\pdfoutput=1
\PassOptionsToPackage{unicode}{hyperref}
\PassOptionsToPackage{hyphens}{url}
\documentclass[
]{article}
\usepackage{xcolor}
\usepackage[letterpaper,top=1in,bottom=1in,left=1.75in,right=1.75in]{geometry}
\usepackage{amsmath,amssymb}
\usepackage{iftex}
\ifPDFTeX
  \usepackage[T1]{fontenc}
  \usepackage[utf8]{inputenc}
  \usepackage{textcomp} 
\else 
  \usepackage{unicode-math} 
  \defaultfontfeatures{Scale=MatchLowercase}
  \defaultfontfeatures[\rmfamily]{Ligatures=TeX,Scale=1}
\fi
\usepackage{lmodern}
\ifPDFTeX\else
\fi
\IfFileExists{upquote.sty}{\usepackage{upquote}}{}
\IfFileExists{microtype.sty}{
  \usepackage[]{microtype}
  \UseMicrotypeSet[protrusion]{basicmath} 
}{}
\usepackage{setspace}
\makeatletter
\@ifundefined{KOMAClassName}{
  \IfFileExists{parskip.sty}{%
    \usepackage{parskip}
  }{
    \setlength{\parindent}{0pt}
    \setlength{\parskip}{6pt plus 2pt minus 1pt}}
}{
  \KOMAoptions{parskip=half}}
\makeatother
\usepackage{longtable,booktabs,array}
\usepackage{calc} 
\usepackage{etoolbox}
\makeatletter
\patchcmd\longtable{\par}{\if@noskipsec\mbox{}\fi\par}{}{}
\makeatother
\IfFileExists{footnotehyper.sty}{\usepackage{footnotehyper}}{\usepackage{footnote}}
\makesavenoteenv{longtable}
\usepackage{graphicx}
\makeatletter
\newsavebox\pandoc@box
\newcommand*\pandocbounded[1]{
  \sbox\pandoc@box{#1}%
  \Gscale@div\@tempa{\textheight}{\dimexpr\ht\pandoc@box+\dp\pandoc@box\relax}%
  \Gscale@div\@tempb{\linewidth}{\wd\pandoc@box}%
  \ifdim\@tempb\p@<\@tempa\p@\let\@tempa\@tempb\fi
  \ifdim\@tempa\p@<\p@\scalebox{\@tempa}{\usebox\pandoc@box}%
  \else\usebox{\pandoc@box}%
  \fi%
}
\def\fps@figure{htbp}
\makeatother
\NewDocumentCommand\citeproctext{}{}

\makeatletter
 \let\@cite@ofmt\@firstofone
 \def\@biblabel#1{}
 \def\@cite#1#2{{#1\if@tempswa , #2\fi}}
\makeatother
\newlength{\cslhangindent}
\newlength{\csllabelwidth}
\newenvironment{CSLReferences}[2] 
 {\begin{list}{}{%
  \setlength{\itemindent}{0pt}
  \setlength{\leftmargin}{0pt}
  \setlength{\parsep}{0pt}
  \ifodd #1
   \setlength{\leftmargin}{\cslhangindent}
   \setlength{\itemindent}{-1\cslhangindent}
  \fi
  \setlength{\itemsep}{#2\baselineskip}}}
 {\end{list}}
\usepackage{calc}

\providecommand{\tightlist}{%
  \setlength{\itemsep}{0pt}\setlength{\parskip}{0pt}}
\usepackage{graphicx}
\usepackage{float}
\usepackage{url}
\usepackage[textfont=it]{caption}
\usepackage{xcolor}
\definecolor{nord0}{HTML}{2E3440}
\definecolor{nord3}{HTML}{4C566A}
\definecolor{nord4}{HTML}{D8DEE9}
\definecolor{nord5}{HTML}{E5E9F0}
\definecolor{nord6}{HTML}{ECEFF4}
\definecolor{nord13}{HTML}{EBCB8B}
\usepackage[skins,breakable]{tcolorbox}
\usepackage{etoolbox}

\renewenvironment{quote}{\begin{tcolorbox}[enhanced,breakable,boxrule=0pt,frame hidden,colback=nord6,sharp corners,borderline west={4pt}{0pt}{nord13},coltext=nord3,fontupper=\itshape,top=12pt,bottom=12pt,left=15pt,right=15pt,before skip=15pt,after skip=15pt]}{\end{tcolorbox}}

\AtBeginDocument{
  \ifcsname Shaded\endcsname
    
  \fi
}

\BeforeBeginEnvironment{verbatim}{\begin{tcolorbox}[breakable,colback=nord6,colframe=nord4,boxrule=0.5pt,arc=4pt,coltext=nord0,left=15pt,right=15pt,top=6pt,bottom=6pt,before skip=15pt,after skip=15pt,fontupper=\footnotesize]}
\AfterEndEnvironment{verbatim}{\end{tcolorbox}}

\usepackage{titlesec}
\titlespacing*{\section}{0pt}{5\baselineskip}{1\baselineskip}
\titlespacing*{\subsection}{0pt}{4\baselineskip}{0.5\baselineskip}
\titleformat{\section}{\LARGE\bfseries}{\thesection}{1em}{}
\titleformat{\subsection}{\Large\bfseries}{\thesubsection}{1em}{}
\titleformat{\subsubsection}{\large\bfseries}{\thesubsubsection}{1em}{}

\makeatletter
\patchcmd{\@maketitle}{\begin{center}}{\begin{center}}{}{}
\patchcmd{\@maketitle}{\LARGE}{\Huge\bfseries}{}{}
\patchcmd{\@maketitle}{\large}{\Large}{}{}
\patchcmd{\@maketitle}{\@date}{\rmfamily\@date}{}{}
\AtBeginDocument{\patchcmd{\@title}{\large}{\LARGE}{}{}}
\makeatother

\usepackage{hyperref}
\usepackage{booktabs}
\usepackage{colortbl}

\let\oldtoprule\toprule
\renewcommand{\toprule}{\arrayrulecolor{nord13}\oldtoprule}
\let\oldmidrule\midrule
\renewcommand{\midrule}{\oldmidrule\arrayrulecolor{nord4}}
\let\oldbottomrule\bottomrule
\renewcommand{\bottomrule}{\arrayrulecolor{nord13}\oldbottomrule\arrayrulecolor{black}}

\hypersetup{hidelinks, breaklinks=true}
\usepackage{adjustbox}
\usepackage{letltxmacro}
\LetLtxMacro{\oldincludegraphics}{\includegraphics}
\renewcommand{\includegraphics}[2][]{\begin{center}\oldincludegraphics[#1]{#2}\end{center}}
\usepackage{bookmark}
\IfFileExists{xurl.sty}{\usepackage{xurl}}{} 
\hypersetup{
  pdftitle={Creating Intelligence},
  pdfauthor={Peter Overmann},
  hidelinks,
  pdfcreator={LaTeX via pandoc}}

\title{Creating Intelligence}
\usepackage{etoolbox}
\makeatletter
\providecommand{\subtitle}[1]{
  \apptocmd{\@title}{\par {\large #1 \par}}{}{}
}
\makeatother
\subtitle{A Computational Foundation for AGI}
\author{Peter
Overmann\thanks{Email: \protect\url{mail@creatingintelligence.org}}}
\date{30 June 2026}

\begin{document}
\maketitle
\begin{abstract}
This work introduces a new computational theory of mind grounded in set
theory and hyperdimensional computing. Whereas traditional neural
networks rely on continuous weights and matrix multiplication, this
framework works with sparse binary data. It represents information as
discrete sets, directly modeling biological neural population codes. I
demonstrate that associative memory emerges naturally from network
topologies featuring a combinatorially expanded hidden layer. Learning
is driven by topological plasticity rather than scalar weight
adjustments. This architecture unifies auto-associative and
hetero-associative learning under a single core algorithm: information
retrieval via subset pattern matching and exact nearest-neighbor search.
Operating with constant-time complexity, these mechanisms bridge
perceptual data (sparse distributed representations) and symbols (sparse
holographic representations) without continuous bottlenecks. Mapping
this framework to neuroanatomy, I propose that both the cerebellum and
the neocortex implement variants of this algorithm, making subset
pattern matching the fundamental engine of cognition. Because it relies
on discrete logic rather than matrix arithmetic, this algorithm
translates directly into in-memory hardware. This opens a new route
toward synthetic intelligence with human-level energy efficiency.

\vspace{1em}

\noindent An open-source, standard-C reference implementation of the
core algorithm is available at
\url{https://github.com/PeterOvermann/CreatingIntelligence}.
\end{abstract}

\setstretch{1.3}

\cleardoublepage
\tableofcontents
\cleardoublepage

\section{A Path Towards AGI}\label{a-path-towards-agi}

Artificial general intelligence (AGI) is usually defined by what it
does: a black box mirroring selected adult human capabilities, most
prominently language generation and academic mastery. But defining AGI
as a target point on a capability trajectory is arbitrary.

Mere knowledge accumulation is a poor proxy for general intelligence.
After all, children possess the same general intelligence as grown-ups
--- they just lack the same volume of training data. True intelligence
is better measured by the agility of the learning process itself: the
capacity to rapidly acquire any novel skill, from language to
programming to driving a car.

Once we replicate this learning agility, human-centric benchmarks serve
only as a baseline. AGI will execute non-human tasks that require
continuous adaptation to fluid environments. A true general AI must
transcend its initial programming, ultimately evolving into new kinds of
synthetic intelligence.

Today's most successful artificial intelligence technologies rely on
neural networks and the deep learning paradigm. This development
originated with Frank Rosenblatt's suggestion (1958) that information is
encoded through adjustable scalar weights, considered to mimic synaptic
strength.

Rosenblatt focused on connection strength rather than network topology.
This set AI on a historical path that prioritizes data fitting and
pattern recognition over structured reasoning. From the original
perceptron, there is a direct line to modern connectionism and to
current transformer architectures. Further leaps of intelligence are
expected from the adoption of multi-modality, embodiment, world models
beyond language, and possibly from doubling down on next-token
prediction.

Combining these approaches may lead to powerful applications that pass
black-box AGI capability benchmarks. But an inspection of the black box
shows that deep neural networks have increasingly diverged from
biological constraints.

The brain is much more energy-efficient (running on about 20 watts,
compared to the megawatts required for LLM clusters). It operates in
continuous time with spiking neurons, makes heavy use of recurrence and
predictive coding, and learns from single experiences without relying on
backpropagation, for which a biological equivalent remains debated.

A future synthetic intelligence may diverge from biology. But it is
worthwhile to learn from nature's blueprints. Evolution has optimized
biological intelligence for exactly the traits current AI lacks: energy
efficiency and continuous learning. Pursuing biological plausibility is
an engineering strategy to overcome the scaling walls of deep learning.
If we were to build a true biologically plausible AGI, it would be an
adaptive, self-organizing system defined by different computational
rules. Here is how we would define its fundamental mechanics:

\begin{itemize}
\tightlist
\item
  Its neurons would send binary signals rather than dense matrices of
  floating-point numbers to each other, spending energy only on pathways
  that are currently active. Such models are known as spiking neural
  networks (SNNs).
\item
  It would use local learning rules, enabling autonomous learning on the
  fly.
\item
  It would be driven by predictive processing: external input only
  travels up the network hierarchy if it contradicts the prediction.
\item
  It would constantly update its world model, adapting to novel
  situations.
\item
  It would allocate its compute resources dynamically --- the analogy of
  neuroplasticity.
\end{itemize}

Following this path, the challenge is to reverse-engineer the mind's
native data structure and the core functions operating on it ---
emulating the discrete behavior of spiking neurons. This data structure
must be a suitable foundation for simulating perception, knowledge
representation, and reasoning. Memory storage and retrieval of this data
structure are an essential part of this blueprint.

Neuroscience gives us an indication of how to approach this challenge. A
biologically plausible computational theory should match the brain's
systemic architecture --- replicating cellular detail is not sufficient.

This approach is theoretically valid because cognitive functions emerge
from the interactions of large populations of neurons, not the molecular
behavior of a single cell. According to David Marr's (1982) trilevel
hierarchy, we can describe a system at a computational level (what it
does), at an algorithmic level (the process it uses), and at an
implementation level (its physical substrate).

This top-down approach faces an obstacle: the \emph{middle level}
problem. We can describe the low-level functioning of a neurotransmitter
in great detail. But we know very little about the brain at a
computational level. We know even less about its underlying algorithmic
processes. Neuroscience is notoriously data-rich but theory-poor.

A purely bottom-up approach is equally challenging: just as inspecting
individual transistors within a CPU tells us nothing about the
computations it executes, cataloging the molecular behavior of single
cells won't explain the brain's overarching computational algorithms.

Therefore, the most realistic path going forward is a \emph{middle-up}
approach. We start at the middle level, inferring core algorithms from
undisputed first principles in neuroscience. And along this route, we
ensure that these processes do not contradict the known biological
implementation.

A coherent computational theory of mind can be grounded in two
foundational principles: population coding and associative memory.

The brain represents information through the collective firing patterns
of neuron populations. Such units of information are known as sparse
binary distributed representations, or sparse hypervectors in the
context of hyperdimensional computation. This data structure is
naturally digital, reflecting the binary all-or-nothing behavior of
neural signals. This view aligns well with spiking neural network
models.

Set theory, and by extension hypergraph theory, provides the most
convenient mathematical description of population coding. Sparse
representations, treated natively as sets, form the foundation for
symbolic knowledge and reasoning.

Associative memory, also called content-addressable memory, is a storage
architecture from which information is retrieved based on its content or
meaning. Conventional computer memory accesses data via specific
addresses. Human memory, by contrast, is associative: it retrieves
information directly by its content, without searching across multiple
addresses.

Fusing both principles results in the notion of a discrete associative
memory for storing sets.

Associative memory systems can reconstruct information from partial
inputs (pattern completion) and are resilient against additive noise.
Therefore, discrete associative memory retrieves information via subset
(or partial) pattern matching.

Based on this premise, the challenge was to find a neurally realistic
mechanism capable of storing sets and performing partial-match queries.
Locality-sensitive hashing, the current standard for set similarity
search, is biologically implausible due to its reliance on preprocessing
and hash tables.

A central discovery of this work is that the desired associative
behavior is an inherent property of certain network topologies. The
location of memory is the topology itself. This implies that information
is represented in discrete form: learning modifies the topology by
switching individual connections on, and the memory decays when those
links disconnect.

Rosenblatt's model relies on continuous synaptic weights. Topological
memory, by contrast, operates natively on binary neuron models. But it
also aligns with alternative learning mechanisms, such as dendritic
plasticity.

Associative topologies require an expanded hidden layer with fixed input
and dynamic output connections. This structure, introduced in my
\emph{Triadic Memory} manuscript (2022) and mirrored in the cerebellum
and the neocortex, is condensed here into a minimal \emph{clean-room}
architecture. The result is a unifying core algorithm for topological
associative memory.

Computationally, the algorithm realizes the exact nearest-neighbor
search within set theory, paralleling the role of vector databases in
continuous-valued embedding spaces. Data retrieval time is independent
of the number of items stored in memory. Due to its scalability and
performance, this architecture should prove useful in technical
applications beyond AGI.

The core algorithm is remarkably concise, fitting into just 50 lines of
high-level code. This brevity strongly indicates biological
plausibility: Brain evolution scales up highly efficient
micro-architectures. If structures as anatomically distinct as the
neocortex and the cerebellum are doing the same thing computationally,
then this points to a substrate-independent mathematical abstraction
which can be expressed by a simple program.

Building upon this new framework, we can already begin to create small
models which replicate core cognitive functions. Cognitive psychologists
have gathered abundant data on how the mind processes information ---
evidence that remains valid as the human mind is not changing. But they
lacked the right architecture to directly translate their insights into
functional software. Closing this gap, the framework introduced here
includes the core routines needed to program with sets, and a
performance-optimized implementation of the topological memory
algorithm.

This is the starting point for experimenting with intelligent systems
that work, at a fundamental computational level, like the human brain.
The winning architecture should be as simple and generic as possible,
leaving room for deriving more complicated and specialized versions.
Along this development roadmap, we may discover compact architectures
that already exhibit forms of general intelligence, for instance basic
language parsing and generation. Developing such a minimal viable AGI is
a realistic research goal. And it would certainly require a smaller
investment than any attempt to find AGI through scaling deep neural
networks.

Scaling topological memory applications into production, whether in
datacenters or in autonomous robots, faces a hardware challenge. The
computational technology discussed here has little in common with
conventional AI systems. Being based solely on binary model parameters
and bit-level operations, topological associative memory does not
involve floating-point matrices and therefore would not benefit from
contemporary graphics or tensor processing hardware. To scale these
models most effectively, computation must occur within memory, much like
biological brains operate: There would be no memory wall (the von
Neumann bottleneck) at all.

This is a new strategy to reduce the power consumption of AI to a human
scale. Ideas for a hypothetical implementation using DRAM and
interconnecting memory units are outlined in this document.

\section{The Mind's Data Structure}\label{the-minds-data-structure}

\subsection{Population coding}\label{population-coding}

Although their internal mechanisms are fundamentally analog, neurons
simulate digital codes by communicating via discrete electrical pulses.
These all-or-nothing events translate to binary codes 1 (spike) or 0 (no
spike).

The apparently digital output of neurons is usually classified into two
broad schemes: population coding and temporal/frequency-based coding.
Additionally, neural digital signals may transmit control codes that do
not carry cognitive information.

The algorithmic and computational model discussed here is derived solely
from the phenomenon of population coding. Collective firing patterns
represent tokens of information and composite data structures, forming
the basis for higher-level cognitive information and reasoning.

While our focus is on static population coding, it is worth addressing
the role of temporal mechanisms in this framework. The \emph{temporal
coding hypothesis} posits that precise, millisecond-level timing and
patterning of individual neuronal spikes carry essential, high-fidelity
information. Temporal firing patterns naturally occur during sensory
processing and motor control. In these systems, temporal patterns act as
a translator, converting static population codes into dynamic, real-time
physical events. But pure temporal coding is unlikely to play a
fundamental role in memory or symbolic thought.

Instead, in the context of population coding, temporal effects serve a
supporting role: they synchronize, aggregate, and sustain neural
signals. Nearly synchronous spiking is clearly a prerequisite for
population codes to materialize. Synchrony can be achieved if neurons
repeatedly fire at irregular time intervals until a critical number of
spikes occur within the same small time window. The phenomenon known as
\emph{temporal multiplexing}, observed in the
\hyperref[neocortex]{neocortex}, aligns well with this aspect of
population coding. While many researchers hypothesize that the
microtiming of multiplexing encodes information, this framework provides
a more direct explanation: the timing simply reflects the sequence of a
neural layer executing a memory query and subsequently presenting the
retrieved response.

Population coding is a natural implementation of a distributed
representation. Where a local representation relies on a single cell
(often referred to as a \emph{grandmother cell}) to represent a concept,
a distributed representation spreads information across multiple
neurons. An individual neuron may participate in multiple distributed
representations: it does not represent any particular concept on its
own. Encoding information with a set of neurons yields abundant
representational capacity as well as the necessary fault tolerance.

Extreme sparsity is the defining feature of population codes.
Ninety-nine percent or more of the neurons are silent at any moment.
This is a main factor in the brain's energy efficiency. The total pool
of neurons involved in a computational process (the embedding dimension)
is quite large, typically ranging from a thousand to ten thousand.

Mathematically, population codes are simply sets --- specifically small
subsets drawn from a high-dimensional universal set.

The notion of sets representing the mind's native data structure is the
starting point from which we derive a coherent computational and
algorithmic model of cognition.

\subsection{The blessing of
dimensionality}\label{the-blessing-of-dimensionality}

Our computational framework builds on hyperdimensional computing, a
brain-inspired approach from the late 1980s that uses very
high-dimensional vectors to represent information. These models include
sparse distributed memory (Kanerva 1988), holographic reduced
representations (Plate 1995), and vector-symbolic architectures (Gayler
2003).

Conventional neural networks suffer from the \emph{curse of
dimensionality}. Here, increasing dimensions (typically between 1,000
and 10,000) actually makes computing easier: Hyperdimensional models
benefit from the \emph{blessing of dimensionality}.

\begin{enumerate}
\def\labelenumi{\arabic{enumi}.}
\item
  In high-dimensional space, any two random vectors are almost
  guaranteed to be nearly orthogonal. Because the number of these
  quasi-orthogonal directions grows exponentially with dimensions, the
  system can represent millions of distinct concepts with almost zero
  risk of mixing them up.
\item
  Hyperdimensional representations are robust to noise: since
  information is distributed across thousands of dimensions, changing or
  losing a few hundred bits of a hypervector barely shifts its position
  in space. It remains much closer to its original value than to any
  other unrelated hypervector.
\item
  Hyperdimensional spaces are \emph{warped} such that complex non-linear
  patterns in low dimensions can become linearly separable in high
  dimensions. This allows one-pass learning, where models can be trained
  by simple vector addition rather than iterative backpropagation.
\end{enumerate}

These ``blessings'' are common to dense and sparse models of
hyperdimensional computation. Dense models represent information as real
or bipolar (-1 and +1) values uniformly distributed over all dimensions.
Mathematical operations involving dense vectors can be defined
coherently to preserve both the density and the uniform distribution of
representations.

Dense models may be mathematically elegant. But because they involve
every dimension in every calculation, they do not capture the sparse
activity patterns and computational efficiency of the biological brain.

Yet, working effectively with sparse hypervectors comes with its own
conceptual challenges, compared to dense models. Viewed in isolation,
sparse binary hypervectors are a data structure with limited use. It
might have been the strict adherence to the vector metaphor that
hindered progress with sparse models.

A much more effective reference frame for hyperdimensional computation
is set theory: Binary hypervectors, interpreted as \emph{characteristic
vectors}, correspond directly to sets. In the context of sparse
distributed representations, we are particularly interested in small
subsets of a universal set corresponding to the hyperdimensional space.

With this shift in perspective, set theory, and by extension subset
combinatorics and hypergraph theory, turns out to be a solid foundation
for representing and manipulating structured information. And
associative memory emerges as the essential mechanism for storing and
transforming these set-based representations.

\subsection{Sets representing
information}\label{sets-representing-information}

As a mathematical model for neural population codes, sets provide an
abundant representational capacity. Consider a universal set of 2,000
elements, representing the total available population of neurons (the
embedding dimension \(N\)). In a \emph{localist} or \emph{one-hot}
encoding scheme where each element maps to a single piece of
information, only 2,000 distinct items can be represented. But if we
distribute information across multiple elements, the capacity increases
by orders of magnitude. The number of all possible sets with \(P\) out
of \(N\) elements --- the \emph{representational capacity} --- is given
by \(\binom{N}{P}\).

Consider a concrete set with dimension \(N = 2,000\) and population
\(P = 10\):

\begin{verbatim}
{158 368 379 387 568 583 618 1012 1025 1966}
\end{verbatim}

We use numbers from 1 to \(N\) to enumerate the set's elements. By
convention, the elements are displayed in order, but this does not
indicate any intrinsic ordering.

Here is the same set, with its elements plotted as dots within an area
covering 2,000 possible locations:

\begin{figure}[H]
\centering
\includegraphics[width=0.292\linewidth,height=\textheight,keepaspectratio,alt={A sparse set with 10 out of 2,000 possible elements.}]{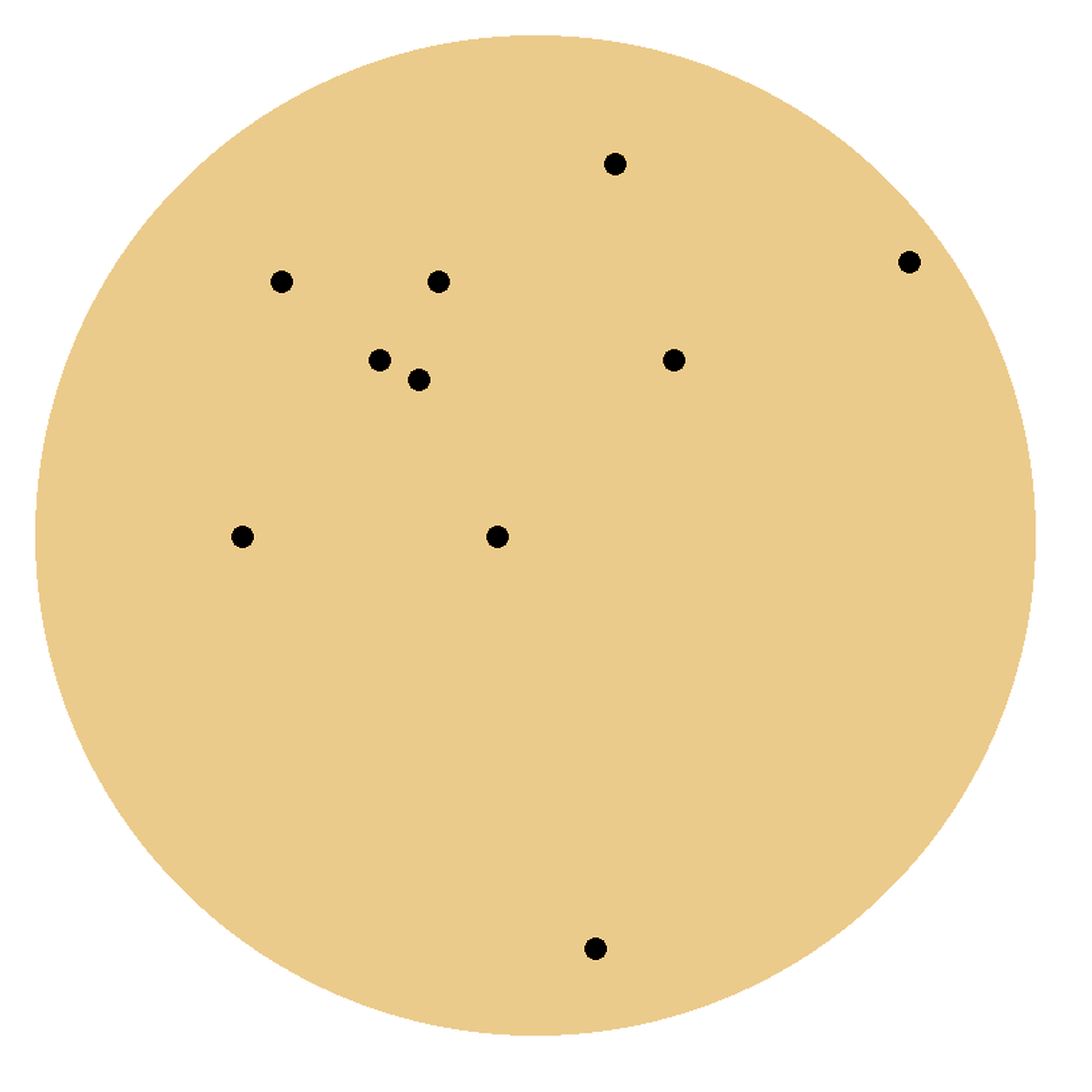}
\caption{A sparse set with 10 out of 2,000 possible elements.}
\end{figure}

This image should convey the scale of the representational space defined
by these small subsets. There are approximately \(2.76 \times 10^{26}\)
unique subsets containing exactly 10 elements --- exceeding any volume
of information that we may want to encode.

The tendency towards orthogonality of hypervectors aligns with subset
combinatorics, particularly regarding the probability of set overlap.
Two random sets usually share no or very few elements. Significant or
even complete overlap is statistically improbable. This probability
distribution is shown here as a table, for random sets with cardinality
\(P = 10\) within an overall dimension \(N = 1,000\):

\begin{verbatim}
 overlap t      probability
      0           0.904
      1           0.09215
      2           0.0038
      3           0.00008248
      4           0.000001027
      5           0.000000007505
      6           0.00000000003172
      7           0.00000000000007345
      8           0.00000000000000008363
      9           0.00000000000000000003758
     10           0.000000000000000000000003796
\end{verbatim}

Generally, the probability that two random sets with equal dimension
\(N\) and respective populations \(P_1\) and \(P_2\) share \(t\)
elements is given by the hypergeometric distribution

\[p(t) = \binom{P_1}{t} \binom{N - P_1}{P_2 - t} \binom{N}{P_2}^{-1}\]

The overlap score, defined as the size of the intersection between two
sets, serves as the metric for similarity. These overlap probabilities
are extremely skewed. As a result, any given set inhabits a minute
neighborhood of similar sets. Beyond this tiny neighborhood, the set is
highly dissimilar to everything else found within the vast
representational space.

An interesting consequence: if a corpus of information is encoded as
random sets, these sets are not only mutually distinct, but also highly
dissimilar to any arbitrary set. This phenomenon allows us to clearly
differentiate between known and unknown information. A nearest-neighbor
search originating from a random set almost never hits known data. There
is almost no middle ground between nearly identical and totally
unrelated sets.

Contrast this with standard vector spaces, which rely on \emph{dense
semantic gradients} for tasks like RAG or image retrieval. In those
continuous spaces, a vector search almost always returns a unique best
match, even if that resulting match bears no meaningful similarity to
the query.

\subsection{Sparse holographic
representations}\label{sparse-holographic-representations}

A central challenge in symbolic AI is the formal representation of
information: what defines a symbol as the fundamental unit of
information, and by what mechanisms are these symbols organized into
nested, composite data structures?

Information units encoded as small subsets within a hyperdimensional
space must resist degradation. If parts of the set are lost, the
representation must retain its integrity --- a defining feature of
distributed systems. In other words, subsets must represent the same
thing as the full set. The whole is in every part.

This leads us to the definition of a sparse holographic representation
(SHR) as a set where all possible subsets (the power set) represent the
same information. An SHR distributes data evenly, meaning no individual
element is tied to a specific subfeature. Mathematically, every possible
subset --- even a single element --- carries the full semantic weight of
the entire concept. Practical memory retrieval requires a sufficiently
large subset to statistically isolate that meaning against the
background noise of other overlapping representations.

Because the empty set is trivially a subset of all SHRs, querying with
an empty or completely unrecognized pattern yields no unique resolution:
it is a superposition of all possible knowledge.

\begin{quote}
\textbf{Sparse holographic representation (SHR)}\\
A set where all possible subsets represent the same semantic
information. Used as an encoding scheme for symbolic tokens.
\end{quote}

In formal logic, if a set and all its subsets map to the exact same
semantic value, the set is treated as a singleton in the information
space. Although it contains multiple elements, it is functionally a
single atom of information.

For practical purposes, a sparse holographic representation is simply
encoded as a random set. Different SHRs may randomly overlap in a few
elements, but such small overlap does not indicate information
similarity. Two random sparse sets are extremely unlikely to share more
than a few elements.

SHRs may represent semantic primitives (like ``red'' or ``apple''),
chunks of information (``red apple''), non-lexical and higher-level
concepts, or pointers to data structures.

Although the information represented by an SHR may have some kind of
substructure, the SHR has by definition no internal structure: every
element carries the same meaning. This is the key to compositionality,
or more precisely its inverse: \emph{decompositionality}. We can easily
compose multiple sets by forming their union. But in order to decompose
that union and recover its original constituents, we must group elements
by their meaning, and this is only possible for holographic sets.
Decomposition would fail if the sets involved had overlapping
substructures.

\subsection{Sparse distributed
representations}\label{sparse-distributed-representations}

Sparse distributed representations (SDRs) are an encoding scheme where
real-world information is represented as sparse binary hypervectors,
which we treat as sets. This data concept is fully compatible with the
set-based framework introduced here. In particular, topological memory
and related algorithms operate directly on SDRs. This complements
existing SDR-based applications with a scalable and performant
associative memory driven by exact nearest-neighbor search and subset
pattern matching.

The concept of SDRs, popularized by Hawkins and \emph{Numenta} (see for
example Ahmad (2015) for an overview), is fully aligned with this
set-theoretic approach. Considered the theoretical universal language of
the neocortex, SDRs can encode any data type as long as the encoding
captures its underlying semantic characteristics. Each element within an
SDR encodes semantic meaning, but because the representation is
distributed, an element may participate in unrelated SDRs. It is the
concurrent activation of the entire set that disambiguates the specific
meaning.

\begin{quote}
\textbf{Sparse distributed representation (SDR)}\\
An encoding scheme representing perceptual data as sparse sets within a
hyperdimensional space. Active elements correspond to shared semantic
features.
\end{quote}

This data structure is complementary to sparse holographic
representations (SHRs), where information is distributed such that any
subset remains representative. Where SDRs are foundational for upstream
perceptual data, SHRs are the basis for higher-level cognitive functions
like language, where compositionality is essential.

Topological memory bridges the gap between perception and symbolic
reasoning: perceptual signals (SDRs) as well as symbols (SHRs) are
universally represented by sets. They differ in their overlap
characteristics and semantic interpretation, but are processed at an
elementary level by the same mechanisms.

\subsection{SDR encoders}\label{sdr-encoders}

Specialized SDR encoders have been developed to handle various data
types.

Continuous numeric data, ranging from physical measurements to financial
information, is mapped to SDRs such that similar values produce
overlapping patterns.

Language elements (words, sentences, etc.) are traditionally converted
into SDRs where items with similar meanings share overlapping elements.
But note that in our framework, symbols must be encoded as random SHRs.
This means \hyperref[semantic-similarity]{semantic similarity} between
symbols arises from graph distance and spreading activation rather than
representational overlap.

Raw sensory data from the physical world, such as images and sounds, can
be encoded into sparse binary representations. A canonical example is
the \hyperref[classification-of-handwritten-digits]{classification of
handwritten digits}, where raster images are encoded as SDRs and
subsequently classified through a hetero-associative memory.

Categorical data, such as discrete labels or behaviors, can be
represented as SDRs, often using encoders that preserve the logical
relationships between different categories.

Spatiotemporal data, for instance geolocations and timestamps, are
encoded as SDRs to model physical positions and chronological movement.

Dense vector embeddings (real-valued low-dimensional arrays) from
conventional deep learning models can be transformed into sparse binary
hypervectors.

Grid cell-like encodings, a technique inspired by neuroscience,
represent an object's position relative to a reference frame. Unlike
conventional coordinates, these encodings use multiple periodic scales
and orientations to create a unique signature for every location in a
space. Application areas include physical navigation, path integration,
robotic movement, abstract cognitive maps, or positional encoding
augmenting transformer models. Similar mechanisms provide the temporal
scaffolding required to anchor episodic memories.

\subsection{Grounding symbols in
perception}\label{grounding-symbols-in-perception}

The \emph{grounding problem} is a fundamental challenge in artificial
intelligence: how do the abstract symbols (words, numbers, etc.) that an
AI manipulates acquire actual, real-world meaning?

Modern deep learning and LLMs process human language as a vast system of
syntax. They learn the mathematical probability of how tokens fit
together, much like a person locked in a room with a dictionary. They
can perfectly simulate grammatical rules. But because their symbols only
point to other symbols, they lack semantics: the meaning of symbols in
the physical world.

Within the topological memory framework, sensory perception and symbolic
cognition are founded on the same data structure: discrete sets.
Associative memory retrieval functions as a transcoder between the data
domains.

Here is a brief outline of this computational model:

Raw perceptual data from the physical world (vision, audio, etc.) is
transduced into overlapping SDRs. In an SDR, elements directly represent
shared semantic and physical features. We can leverage a host of
pre-existing SDR encoders to handle this raw sensory transduction.

High-level concepts and symbolic tokens (words, numbers, etc.) are
encoded as orthogonal SHRs. These are random, orthogonal sets where no
single element holds specific sub-feature meaning. Both SDRs and SHRs
are natively processed within the same computational pipeline.

A symbol is grounded when its abstract SHR is bound to the physical SDR
of that object. Computationally, this is the result of
\hyperref[heteroencoder]{heteroencoding}. For example, the perception of
the color red is encoded as a class of highly overlapping SDRs, each
representing different impressions of red. Heteroencoding maps these to
a stable SHR: the symbol ``red'' is no longer a floating syntactic
token, but is deterministically anchored to a physical sensory stream.

This mechanism preserves the nuanced gradients of perception and
simultaneously generates the composition-ready symbolic tokens required
for knowledge representation and reasoning.

Generalizing this approach beyond human-centric physical embodiment, a
synthetic intelligence can operate on any kind of sensory data and
condense it into non-human symbols --- its physical world is the data
stream. And of course, this sensory grounding is optional. A
language-only intelligence can bypass the perception pipeline.
Tokenizing inputs directly into symbolic representations, it would
operate within the realm of ungrounded syntax.

\section{Topological Associative
Memory}\label{topological-associative-memory}

The brain encodes information through populations of co-active neurons.
Learning and retrieving these patterns requires an associative memory
mechanism capable of direct access. Cognitive psychology confirms this:
the mind retrieves information directly by its content, rather than
scanning through memory locations. Human memory has an enormous storage
capacity, and it learns quickly.

The data objects stored in this content-addressable memory are
population codes: relatively small, discrete subsets of a very large
group of neurons.

A biologically plausible mechanism for this kind of memory has remained
elusive for decades. Without a direct mechanism to store and retrieve
these population codes, artificial intelligence has had to build its
computational foundation on non-biological approximations, such as
continuous weights. We address this gap by showing that discrete
associative memory is inherent in the network topology itself.

This associative behavior emerges naturally from random network topology
through a fundamental mechanism: a combinatorially expanded hidden
layer. Each node in this hidden layer receives at least two fixed
connections from the input layer. It also forms dense, dynamic links
with the output layer. These links are the location of memory and
learning.

Paths between the hidden layer and the target output are opened when
storing information. Memory traces fade when these paths close. Memory
retrieval activates hidden layer nodes that receive at least two binary
inputs. Those activations are then summed up and thresholded to produce
an output with the desired sparsity level.

\begin{quote}
\textbf{Topological associative memory}\\
Discrete associative memory for sparse sets, emerging from combinatorial
network expansion. Learning occurs by switching binary connections,
while \hyperref[k-winners-take-all]{k-winners-take-all} serves as the
retrieval engine.
\end{quote}

This mechanism does not involve continuous connection weights. In a
departure from conventional neural networks, hidden layer activation is
modeled with binary inputs and a simple excitatory threshold, analogous
to the original artificial neuron proposed by Warren McCulloch and
Walter Pitts (1943).

We model learning simply through the opening and closing of signal
paths, abstracting away the biological specifics of neural plasticity.
Topological plasticity becomes the generalized mechanism of associative
learning. The underlying biological realization may be synaptic
long-term potentiation (LTP), dendritic plasticity, or other phenomena.
Additionally, topological plasticity may involve a combination of such
neural mechanisms. This abstraction is supported by experimental
observations: the rapid formation and elimination of dendritic spines
combined with the activation of ``silent'' synapses indicate that the
brain uses switch-like mechanisms to alter its effective network
topology.

\begin{quote}
\textbf{Topological plasticity}\\
The generalized mechanism of associative learning in this framework.
Memory traces are formed or eliminated by switching binary network
connections (opening and closing signal paths) rather than adjusting
continuous synaptic weights.
\end{quote}

The concept of generalized topological plasticity is compatible with
Donald Hebb's original postulate, famously summarized as ``neurons that
fire together, wire together''. Although the concept of Hebbian learning
is commonly conflated with the theory that continuous synaptic weights
represent information, this \emph{perceptron} model did not originate
with Hebb.

Associative network topology requires a combinatorially expanded hidden
layer: it is randomly connected to all \(N\) input nodes, with a minimum
of two inputs projecting into each hidden layer node. This means there
are at least \(N(N-1)/2\) hidden nodes --- a combinatorial fan-out by a
factor \((N-1)/2\). This design principle is known as expansion coding.

\begin{quote}
\textbf{Expansion coding}\\
A network design using a combinatorially expanded hidden layer. Each
hidden node receives fixed connections from a pair of input nodes.
\end{quote}

In an idealized ``canonical'' topology, exactly \(N(N-1)/2\) hidden
nodes receive exactly two connections from the input layer. This is a
valid approximation of biological networks because simultaneous
excitations across three or more input nodes are rare due to the input's
sparsity. A hidden layer node connected to three input nodes is shared
by three possible input pairs, and four connections correspond to six
input pairs. This setup can reduce the hidden layer's size without
affecting the core mechanism of associative learning, though it results
in a comparatively lower memory capacity.

The purpose of this \(\mathcal{O}(N^2)\) fan-out is not to separate
patterns to reduce interference, as in conventional networks, but to
create massive storage redundancy. When an association is stored, every
pair of active input elements activates a specific hidden node, which
then forms a discrete connection to the target output. Because the
memory is distributed across this redundant web of links, querying the
system with only a small subset of the original pattern activates enough
links to reconstruct the whole. This structural redundancy is the
foundation that enables pattern completion, subset matching, and the
memory's abundant storage capacity.

In this model, learning is a distributed yet localized process: a single
write operation affects only a fraction of memory locations without
causing a ripple effect within the entire space and therefore avoiding
catastrophic forgetting.

\begin{quote}
\textbf{Local learning}\\
Topological memory stores new information through local memory updates,
rather than global weight adjustments.
\end{quote}

The following figure shows the canonical topology for a small number of
connections. Real-world associative memories must involve much larger
layers than this example suggests, with typical inputs and outputs
ranging from 1,000 to 10,000 nodes.

\begin{figure}[H]
\centering
\includegraphics[width=0.5\linewidth,height=\textheight,keepaspectratio,alt={Hetero-associative topology with 6 inputs (bottom) and 7 outputs (top) which are fully connected to the hidden layer.}]{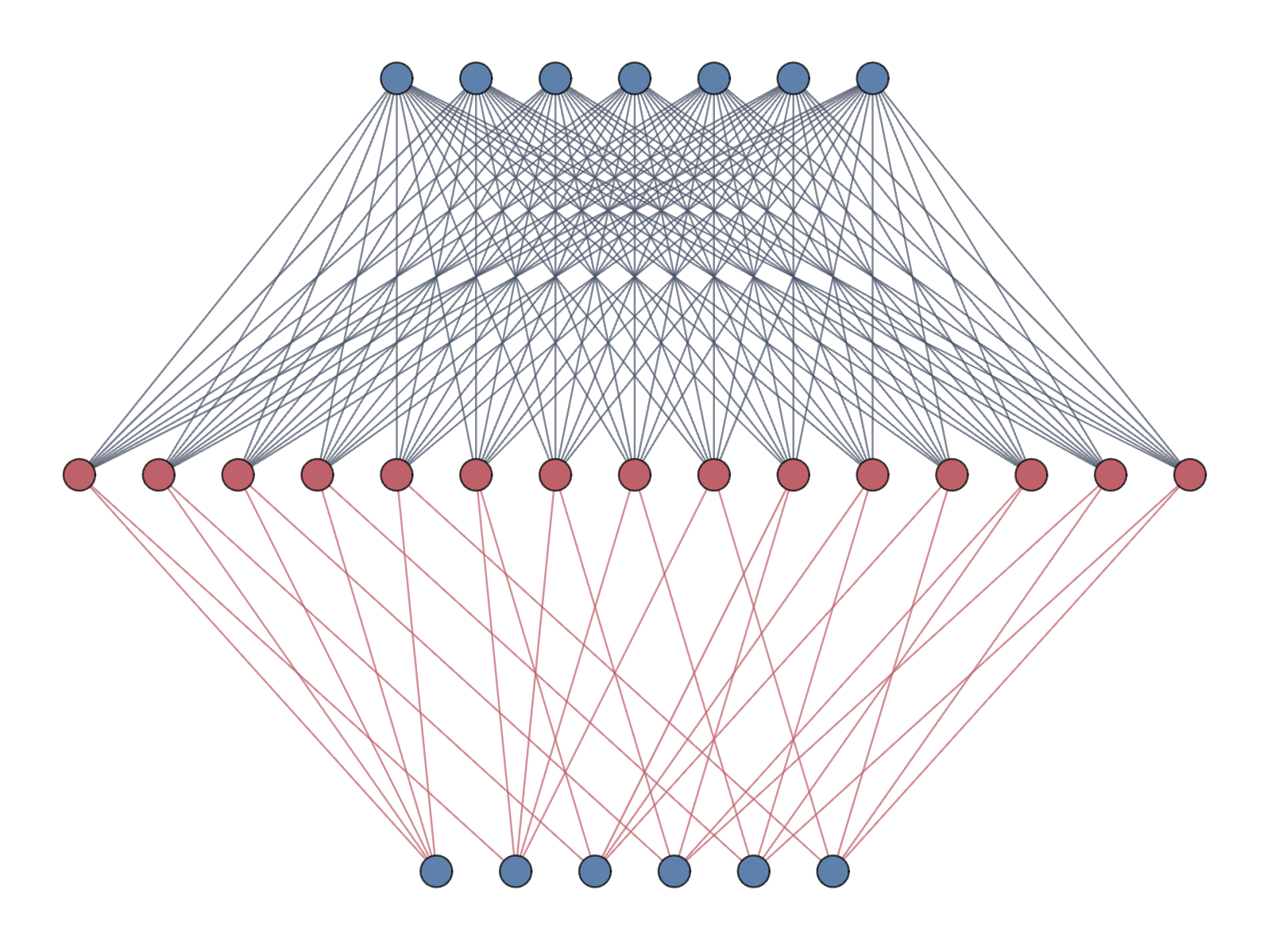}
\caption{Hetero-associative topology with 6 inputs (bottom) and 7
outputs (top) which are fully connected to the hidden layer.}
\end{figure}

This type of feed-forward topology, with its physically distinct input
and output layers, is inherently hetero-associative. It requires a
secondary incoming pathway for training the output layer. The input and
output layers generally have different dimensions, representing
information from distinct domains. This topology aligns with the overall
architecture of the \hyperref[cerebellum]{cerebellum} --- a brain
structure that initially evolved to associate sensory inputs to motor
control outputs.

Auto-associative behavior emerges from a recurrent topology with a
massively expanded hidden layer. Here the input signal is routed to the
latent space and back into the input layer, updating the system's state.
A single neural layer acts simultaneously as the input receiver and the
output state. Consequently, the input and the output have identical
dimensions and represent the same kind of information. This architecture
broadly matches the topology of the \hyperref[neocortex]{neocortex},
which is widely assumed to perform auto-associative functions.

\begin{figure}[H]
\centering
\includegraphics[width=0.458\linewidth,height=\textheight,keepaspectratio,alt={Auto-associative topology with 6 input/output dimensions (blue). The 15-node latent space is shown in red.}]{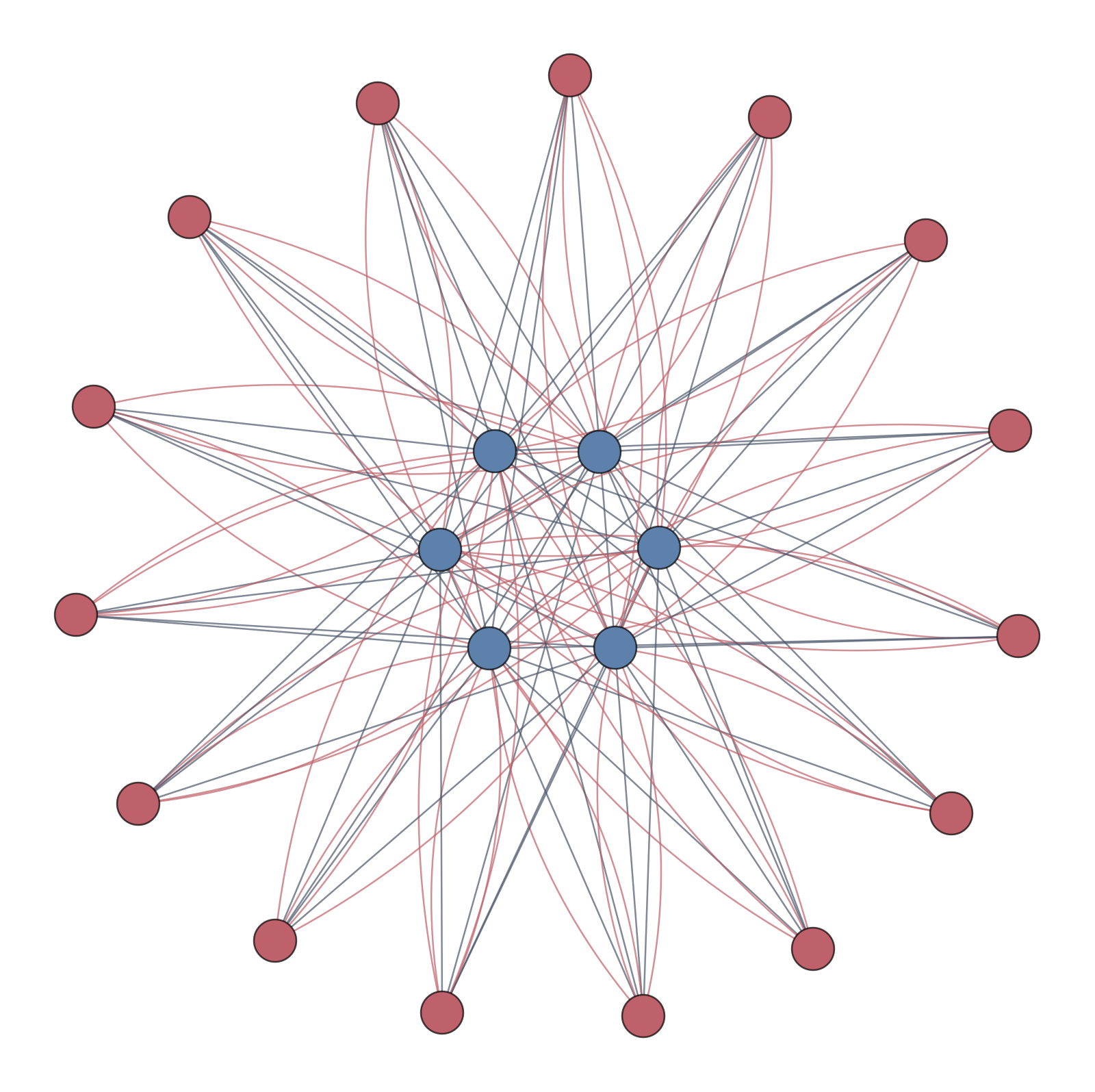}
\caption{Auto-associative topology with 6 input/output dimensions
(blue). The 15-node latent space is shown in red.}
\end{figure}

Both hetero-associative and auto-associative topologies are technically
single-layer networks in standard neural network terminology. The
connections from the input to the hidden layer are static, performing
expansion coding as a preprocessing step. The only part of the network
that can learn via adjustable topological paths is the final connection
to the output layer.

A side-by-side comparison shows the architectural difference between
recurrent auto-associative and feed-forward hetero-associative
architectures. Despite their structural differences, both types execute
the same topological associative memory algorithm.

\begin{figure}[H]
\centering
\includegraphics[width=0.792\linewidth,height=\textheight,keepaspectratio,alt={Left: Recurrent network topology, storing autoassociations A\textbackslash to A. Right: Feed-forward network with two inputs for heteroassociations A\textbackslash to B.}]{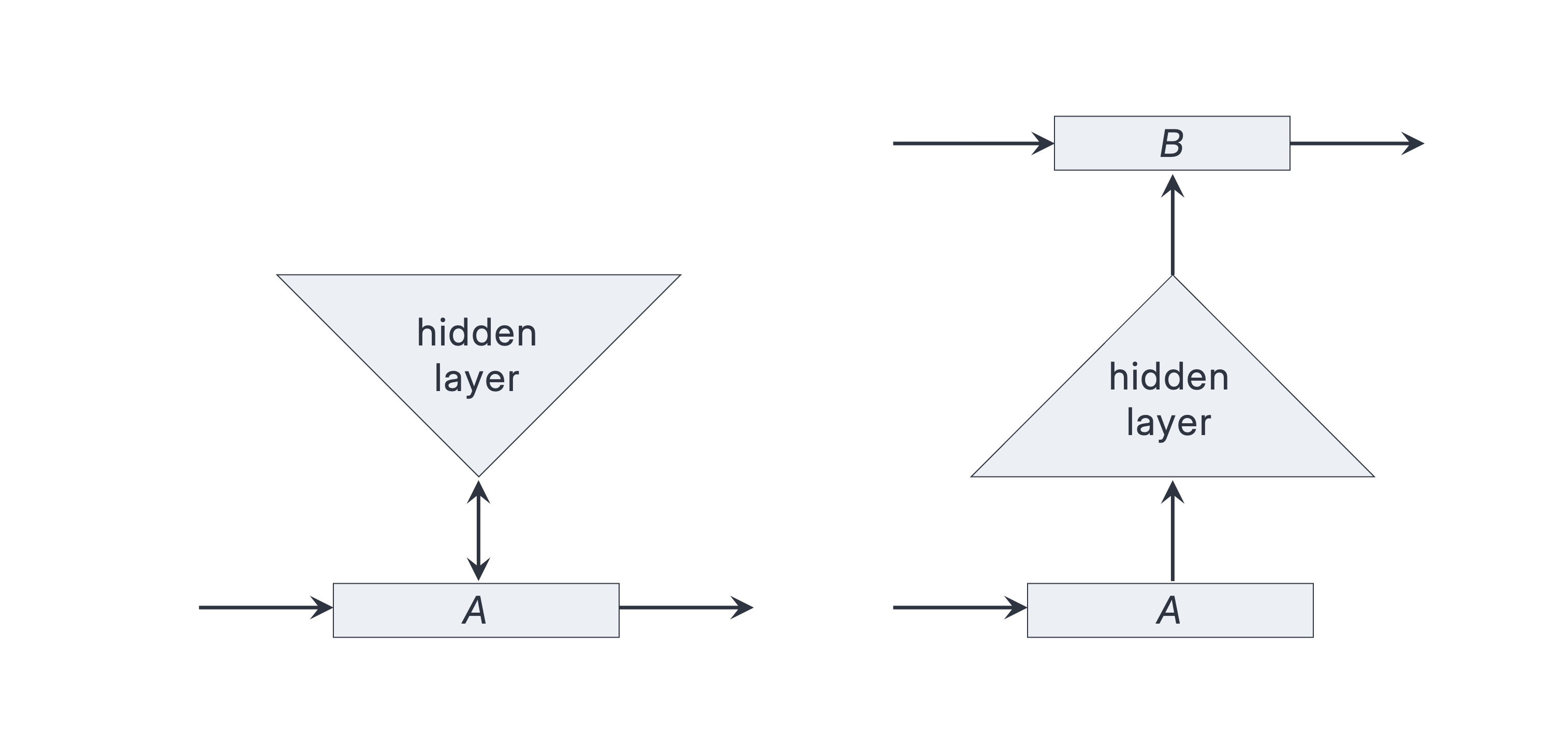}
\caption{Left: Recurrent network topology, storing autoassociations
\(A\to A\). Right: Feed-forward network with two inputs for
heteroassociations \(A\to B\).}
\end{figure}

In a biological network, the hidden layer can manifest either as a
massive population of neurons (instantly recognizable in the
\hyperref[cerebellum]{cerebellum}) or as an expansive dendritic network
(characteristic of the \hyperref[neocortex]{neocortex}). Despite their
pronounced anatomical differences, both structures are suitable to
execute the same abstract computation: an associative memory for
discrete sets, with retrieval operating via subset pattern matching.
Both the cerebellum and the neocortex have been hypothesized to be
associative memories. Both are modeled by the same computational
function within the topological memory framework.

\section{What topological memory is
doing}\label{what-topological-memory-is-doing}

Here we examine the precise mechanics of storing and retrieving sets, at
the level of explicit set elements. Later sections abstract these into
higher-level symbols.

To demonstrate this, imagine we are interacting with a topological
memory engine through a command-line interface. In the examples below,
lines starting with \texttt{\textgreater{}} are the user's input prompt,
and subsequent lines show the memory's response.

(Note: a command-line tool with equivalent functionality but a different
syntax is included in the standard-C reference implementation.)

Topological memory can indeed be used interactively: information is
stored in a single pass and can be immediately retrieved.

At the session start, we instantiate a memory for heteroassociations
\(A \to B\) with hyperparameters \(N_A = 1,000\), \(N_B = 500\)
(dimensions), \(P_A = 10\), and \(P_B = 5\) (populations). Note that
this memory is asymmetric with respect to its input and output
parameters.

\begin{verbatim}
> memory 1000 10 -> 500 5
\end{verbatim}

Specify sets \(A\) and \(B\):

\begin{verbatim}
> A := {198 207 288 335 414 633 643 782 808 960}
> B := {4 185 253 354 393}
\end{verbatim}

In our convention, set elements are numbered from 1 to \(N\) (the
universal set's cardinality).

Store the heteroassociation \(A \to B\):

\begin{verbatim}
> A -> B
\end{verbatim}

Querying the memory with \(A\) accurately retrieves \(B\):

\begin{verbatim}
> A
{4 185 253 354 393}
\end{verbatim}

Retrieving the original association from noisy input (additive noise)
returns the same answer:

\begin{verbatim}
> {29 137 198 207 248 281 288 335 357 414 499 600 633
     643 782 784 808 960}
{4 185 253 354 393}
\end{verbatim}

We can also retrieve the same association from an incomplete input
(subtractive noise):

\begin{verbatim}
> {198 207 335 414 633 808 960}
{4 185 253 354 393}
\end{verbatim}

As one can see from these examples, this associative memory works
differently from the \emph{association} (or \emph{dictionary}) data
structures in modern programming. Lookup tables rely on hashing for
exact matches. They cannot perform the kind of subset queries shown
here.

In this specific configuration, at least 7 input elements must match.
The pattern matching threshold is automatically derived from
hyperparameters \(N_A\) and \(P_A\), unless a user-defined threshold
overrides the default value.

Accordingly, pattern matching with only 6 of the original input elements
fails: the memory query returns an empty set.

\begin{verbatim}
> {198 207 335 414 633 808}
{}
\end{verbatim}

Querying with an unknown input pattern reliably gives an empty set:

\begin{verbatim}
> {171 175 403 412 417 434 483 504 808 902}
{}
\end{verbatim}

A false positive outcome is extremely unlikely. In this specific
configuration, the probability of random input correctly matching 7
elements is 0.00000000000007345. This constraint suppresses
hallucinations resulting from false-positive memory retrieval.

The memory can handle associations with different inputs pointing to the
same output (many-to-one associations):

\begin{verbatim}
> A2 := {2 88 150 240 311 490 555 601 777 890}; A2 -> B;
> A2
{4 185 253 354 393}
\end{verbatim}

Storing one-to-many associations combines the right-hand sides (outputs)
in memory:

\begin{verbatim}
> A2 -> {1 2 13 266 480};
> A2
{1 2 4 13 185 253 266 354 393 480}
\end{verbatim}

Querying with a union of equal-sized, non-overlapping patterns creates a
tie. The system cannot return a unique best match:

\begin{verbatim}
> {167 279 453 585 835 836 906 980} -> {3 159 180 241 403}
> {338 442 553 718 784 886 890 977} -> {41 70 395 465 492}
> {167 279 338 442 453 553 585 718 784 835 836 886 890
      906 977 980}
{}
\end{verbatim}

Removing any one element from the union breaks the tie:

\begin{verbatim}
> {279 338 442 453 553 585 718 784 835 836 886 890 906 977 980}
{41 70 395 465 492}
\end{verbatim}

Now we subtract the matching subset to retrieve the remaining pattern.

\begin{verbatim}
> {167 279 453 585 835 836 906 980}
{3 159 180 241 403}
\end{verbatim}

What we see here is the extraction of an item from a union via
randomized subsampling. This process is the basis for
\hyperref[iterative-decomposition]{iterative decomposition} of bundled
data.

Next we instantiate a second, independent memory where inputs and
outputs have the same dimension \(N = 1,000\) and uniform population
\(P = 10\):

\begin{verbatim}
> memory 1000 10 -> 1000 10
\end{verbatim}

This symmetric memory can equally store heteroassociations \(A \to B\)
and autoassociations \(A \to A\), and even permits a mixture of both
types. The underlying topological associative memory algorithm makes no
distinction between these different kinds of associations.

A memory that is exclusively auto-associative is called an item memory,
cleanup memory, or item store.

This defines a set \(D\) and stores it into the auto-associative memory:

\begin{verbatim}
> D := {5 9 77 254 329 447 637 754 770 856}; D -> D;
\end{verbatim}

Querying with set \(D\):

\begin{verbatim}
> D
{5 9 77 254 329 447 637 754 770 856}
\end{verbatim}

This particular memory configuration has the capacity to store about
959,818 distinct autoassociations.

Querying with a very noisy version of the original pattern cleans up the
input:

\begin{verbatim}
> {2 5 6 9 25 77 120 124 148 160 202 214 254 276 282 284 294
    300 304 329 346 354 413 426 441 447 463 474 475 480 511
    517 554 559 560 561 568 581 593 629 637 654 671 683 686
    713 734 754 770 774 819 832 856 900 902 916 926 941 960}
{5 9 77 254 329 447 637 754 770 856}
\end{verbatim}

A partial (subset) query completes the missing elements:

\begin{verbatim}
> {9 77 329 447 637 754 856}
{5 9 77 254 329 447 637 754 770 856} 
\end{verbatim}

This mechanism is known as pattern completion --- an essential process
in cognition.

Storing another pattern \(E\) in memory and querying with the bundle
\(D \cup E\) (denoted in the CLI by simply placing the variables
side-by-side) gives an empty set: no unique best match can be determined
in this case.

\begin{verbatim}
> E := {102 105 134 258 275 602 625 791 798 886}; E -> E;
> D E
{} 
\end{verbatim}

Randomized subsampling of \(D \cup E\) extracts one of the items:

\begin{verbatim}
> {5 9 102 254 258 275 329 447 602 625 637 754 770 791 798
    856 886}
{5 9 77 254 329 447 637 754 770 856} 
\end{verbatim}

The memory separates known from unknown information:

\begin{verbatim}
> {34 55 56 80 138 339 360 424 462 515 548 589 860}
{} 
\end{verbatim}

The examples shown so far involve random tokens with no or small mutual
overlap. But topological memory can also distinguish between nearly
identical patterns, differing in just one element:

\begin{verbatim}
> F := {1 2 3 4 5 6 7 8 9 10}; F -> F
> G := {1 2 3 4 5 6 7 8 9 11}; G -> G
\end{verbatim}

The memory retrieves exact matches:

\begin{verbatim}
> F
{1 2 3 4 5 6 7 8 9 10}

> G
{1 2 3 4 5 6 7 8 9 11}
\end{verbatim}

\section{The Core Algorithm}\label{the-core-algorithm}

\subsection{Functional principles}\label{functional-principles}

This section outlines the functional principles of topological memory,
serving as a usage and implementation guide.

A unified core algorithm implements a general
\hyperref[hetero-associative-memory]{hetero-associative} memory for
writing and retrieving associations \(A \to B\).
\hyperref[auto-associative-memory]{Auto-associative} behavior is
realized at the usage level by constraining data storage to
autoassociations \(A \to A\).

A hetero-associative memory \(M\) generally represents a mapping \(M: U
\to V\) from a universal set \(U\) to a universal set \(V\), which
differs from \(U\) in dimension and object type. We call \(U\) the
memory's input layer and \(V\) its output layer. The memory is
constructed incrementally by storing elementary associations
\(A \to B\), where \(A \subset U\) and \(B \subset V\) are small subsets
of their respective universes.

An \hyperref[auto-associative-memory]{auto-associative memory} maps a
universal set \(U\) onto itself. In this case, the memory's input and
output layers are shared, and the memory stores associations of the form
\(A \to A\). Here the underlying topological memory is configured with
identical hyperparameters for the input and the output. Implemented as a
functional network component, auto-associative memories do not provide
the secondary input channel (the \(B\) in \(A \to B\)) necessary for
storing heteroassociations.

The memory writing process has the following properties:

\begin{itemize}
\item
  Writing an association \(A \to B\) into memory is an atomic operation.
  Once new information is stored in memory, it is merged with existing
  memory traces and becomes immediately available for retrieval. Memory
  retrieval, also an atomic operation, can be interspersed with memory
  storage. This enables online learning.
\item
  Data storage is distributed and redundant --- the basis for the
  memory's noise and fault tolerance.
\item
  The memory state is an accumulation of all prior write operations.
  Individual write operations cannot be reconstructed from the aggregate
  memory state, and their order is not maintained.
\item
  The sizes (number of elements) of \(A\) and \(B\) may vary. When
  storing an association, the size of \(A\) or \(B\) may lie below the
  thresholds implied by the memory's
  \hyperref[hyperparameters]{hyperparameters}. Storing associations
  involving empty sets has no effect on the memory state. A software or
  hardware implementation should not treat this case as an error.
\item
  The core memory algorithm functions correctly for non-sparse inputs,
  but its performance and capacity strongly depend on sparsity.
\item
  Patterns written into memory may have any degree of overlap.
\item
  Writing the same association multiple times is the same as writing it
  once: the memory's storage model deduplicates redundant information by
  design.
\item
  For a subset \(A_1 \subset A\), storing both \(A_1 \to B\) and its
  extension \(A \to B\) subsumes the association for \(A_1\).
\item
  Storing new information increases the memory density, unless the
  memory is full. Writing data into a full memory has no effect.
\end{itemize}

Note that a stochastic \hyperref[memory-decay]{memory decay} mechanism,
if enabled, may change certain behaviors listed above.

Memory retrieval functions via subset pattern matching:

When queried with a set \(A \subset U\), the memory retrieves an
association \(X
\to Y\) by identifying the best partial match \(X \subseteq A\) and its
associated set \(Y \subset V\).

\begin{quote}
\textbf{Best match}\\
For a query pattern \(A\), the best match \(X \to Y\) is defined as the
unique largest subset \(X \subseteq A\) where every element pair within
\(X\) contributes to every element of \(Y\).
\end{quote}

Here are the main properties of this mechanism:

\begin{itemize}
\item
  The size of the matching patterns \(X\) and \(Y\) may vary, but must
  exceed the thresholds defined by the memory's
  \hyperref[hyperparameters]{hyperparameters}.
\item
  An implementation should give an empty result if there is no match
  meeting those requirements. This mechanism, based on overlap
  probabilities within the memory's storage capacity, serves as a test
  of whether or not an item is known.
\item
  If multiple subsets of \(A\) of the exact same size meet the matching
  criteria, a tie occurs and the retrieval algorithm defaults to an
  empty result. Such ties can be resolved by systematic stochastic
  subsampling of the query patterns.
\item
  Memory retrieval operates on all memory traces linked to element pairs
  within \(A\). Any prior write operation involving those element pairs
  may contribute to a successful match.
\item
  The memory merges information rather than keeping a record of atomic
  write operations. Therefore, the retrieval algorithm does not simply
  search for a match with a single association exactly as it was written
  into memory. Instead, it calculates a best match within the entire
  memory state, which aggregates all prior write operations.
\item
  As a discrete algorithm, memory retrieval can distinguish patterns
  that differ by just a single element, accurately finding the exact
  best match within known information.
\item
  Memory retrieval is a deterministic algorithm. Non-deterministic
  behaviors arise from randomized subsampling of query patterns.
\end{itemize}

Associative memory is generally content-addressable: memory retrieval
accesses matching data directly, rather than iterating through memory
records. Topological memory implements storage and retrieval of
associations \(A \to B\) with a time complexity of
\(\mathcal{O}(|A|^2)\), independent of the memory density.

\begin{quote}
\textbf{Constant-time operation}\\
In topological associative memory, the time complexity of data storage
and retrieval is independent of the total amount of stored data.
\end{quote}

\subsection{Storing associations}\label{storing-associations}

Storing an association \(A \to B\) into memory uses expansion coding:
each element pair within \(A\) activates a memory set. Each active
memory set is then updated with the union of \(B\) and its current
state:

\textbf{Input:} Sets \(A\) and \(B\)

\begin{enumerate}
\def\labelenumi{\arabic{enumi}.}
\item
  \textbf{Expansion coding:} Map 2-subsets \(\binom{A}{2}\) to memory
  sets \(M_{ij}\)
\item
  \textbf{Memory update:} \(M_{ij} \gets M_{ij} \cup B\)
\end{enumerate}

The time complexity of storing an association in memory is
\(\mathcal{O}(|A|^2)\), independent of the total memory density.

The dimensions of \(A\) and \(B\) must match the memory's
hyperparameters \(N_A\) and \(N_B\). There is no constraint on \(|A|\)
or \(|B|\) when writing into memory.

This distributed memory model incorporates two distinct levels of
redundancy: First, \(A\) and \(B\) are distributed representations on
their own. Second, \(B\) is stored equally in all active memory sets
linked to element pairs within \(A\), distributing the same information
redundantly over many memory locations.

In a typical low-level implementation, the memory sets \(M_{ij}\) are
coded as densely packed bitvectors, whereas sparse sets \(A\) and \(B\)
are best represented as variable-length lists. The memory requires
\(N_A(N_A-1)N_B/2\) binary storage locations.

The combinatorial addressing expansion, accounting for a factor
\((N_A-1)/2\), takes the memory size to the third order
\(\mathcal{O}(N_A^2 N_B)\) of the involved dimensions --- this is the
basis for the memory's high capacity.

\subsection{Retrieving associations}\label{retrieving-associations}

For an input set \(A\), memory retrieval isolates the association
\(X \to Y\) that best matches previously stored information: \(X\) is
the largest subset of \(A\) where every element pair contributes to
\(Y\).

The algorithm operates iteratively. It calculates the most likely output
\(Y\), identifies the elements of \(X\) that do not contribute to \(Y\),
and drops them. A new estimate for \(Y\) is recalculated in each cycle.
The process converges when all elements of \(X\) map consistently to
\(Y\), or fails if the remaining elements drop below the hyperparameter
threshold.

\textbf{Input:} Query set \(A\), hyperparameters \(T\) and \(P_B\)

\textbf{Output:} Unique best match \(Y\), or an empty set
\(\varnothing\)

\begin{enumerate}
\def\labelenumi{\arabic{enumi}.}
\item
  \textbf{Initialize:} \(X \gets A\)
\item
  \textbf{Threshold check:} If \(|X| < T\) return \(\varnothing\)
\item
  \textbf{Expansion coding:} Map 2-subsets \(\binom{X}{2}\) to memory
  sets \(M_{ij}\)
\item
  \textbf{Aggregate:}

  \(R \gets \sum_{ij} M_{ij}\) (overall response multiset)

  \(R_i \gets \sum_{j} M_{ij}\) (per-element contribution)
\item
  \textbf{Select}: \(Y \gets R_{\top P_B}\)
  (\hyperref[k-winners-take-all]{kWTA} with cutoff \(T(T-1)/2\))
\item
  \textbf{Threshold check:} If \(|Y| < P_B\) return \(\varnothing\)
\item
  \textbf{Per-element weights:} \(w_i \gets |R_i \cap Y|\)
\item
  \textbf{Convergence test:} If \(w_i \ge (|X| - 1) |Y|\) return \(Y\)
\item
  \textbf{Refine:} \(X \gets X \setminus X_\text{min}\) (drop weakest
  contributors)
\item
  \textbf{Iterate:} Go to step 2
\end{enumerate}

Note on pattern attribution: While the core algorithm returns the
retrieved pattern \(Y\), the matching subset \(X \subseteq A\) is
manifested internally. It can be exposed via an API function to support
an application's explainability.

Implementation details:

The time complexity of retrieval is \(\mathcal{O}(|A|^2)\), but is
independent of the total memory density. This satisfies the requirement
for a constant-time retrieval mechanism in high-performance intelligent
systems.

The threshold checks (based on hyperparameter choices) reliably separate
known from unknown inputs. The default value of \(T\) is chosen to
prevent false positive matches for a memory filled to capacity with
SHRs.

There are various ways to implement the refinement step, balancing
execution speed, memory efficiency, retrieval accuracy, and noise
tolerance. The version shown here (and coded in the reference library)
uses a heuristic based on subset combinatorics and operates with high
mathematical accuracy. Any biological realization of memory retrieval
will naturally be less accurate.

No iteration is needed if there is an immediate perfect match previously
stored information. If the input \(A\) includes non-contributing
elements, one or two iterations are normally sufficient.

\subsection{K-winners-take-all}\label{k-winners-take-all}

Managing activation sparsity is a fundamental requirement in both
biological neural networks and the computational model presented here.

For associative memory to function reliably, the sets representing
information tokens must maintain a stable and consistent number of
elements, with only small variance permitted. Population coding relies
on the principle that these sets are small relative to the dimension of
the universal set, a property that is the basis for compositionality.

In any neural circuit, every component must continuously regulate its
output sparsity to prevent runaway activity, which would otherwise lead
to a loss of pattern decomposition.

Framed in set-theoretic terms, this sparsity control reduces a multiset
\(M\) to a set \(A =  M_{\top K}\) by selecting the top \(K\) elements
based on their frequency. This mechanism is known as the
k-winners-take-all (kWTA) or K-rank algorithm. Note that in statistics,
\(K\) is the conventional notation for the parameter we define here as
the population \(P\).

\begin{quote}
\textbf{K-winners-take-all (kWTA)}\\
A sparsity-controlling algorithm that reduces a multiset to a discrete
set by selecting the top \(K\) most active elements.
\end{quote}

The \emph{hard} implementation of kWTA used here may return more than
\(K\) elements in the event of ties: if multiple elements share the same
value at the \(K\)-th rank, basic thresholding selects all of them.

An implementation with guaranteed \(\mathcal{O}(N)\) time complexity is
provided in the standard C source. Linear complexity is possible in our
specific case because the values to be ranked are discrete. Ranking
continuous values with methods such as \emph{quickselect} would be less
efficient.

kWTA occurs throughout this project: at the core of the associative
memory retrieval algorithm, in sparse coding of continuous-valued sensor
inputs, and as an analog electronic circuit deployed in a hardware
accelerator.

In ``many minds'' architectures, intelligence emerges not from a single
central model but from many independent agents voting for majority
consensus. Such models have been proposed, for example, by Marvin Minsky
(1986) and Jeff Hawkins (2021). Also, modern mixture-of-experts models
implement a version of this idea. In such architectures, individual
votes are combined via kWTA. This mechanism can produce Bayesian-like
outcomes, generating \emph{best guess} predictions.

The biological equivalent of kWTA is lateral inhibition, where active
neurons suppress their neighbors via inhibitory interneurons.

\subsection{Relation to hyperdimensional
computing}\label{relation-to-hyperdimensional-computing}

Set theory, together with subset combinatorics, provides a natural
mathematical framework for modeling neural population codes. This notion
of sets can be extended to hypergraphs to form a mathematical foundation
for knowledge representation.

The mathematics of small sets is closely related to hyperdimensional
computing theory, which strongly influenced the development of this
project. This theory models information as binary hypervectors,
equivalent to points in a high-dimensional space. These binary
hypervectors are, of course, equivalent to the ``characteristic
vectors'' of sets. The hypervector dimensionality \(N\) corresponds to
the size of the universal set, and a hypervector's Hamming weight (the
number of active bits) aligns with a specific set's size.

To benefit from the ``blessing of dimensionality'', such as the
mathematical tendency toward orthogonality, the vector dimensionality
\(N\) is typically very large, in a range from 1,000 to about 10,000. A
detailed discussion of this framework can be found in Kanerva's 2009
overview of hyperdimensional computing.

In the context of hyperdimensional computing, storing an association
physically links the binary vectors of two concepts inside a
three-dimensional grid. The memory records this relationship by
switching the specific grid coordinates where the individual components
of both concepts intersect to 1.

Formally, writing \(A \to B\) to a memory \(T_{ijk}\) (a rank-3 tensor)
can be expressed by \(T_{ijk} \gets T_{ijk} \lor (A_i A_j B_k)\) for all
\(i < j\) and all \(k\).

The core function of memory retrieval from an input \(A\) is equivalent
to \(B_k = ( \sum_{i<j} T_{ijk} A_i A_j )_{\top P_B}\).

The geometrical interpretation of this structure is a triangular prism:
a cuboid with dimensions \(N_A \times N_A \times N_B\) cut diagonally in
half, with the diagonal removed. This prism contains
\(N_A (N_A - 1)N_B/2\) addressable bits. Memory is not stored in a
two-dimensional matrix, as one may think, but in this three-dimensional
structure.

\section{Memory Properties}\label{memory-properties}

\subsection{Hyperparameters}\label{hyperparameters}

A topological memory unit stores associations \(A \to B\) with fixed
dimensions, governed by the sizes \(N_A\) and \(N_B\) of the underlying
network's input and output layers.

In set-theoretic terms, \(A\) and \(B\) are simply subsets drawn from
universal sets \(U\) and \(V\) with cardinalities \(N_A\) and \(N_B\),
respectively.

The populations (sizes) of \(A\) and \(B\) may vary within the same
memory unit. Retrieval via subset matching requires a minimum number of
elements in \(A\) and \(B\) to match with previously stored patterns.
The choice of these thresholds depends on the meaning and purpose of the
representations to be processed: symbolic tokens, their bundles,
pointers, segmented data structures, numeric codes, grid cell encodings,
or overlapping sensory information.

The parametrization chosen here is most convenient and intuitive for
working with sparse holographic representations (SHRs) of uniform
sparsity.

{\def\LTcaptype{none} 
\begin{longtable}[]{@{}
  >{\centering\arraybackslash}p{(\linewidth - 4\tabcolsep) * \real{0.2286}}
  >{\raggedright\arraybackslash}p{(\linewidth - 4\tabcolsep) * \real{0.3286}}
  >{\raggedright\arraybackslash}p{(\linewidth - 4\tabcolsep) * \real{0.4429}}@{}}
\toprule\noalign{}
\begin{minipage}[b]{\linewidth}\centering
\textbf{Hyperparameter}
\end{minipage} & \begin{minipage}[b]{\linewidth}\raggedright
\textbf{Meaning}
\end{minipage} & \begin{minipage}[b]{\linewidth}\raggedright
\textbf{Usage}
\end{minipage} \\
\midrule\noalign{}
\endhead
\bottomrule\noalign{}
\endlastfoot
\(N_A\) & input dimension & \\
\(P_A\) & input population & \\
\(N_B\) & output dimension & \(N_B = N_A\) if auto-associative \\
\(P_B\) & output matching threshold & \(P_B = P_A\) if
auto-associative \\
\(T\) & input matching threshold & computed automatically or set
explicitly \\
\end{longtable}
}

Auto-associative memory is realized by a configuration with
\(N_B = N_A\) and \(P_B = P_A\).

A software or hardware implementation of topological associative memory
should allow arbitrary-sized patterns to be stored into memory. In
particular, it should not treat storing an association involving an
empty set as an error. For a write operation to affect the memory state,
\(A\) must have at least two elements, and \(B\) must be non-empty.

In memory retrieval, a successful match must include at least \(T\)
elements of \(A\) and \(P_B\) elements of \(B\).

By default, the pattern matching threshold \(T\) is automatically set to
a value that minimizes false-positive matches. This is based on the
memory's capacity to store random associations (SHRs) with uniform
populations \(P_A\) and \(P_B\). Computing \(T\) involves the full set
of hyperparameters.

The automatically computed value of \(T\) is optimized for a memory that
stores SHRs. This default can be overridden, which is often necessary
when operating on SDRs. The pattern matching threshold is a global
parameter within the memory instance; it cannot be adjusted on a
per-query basis. If an explicit value is specified for \(T\), the value
of \(P_A\) has no effect on the memory's operation: either \(P_A\) or an
explicit value of \(T\) governs the pattern matching logic, but not both
at the same time.

Apart from these hyperparameters, the memory algorithm contains no other
configuration options --- it is neither necessary nor possible to
fine-tune internal parameters for a specific application scenario.

\subsection{Memory capacity}\label{memory-capacity}

Topological memory can store and perfectly retrieve a very large number
of associations \(A \to B\) between sets \(A\) and \(B\). The capacity
depends on the memory's \hyperref[hyperparameters]{hyperparameters} and
the statistical overlap among the sets \(A\).

We define the memory's nominal capacity as the number of random
heteroassociations that can be stored until half the storage locations
are set to 1. This is the point of maximum information density, because
the capacity is symmetric with respect to bit-flipping.

\begin{quote}
\textbf{Nominal memory capacity} The maximum number of random
associations a topological memory can store until half of its storage
locations are set to 1 (density \(\rho^* = 0.5\)).
\end{quote}

Storing one association sets \(P_A (P_A-1)P_B/2\) storage bits to 1,
regardless of their current state. As the memory density increases, the
probability of write operations flipping bits from 0 to 1 decreases.

Given the total memory size \(N_A(N_A-1)N_B/2\), we can quickly
calculate the expected number of write operations necessary to fill half
the memory:

\[C_\text{hetero} \approx \ln(2) \frac{ N_A (N_A-1)  N_B} { P_A (P_A-1) P_B }\]

This formula shows that the sparsity of \(A\) has a stronger impact on
capacity than that of \(B\), primarily due to the combinatorial
expansion occurring from the input to the hidden layer.

The following plot shows the relationship between memory density and the
number of stored SHRs. A configuration with \(N = 1,000\) and \(P = 10\)
has a nominal capacity of around 769,393 heteroassociations at density
\(\rho^* = 0.5\).

\begin{figure}[H]
\centering
\includegraphics[width=0.675\linewidth,height=\textheight,keepaspectratio,alt={SHR storage capacity as a function of memory density \textbackslash rho for hyperparameters N = 1,000 and P = 10.}]{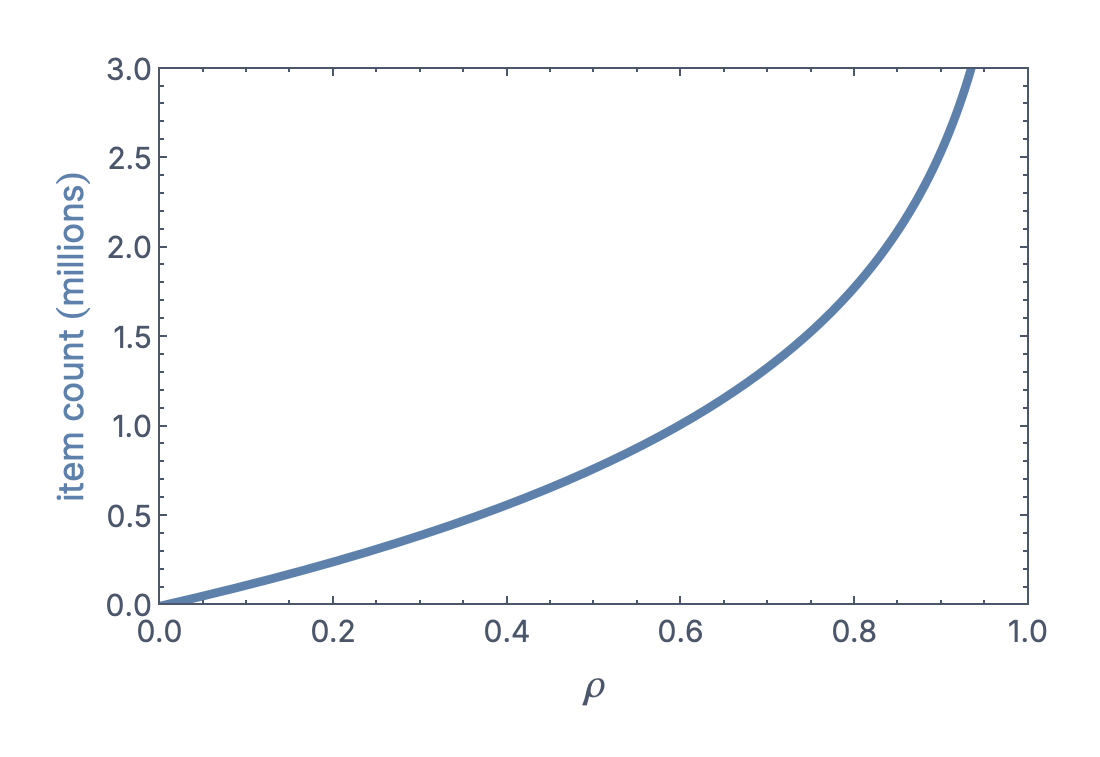}
\caption{SHR storage capacity as a function of memory density \(\rho\)
for hyperparameters \(N = 1,000\) and \(P = 10\).}
\end{figure}

Two complementary metrics quantify retrieval accuracy: the average
overlap between stored and retrieved sets (number of correctly matched
elements) and their average Hamming distance \(\text{H}\) (number of
disagreeing elements). Both metrics are calculated at the output layer
(the right-hand side of associations \(A \to B\)). Perfectly accurate
retrieval implies \(\text{H} = 0\) and full overlap.

In a memory filled with random associations at uniform population,
stored and retrieved patterns fully overlap regardless of the memory
density. But the average Hamming distance turns non-zero from a density
of around 0.85 and then grows like a hockey stick: memory recall picks
up spurious elements in addition to the correct ones due to interference
with unrelated information:

\begin{figure}[H]
\centering
\includegraphics[width=0.675\linewidth,height=\textheight,keepaspectratio,alt={Retrieval accuracy as a function of memory density \textbackslash rho.}]{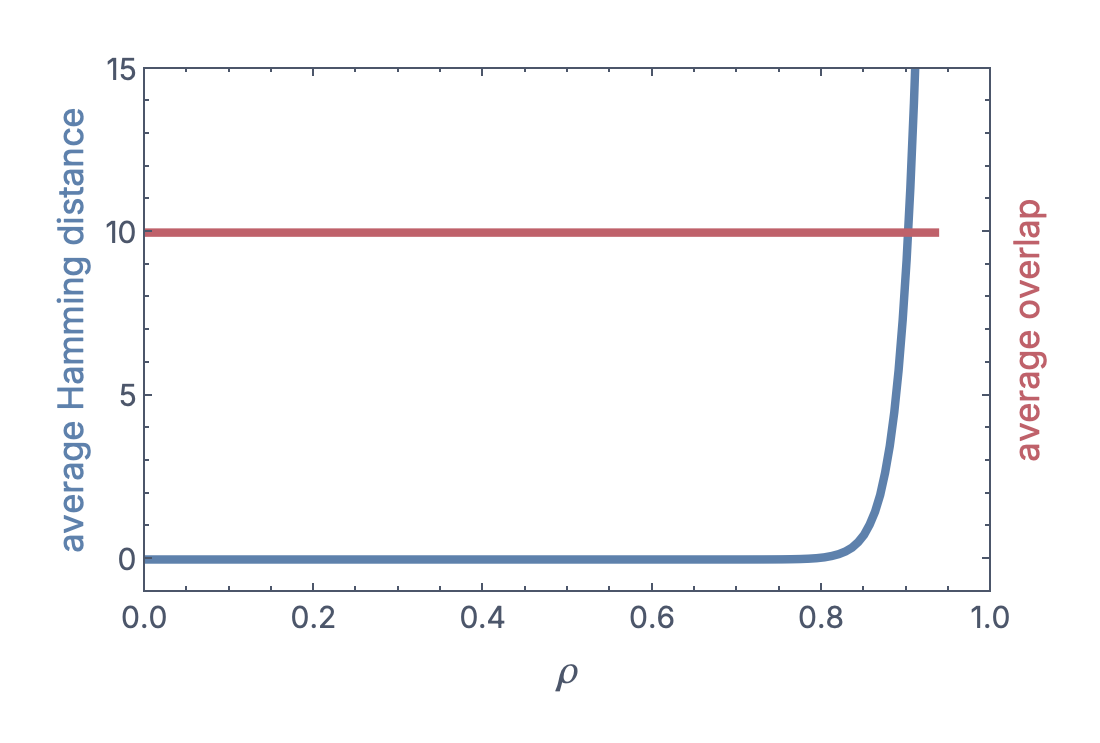}
\caption{Retrieval accuracy as a function of memory density \(\rho\).}
\end{figure}

The \emph{memory cliff} phenomenon occurs well beyond the nominal
capacity mark \(\rho^* = 0.5\).

Auto-associative memory has a higher capacity than hetero-associative
memory with equivalent hyperparameters. Recall that topological memory
assigns a memory set to each pair of input elements. Each element of the
memory set represents a topological connection to the corresponding
output element. But with autoassociations, the input and the output are
identical: the memory set corresponding to elements \((i,j)\) always
includes \(i\) and \(j\). In each memory set, two elements are
``clamped''. This means that only \(N-2\) output dimensions carry
information (a negligible difference), and \(P-2\) instead of \(P\)
elements of the memory sets are affected when storing an autoassociation
(a noticeable difference).

The nominal capacity for auto-associative memory storing random SHRs
with dimension \(N\) and population \(P\) is given by

\[C_\text{auto} \approx \ln(2) \frac{ N(N-1)(N-2)} {P(P-1)(P-2)}\]

We conclude that the capacity for storing autoassociations is about
\(P/(P-2)\) times larger than that for heteroassociations with the same
hyperparameters.

Auto-associative memory with \(N = 1,000\) and \(P = 10\) can store and
accurately recall about 959,818 random patterns.

For a fixed population size, memory capacity scales rapidly with the
dimensionality \(N\). A configuration with 5,000 dimensions can store
approximately 100 million associations.

\begin{figure}[H]
\centering
\includegraphics[width=0.675\linewidth,height=\textheight,keepaspectratio,alt={Memory capacity to store associations with SHR population P=10, as a function of dimension N.}]{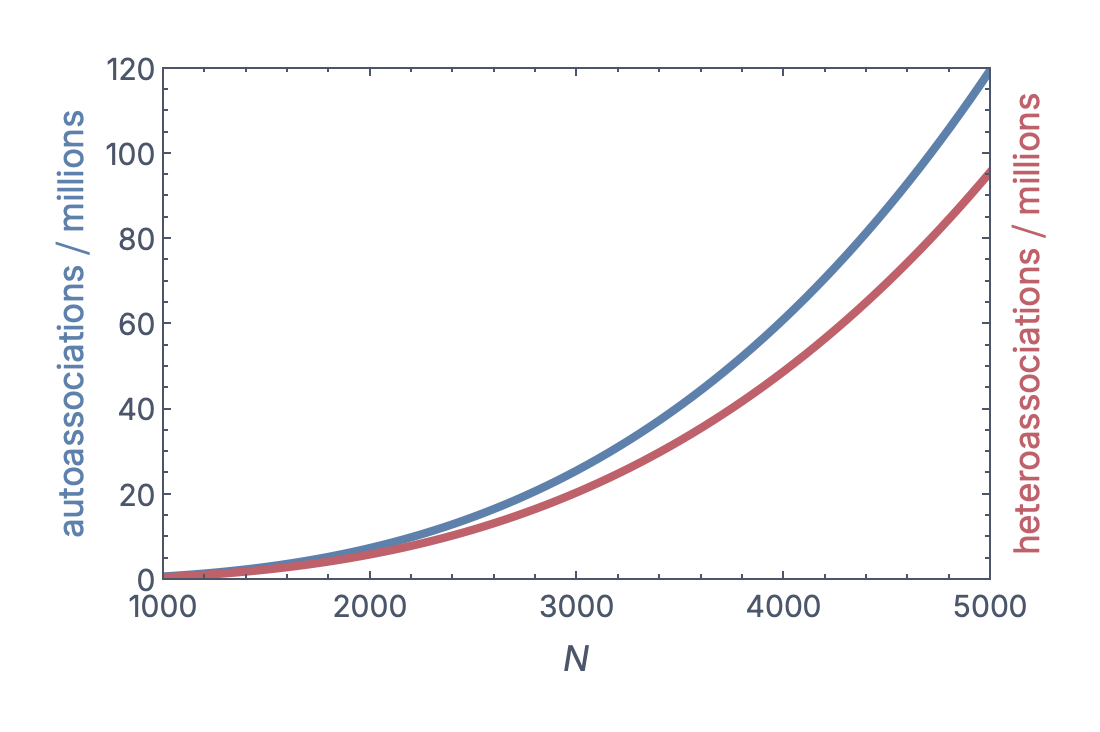}
\caption{Memory capacity to store associations with SHR population
\(P=10\), as a function of dimension \(N\).}
\end{figure}

\subsection{Storage efficiency}\label{storage-efficiency}

A notable property of topological memory is that the physical storage
cost per item is independent of the overarching dimensionality. We can
derive the number of bits required to store one association by dividing
the total number of storage locations \(V\) by the nominal capacity
\(C\). The capacity \(C\) is given by

\[C \approx \ln(2) \frac{V}{L}\]

where \(L\) is the number of bits affected by one write operation. With
this approximation for the capacity, the storage cost is simply:

\[\frac{V}{C} \approx \frac{L}{\ln(2)}\]

In this expression, the total memory size \(V\) and therefore the
memory's dimensionality cancel out.

The cost of storing a heteroassociation is then given by:

\[\left(\frac{V}{C}\right)_\text{hetero} \approx \frac{P_A(P_A-1)P_B}{\ln(4)}\]

The number of bits required to store an autoassociation:

\[\left(\frac{V}{C}\right)_\text{auto} \approx \frac{P(P-1)(P-2)}{\ln(4)}\]

In a configuration with \(P = 10\), one autoassociation occupies 519
bits, regardless of the dimensionality. Considering the high redundancy
of data storage, this is remarkably low --- especially when compared
with a dense hypervector representation, occupying \(N\) bits.

\subsection{Noise tolerance}\label{noise-tolerance}

Topological associative memory is fault-tolerant: information can be
accurately retrieved from mere subsets of originally stored patterns. In
an auto-associative context, this mechanism drives pattern completion.

For a memory configuration with \(N = 1,000\) and \(P = 10\), the
default pattern matching threshold is \(T = 7\). Retrieval succeeds even
if three random elements are missing, but input reduction increases
susceptibility to interference. This shifts the memory cliff to a lower
memory density:

\begin{figure}[H]
\centering
\includegraphics[width=0.675\linewidth,height=\textheight,keepaspectratio,alt={Retrieval accuracy with 3 random elements dropped, as a function of memory density \textbackslash rho.}]{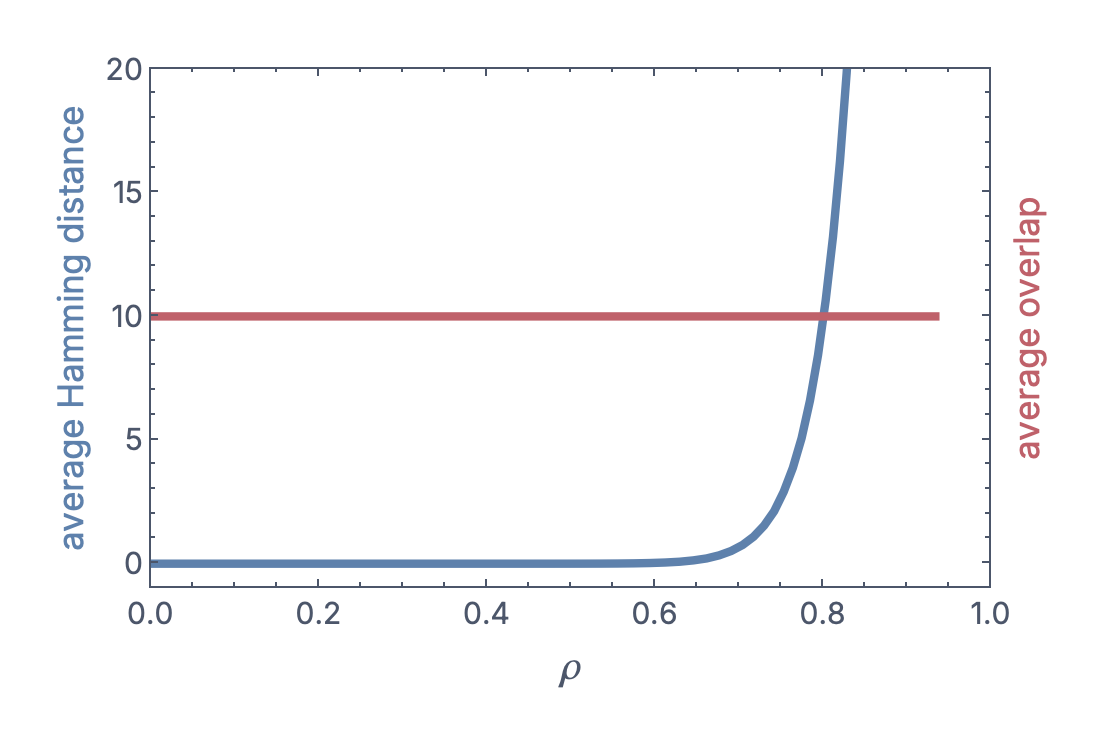}
\caption{Retrieval accuracy with 3 random elements dropped, as a
function of memory density \(\rho\).}
\end{figure}

Beyond handling missing data, the algorithm uses subset pattern matching
to extract known information from inputs polluted with additive noise.
It filters out elements that are not part of a known pattern. This is
the basis for pattern decomposition --- an essential cognitive process.
It is also the exact mechanism that decomposes unions (bundles) of sets
back into their constituent tokens.

This plot illustrates the retrieval accuracy for patterns with
\(P = 10\) from inputs with 15 additional random elements:

\begin{figure}[H]
\centering
\includegraphics[width=0.675\linewidth,height=\textheight,keepaspectratio,alt={Retrieval accuracy with 15 random elements added, as a function of memory density \textbackslash rho.}]{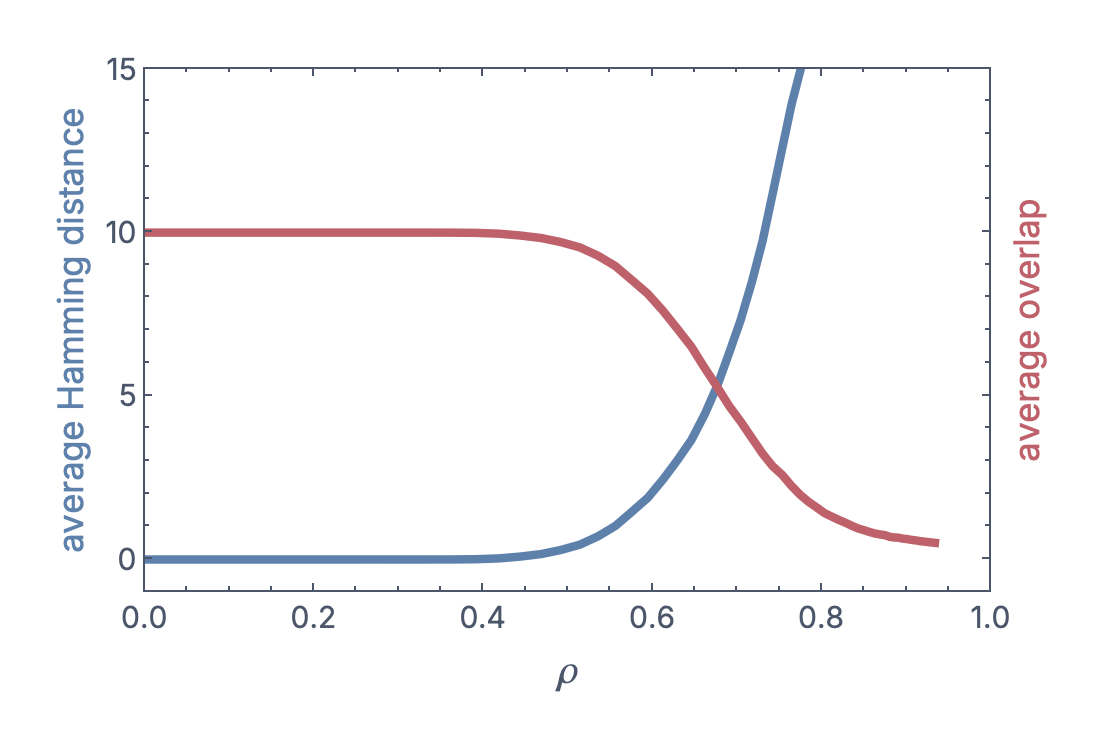}
\caption{Retrieval accuracy with 15 random elements added, as a function
of memory density \(\rho\).}
\end{figure}

The average overlap decreases as the memory density increases. Examining
the output of individual queries, we observe that most outcomes are
either perfect or empty. Zero results occur when there is a tie between
equally valid matches. In practical applications, queries will be
repeated a few times with subsampled versions of the input, until a
unique best match is found: the decrease of overlap as shown in the
figure above correlates with an increasing number of randomized queries,
necessary for a partial match to win.

\subsection{Pattern matching
threshold}\label{pattern-matching-threshold}

The memory retrieval algorithm attempts to find the best possible match
of an input set \(A\) with previously stored associations. To avoid
false positives (hallucinations), a minimum number \(T\) of elements
must coincide. This pattern matching threshold is governed by the
overlap probability between a pair of random sets and by the memory's
nominal capacity \(C\).

If we pick two random sets of size \(P\) from a universe of size \(N\),
the probability that they share exactly \(t\) elements is given by the
hypergeometric distribution:

\[p(t) = \binom{P}{t} \binom{N - P}{P - t} \binom{N}{P}^{-1}\]

If the memory is filled to its nominal capacity, it contains \(C\)
distinct SHRs. If we query the memory with a random, unknown input
pattern, the probability that this input falsely matches any one of the
stored entries by at least \(T\) elements is bounded by \(C \cdot p(T)\)
(using the union bound and approximating the cumulative probability by
its dominant term \(p(T)\)).

To derive a suitable threshold \(T\), we mandate the following
condition: if each of the \(C\) entries stored in a full memory is
probed with an individual random pattern, the expected number of
false-positive matches must remain below 1. This is expressed by the
inequality

\[C^2 \cdot p(T) < 1\]

Substituting the hypergeometric probability for \(p(T)\) and the nominal
capacity formula for \(C_\text{hetero}\), we arrive at the threshold
condition for heteroassociations:

\[\left( \ln(2) \frac{N_A (N_A - 1) N_B}{P_A (P_A - 1) P_B}\right)^2 \cdot \binom{P_A}{T} \binom{N_A - P_A}{P_A - T} \binom{N_A}{P_A}^{-1} < 1\]

This requirement can be fulfilled only for inputs with populations \(P_A
\ge 7\). For smaller populations, we set \(T = P_A\), which means all
elements of the input pattern must match.

The following table shows the pattern matching threshold \(T\) for
various hyperparameter configurations \(N = N_A = N_B\) and
\(P = P_A = P_B\):

\begin{verbatim}
     P = 7  8  9  10  11  12  13  14  15  16  17  18  19  20  21  22  23  24
N =     
 1k      6  7  7   7   7   7   8   8   8   8   8   9   9   9   9   9   9  10
 2k      6  7  7   7   7   7   7   8   8   8   8   8   8   9   9   9   9   9
 3k      6  7  7   7   7   7   7   8   8   8   8   8   8   8   9   9   9   9
 4k      6  7  7   7   7   7   7   8   8   8   8   8   8   8   8   9   9   9
 5k      6  7  7   7   7   7   7   7   8   8   8   8   8   8   8   8   9   9
 6k      6  7  7   7   7   7   7   7   8   8   8   8   8   8   8   8   8   9
 7k      6  7  7   7   7   7   7   7   8   8   8   8   8   8   8   8   8   9
 8k      6  7  7   7   7   7   7   7   8   8   8   8   8   8   8   8   8   8
 9k      6  7  7   7   7   7   7   7   8   8   8   8   8   8   8   8   8   8
10k      6  7  7   7   7   7   7   7   8   8   8   8   8   8   8   8   8   8
11k      6  7  7   7   7   7   7   7   7   8   8   8   8   8   8   8   8   8
12k      6  7  7   7   7   7   7   7   7   8   8   8   8   8   8   8   8   8
\end{verbatim}

For auto-associative memory, the threshold condition is given by

\[\left( \ln(2) \frac{N (N - 1) (N - 2)}{P (P - 1) (P - 2)}\right)^2 \cdot \binom{P}{T} \binom{N - P}{P - T} \binom{N}{P}^{-1} < 1\]

\subsection{Separating known from
unknown}\label{separating-known-from-unknown}

Classical associative memory algorithms, like Hopfield nets, are
designed to \emph{always recall}. They do this by converging to a stable
state known as an attractor. This attractor might be the closest stored
pattern, or it could just as easily be a spurious state hallucinated
from unrelated memories.

Topological associative memory works differently. It only converges if a
statistically significant portion of the input overlaps with known
information: only known patterns function as attractors in this model.
The memory's ability to separate known from unknown patterns is one of
its defining features.

This filtering mechanism works accurately up to the nominal memory
capacity at density \(\rho^* = 0.5\). Above this mark, the probability
of false-positive retrieval begins to increase. This is intuitively
clear: with a hypothetical full memory, any random input matches.

Here is a simulation of this behavior, measuring the probability of
non-empty responses to random query patterns.

\begin{figure}[H]
\centering
\includegraphics[width=0.675\linewidth,height=\textheight,keepaspectratio,alt={Probability of false-positive responses to random queries, as a function of memory density \textbackslash rho. This simulation uses heteroassociations with N=1,000 and P=10.}]{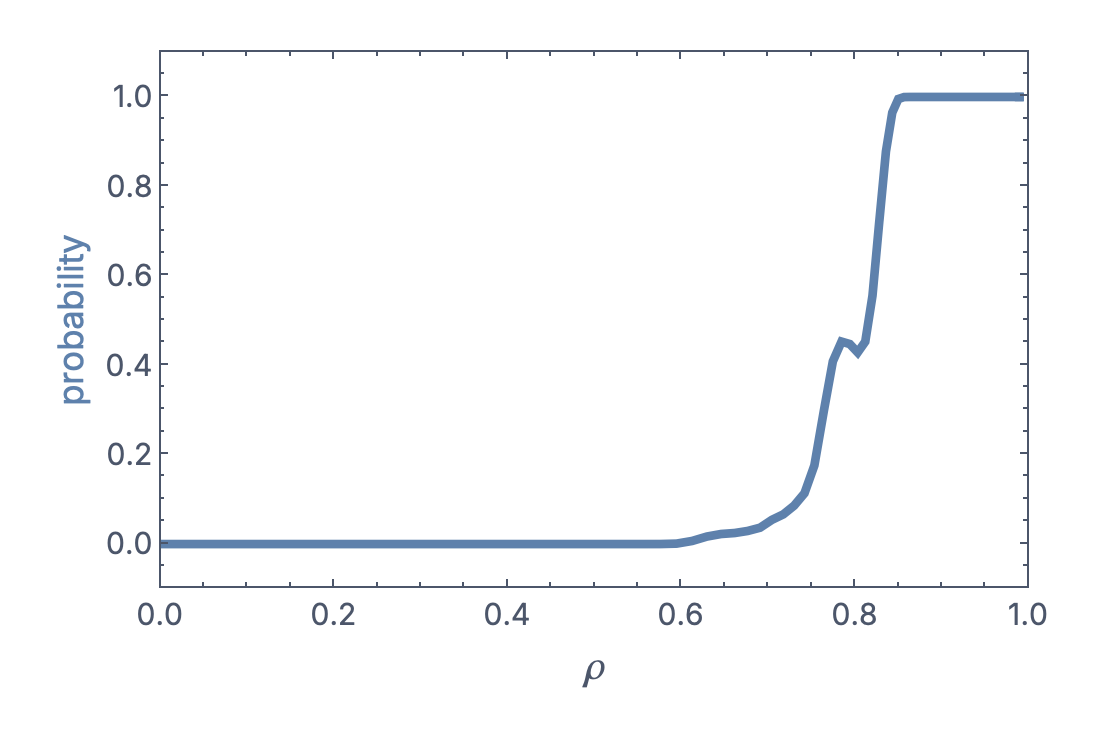}
\caption{Probability of false-positive responses to random queries, as a
function of memory density \(\rho\). This simulation uses
heteroassociations with \(N=1,000\) and \(P=10\).}
\end{figure}

\section{Computing With Sets}\label{computing-with-sets}

In our computational framework, sets act as the foundational data
structure --- specifically, small subsets drawn from hyperdimensional
universes. The core algorithm is topological memory, which retrieves
information via subset pattern matching. The combination of set-based
composite data structures and set-manipulating algorithms enables a huge
space of possible programs.

To map this computational space, we begin by cataloging elementary data
types and operations. This is the starting point for developing simple
design patterns and cognitive architectures.

Here are the most common set operations and their notations:

{\def\LTcaptype{none} 
\begin{longtable}[]{@{}
  >{\raggedright\arraybackslash}p{(\linewidth - 6\tabcolsep) * \real{0.2523}}
  >{\raggedright\arraybackslash}p{(\linewidth - 6\tabcolsep) * \real{0.2793}}
  >{\centering\arraybackslash}p{(\linewidth - 6\tabcolsep) * \real{0.1441}}
  >{\raggedright\arraybackslash}p{(\linewidth - 6\tabcolsep) * \real{0.3243}}@{}}
\toprule\noalign{}
\begin{minipage}[b]{\linewidth}\raggedright
\textbf{Function}
\end{minipage} & \begin{minipage}[b]{\linewidth}\raggedright
\textbf{Set Operation}
\end{minipage} & \begin{minipage}[b]{\linewidth}\centering
\textbf{Notation}
\end{minipage} & \begin{minipage}[b]{\linewidth}\raggedright
\textbf{Neural Basis}
\end{minipage} \\
\midrule\noalign{}
\endhead
\bottomrule\noalign{}
\endlastfoot
bundling & union & \(A \cup B\) & temporal integration \\
block coding & disjoint union & \((A,B)\) & pathway integration \\
novelty detection & complement & \(A \setminus B\) & subtractive
inhibition \\
matching & intersection & \(A \cap B\) & coincidence detection \\
prediction & symmetric difference & \(A \Delta B\) & recurrent
inhibition \\
similarity & overlap metric & \(\vert A\cap B\vert\) & coincidence
detection \\
majority voting & multisets & \(A \uplus B\) & linear aggregation \\
kWTA & thresholding & \(u_{\top K}\) & lateral inhibition \\
capping & subsampling & \(A_k\) & stochastic inhibition \\
decimation & subsampling & \(D \cdot A\) & stochastic inhibition \\
permutation & bijection & \(\pi A\) & recurrent pathway \\
binding & autoassociation & \(A \otimes B\) & topological memory \\
replacement & heteroassociation & \(A \to B\) & feed-forward memory \\
\end{longtable}
}

\subsection{Bundling}\label{bundling}

Bundling, also known as superposition, combines multiple representations
into a single representation. Together with \hyperref[binding]{binding}
and \hyperref[permutation]{permutation}, bundling is one of the basic
algebraic operations used to compose and manipulate distributed
representations.

A bundle represents a set of distributed representations: either SDRs
encoding real-world data, or SHRs representing symbolic tokens

With objects encoded as sets \(A\) and \(B\), their bundle is given by
the set union operation \(A \cup B\). In the context of hyperdimensional
computing, this is equivalent to the logical \(\text{OR}\) between the
respective characteristic vectors.

An object is either part of a bundle, or it is not. It cannot occur more
than once within a bundle. This is expressed by the identity
\(A \cup A = A\). The bundling operation is commutative:
\(A \cup B = B \cup A\). Bundles represent unordered collections. To
represent ordering, we need to use either \hyperref[block-coding]{block
coding} or \hyperref[permutation]{permutations}.

Associativity: the grouping of sets during the bundling process does not
affect the resulting representation. \((A \cup B) \cup C\) produces the
same set as \(A \cup (B \cup C)\). Similarly, because the operation is
commutative, the order of sets in the bundle is also irrelevant. This
allows us to build up bundles incrementally, in any order.

The bundling operation preserves similarity: Both \(A\) and \(B\) are
similar to \(A
\cup B\), as measured by their overlaps. In comparison, the
\hyperref[binding]{binding} \(A \otimes B\) produces a dissimilar
representation.

\begin{quote}
\textbf{Bundling}\\
Combining multiple distributed representations via the set union
operation to create an unordered collection that remains semantically
similar to its parts.
\end{quote}

Bundling two representations results in a less sparse representation.
For any two sets \(A\) and \(B\), the cardinality of their union is
bounded by the following inequalities:

\[\begin{aligned}
|A| &\le |A \cup B| \le |A| + |B| \\
|B| &\le |A \cup B| \le |A| + |B|
\end{aligned}\]

There is no hard limit on how many SHRs we can bundle before losing
sparsity. But topological memory imposes a practical limit that we can
measure experimentally.

First, we calculate the expected size of a bundle composed of \(k\)
random SHRs with population \(P\) and dimension \(N\):

\[\texttt{bundlesize}(N, P, k) = N \left(1 - \left(1 - \frac{P}{N} \right)^k \right )\]

The more items we add to a bundle, the higher the probability of overlap
becomes. The bundle eventually converges to the universal set when
saturated with all possible elements:

\begin{figure}[H]
\centering
\includegraphics[width=0.675\linewidth,height=\textheight,keepaspectratio,alt={Expected size of a superposition as a function of the number of bundled SHRs, assuming N = 1,000 and P = 10.}]{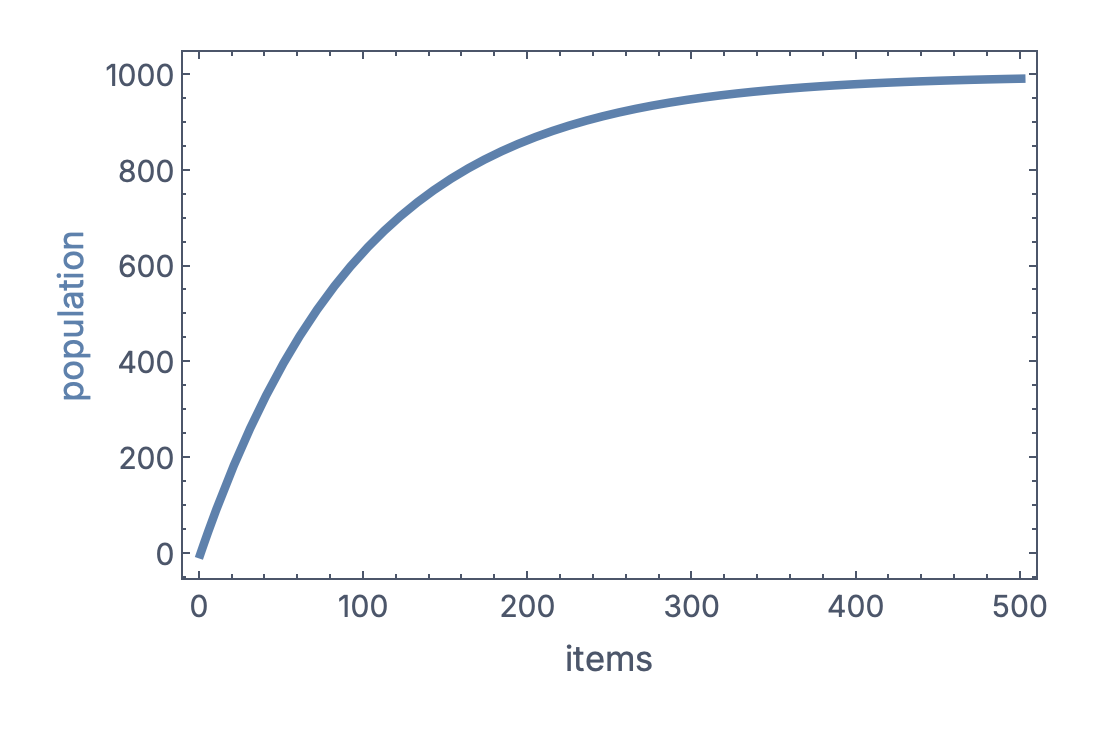}
\caption{Expected size of a superposition as a function of the number of
bundled SHRs, assuming \(N = 1,000\) and \(P = 10\).}
\end{figure}

As the bundle grows, its subsets increasingly resemble unrelated
patterns. This phenomenon (crosstalk) can be quantified as follows:

Consider an item memory filled to capacity with SHRs. We calculate the
probability that a bundle of \(k\) SHRs overlaps with any stored item in
memory by \(T\) elements (the memory's pattern matching threshold). Such
overlap causes false-positive matches during memory retrieval. We
require the probability of a false-positive outcome to remain below 1 to
derive a lower bound on the usable bundle size. For a fixed SHR
population, this limit is a function of the dimension as shown in the
following figure.

\begin{figure}[H]
\centering
\includegraphics[width=0.675\linewidth,height=\textheight,keepaspectratio,alt={Usable bundle size at which false-positive matches are unlikely, as a function of dimensionality N. This simulation uses a fixed SHR population P=10.}]{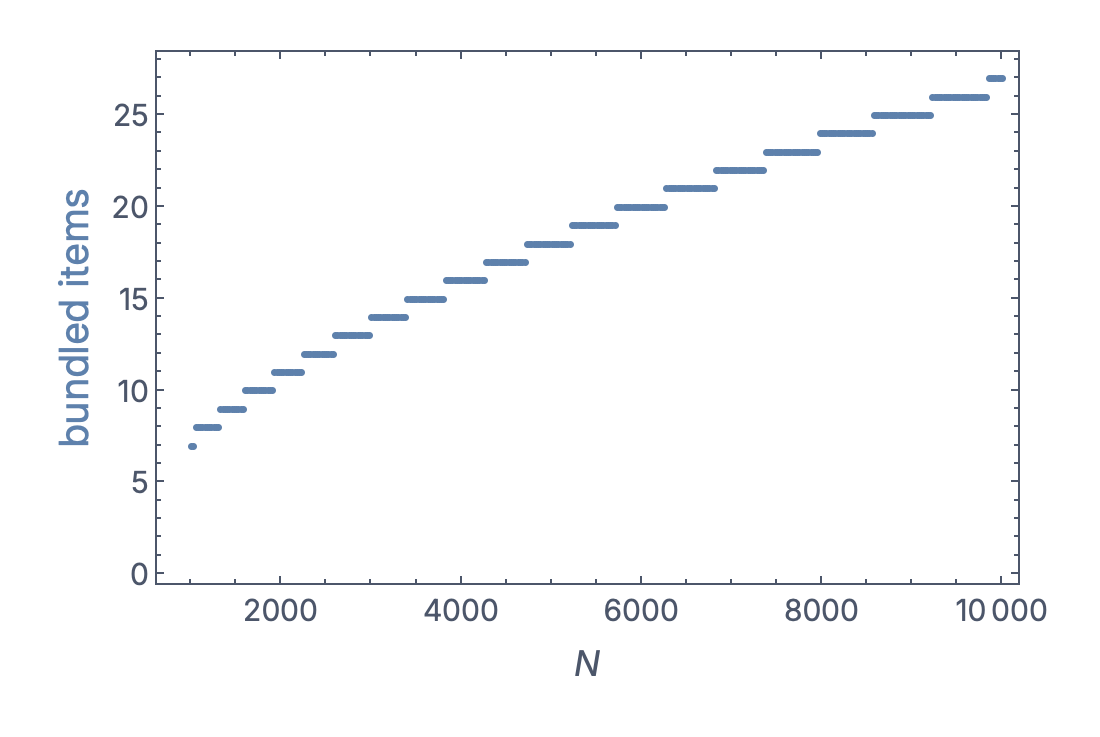}
\caption{Usable bundle size at which false-positive matches are
unlikely, as a function of dimensionality \(N\). This simulation uses a
fixed SHR population \(P=10\).}
\end{figure}

The usable bundle size is around 10 for small memory configurations.
This aligns with the structural limitations of decomposing bundles using
auto-associative memory, discussed in detail further
\hyperref[iterative-decomposition]{below}.

In a biological network, bundles can form dynamically without being
persisted in memory. This happens through temporal signal integration,
within recurrent networks, or as the sparse outcome of
\hyperref[k-winners-take-all]{majority voting} among multiple sets.
Additionally, bundles can emerge during memory retrieval when inputs map
to multiple outputs simultaneously.

Bundles can be naturally persisted in auto-associative memory. A memory
filled with superpositions of symbolic tokens (SHRs) functions as a
hypergraph database.

\subsection{Block coding}\label{block-coding}

Disjoint bundling, a variant of the bundling operation discussed above,
combines multiple disjoint sets into a single, partitioned set. Two sets
are called disjoint if they are non-empty and share no elements. This
forms the set-theoretic foundation for block coding.

We can combine distinct information categories, for instance multimodal
sensor data, into a larger set where each block represents a
non-overlapping feature space or modality. Conversely, we may partition
larger sets into smaller blocks, where each block can represent a
specific role or attribute. This is especially useful for building
graphs where representations from the same domain are arranged in
non-overlapping blocks.

\begin{quote}
\textbf{Block coding}\\
Combining non-overlapping sets via the disjoint union operation.
\end{quote}

Technically the union of disjoint sets involves re-indexing
(renumbering) their elements, as a way to tag their origin.

In standard set theory, the elements of a partitioned set all belong to
the same universe. But there is also a \emph{multiverse} view where each
partition represents its own universe. The concept of a disjoint union
is related to the \emph{coproduct} in category theory, and \emph{sum} or
\emph{sum-set} in type theory. But regardless of the mathematical
formalism, we use partitioned sets to bundle different types of
representations that should not be superimposed with the regular
bundling mechanism. For example, a hypergraph representation might keep
edge labels and a bundle of nodes in two disjoint set partitions.

This is equivalent to the disjoint union, usually denoted as
\(A \sqcup B\), which we represent as the tuple \((A, B)\)

Disjoint bundling is not a commutative operation: \((A,B)\) and
\((B,A)\) represent different sets.

The partitions \(A\) and \(B\) may have different dimensions (the
\emph{multiverse} view). The dimension \(\vert(A,B)\vert\) is the sum of
the individual dimensions \(N_A + N_B\).

Topological associative memory, of course, handles partitioned sets just
like any other set. From a computational point of view, we just need to
properly re-index the elements with a contiguous range of numbers. This
process is structurally equivalent to concatenating the sets'
characteristic vectors.

Notably, auto-associative memory preserves the partitioned structure.
Storing an association \((A,B) \to (A,B)\) in memory treats the inputs
as one contiguous set, linking information across the partitions.
Querying the memory with a partially filled structure such as
\((A, \varnothing)\) infers the missing information from memory,
completing the pattern.

Block coding is normally applied with a small number of partitions,
where each partition can be viewed as its own universe of features. For
instance, the ``\hyperref[frames]{frames}'' model arranges cognitive
relations in three layers. And of course, semantic
\hyperref[triplestore]{triples}, entity-attribute-value models, and
related schemes are founded on tripartite representations.

The neural basis for disjoint bundles is the joining of distinct
pathways, meaning that this mechanism is hardwired into the network
architecture. Therefore, we should not use block coding for storing
sequences.

\subsection{Multisets}\label{multisets}

Multiset aggregation, a mechanism closely related to
\hyperref[bundling]{bundling}, refers to combining multiple sets into a
multiset (an unordered collection of elements, or \emph{bag}). Unlike
bundling, which results in a set of unique elements, multiset
aggregation preserves not only membership but also the multiplicity of
duplicate elements.

Like bundling, multiset aggregation preserves the similarity with its
components. But depending on the chosen metric, multiset similarity (or
distance) may be weighted by the elements' multiplicities.

We use the notation \(u = A \uplus B\) for a multiset \(u\) formed by
aggregating sets \(A\) and \(B\). Multiset aggregation is fundamentally
commutative, meaning that \(A
\uplus B\) and \(B \uplus A\) result in the same representation. It is
also fully associative:
\((A \uplus B) \uplus C = A \uplus (B \uplus C)\). Note that in common
hyperdimensional models, multiset aggregation is only approximately
associative due to clipping and binarization; consequently, the order of
operations can influence the result.

At the cellular level, neurons aggregate information through the spatial
and temporal summation of postsynaptic potentials. This is modeled as a
multiset where the multiplicity of elements corresponds to linear
summation or dendritic summation of signals.

The \hyperref[k-winners-take-all]{kWTA} mechanism acts as the cleanup or
normalization layer for multiset aggregation, mimicking lateral
inhibition in the brain. It enforces sparsity by allowing only the \(K\)
most active neurons to fire. This mirrors biological inhibition where
groups of neurons compete to be active. As discussed above, this
mechanism (together with combinatorial expansion of a hidden layer) is
the basis of topological associative memory.

\subsection{Stochastic subsampling}\label{stochastic-subsampling}

Managing the sparsity of activation is a prerequisite for associative
memory to properly function. For an input set with \(P\) elements,
topological memory activates \(\mathcal{O}(P^2)\) latent units:
exceeding the system's design envelope quickly leads to runaway
activation.

Stochastic subsampling of sets is the primary sparsity-controlling
mechanism in our toolbox. There are two elementary operations:
proportional decimation and signal capping.

Proportional decimation \(D\cdot A\) reduces the size of a set \(A\) by
a linear factor \(D \le 1\), randomly selecting the elements to be kept.
(This notation must not be confused with element-wise multiplication.)

The analogy of this operation in an electric circuit is resistance,
limiting current (elements transmitted per time step) in proportion to
the voltage (elements pushed into the system per time step).

Signal capping applies stochastic subsampling to limit the population
size to a constant value \(k\). It randomly chooses a subset
\(A_k \subseteq A\), retaining a population of \(\min(|A|, k)\).

\begin{quote}
\textbf{Stochastic subsampling}\\
Randomized removal of set elements by either proportional decimation
(reducing the population by a linear factor) or signal capping
(enforcing a maximum population).
\end{quote}

In an electric circuit, this function is similar to a current-regulating
diode. It effectively caps the number of elements transmitted per time
step. Applied to a system that accumulates signals over time (analogous
to an electric capacitor), it defines the signal capacity.

Beyond managing activation limits in networks, subsampling techniques
play an important role in topological memory retrieval, facilitating
pattern matching with composite data structures.

Bundles composed of SHRs with identical populations and zero overlap
create a representational symmetry, which affects memory retrieval. The
pattern matching algorithm cannot isolate a single best match from a
fully symmetric bundle, causing memory retrieval to fail and return an
empty result.

Stochastic subsampling breaks this symmetry. Decimating the bundle
shifts the underlying probabilities, making a unique match much more
likely. This fundamental mechanism is the basis for the
\hyperref[iterative-decomposition]{iterative decomposition}, graph
traversal, and related memory operations involving SHR bundles.

Subsampling typically affects some items more than others, occasionally
even pushing them below the memory's pattern matching threshold. Also,
spurious overlap between items introduces biases that further aid
symmetry breaking. Within just a few repetitions of this process, one
item from the bundle emerges as the unique best match.

Consider a bundle \(B\) composed of non-overlapping sets \(A_1\) and
\(A_2\), each containing 10 elements. Pattern matching \(B\) against an
item memory results in a tie between the two sets. But subsampling \(B\)
by removing just one element always breaks the tie. Removing two
elements has an \(18/38\) probability of breaking the tie (either
\(A_1\) or \(A_2\) losing two elements), and a \(20/38\) probability of
maintaining the tie (both \(A_1\) and \(A_2\) losing one element).
Removing three elements always breaks the tie. The combinatorics becomes
unwieldy when more than two items are involved, and if items overlap.

Simulations show that subsampling by a decimation factor
\(D \approx 0.8\) works well across a wide range of usage scenarios and
configurations. As a prerequisite for this mechanism, memory
hyperparameters must be chosen so that the SHR population \(P\) exceeds
the pattern matching threshold \(T\) (for instance \(P = 10\) and
\(T = 7\)).

The core algorithm's inability to resolve pattern matching ties on its
own should not be considered a limitation. Operating on idealized
clean-room inputs, it functions with mathematical precision. We
systematically mitigate this effect by subsampling inputs involving SHR
bundles. This is a sound approach, simulating the noise inherent in
biological systems.

In a biological brain, the combination of many mechanisms contributes to
our mathematical abstraction of stochastic subsampling. A ``warm and
wet'' neural system is naturally noisy. For example, minute variations
in signal synchronization or synaptic jitter are sufficient to have a
sampling effect on neural codes.

Associative memory systems are generally known for their robustness
against signal degradation. The topological memory framework adds a new
angle to this perspective: signal degradation is a prerequisite for the
memory to operate on composite data. By breaking representational
symmetry, stochastic subsampling causes associative memory retrieval to
make random choices. This aligns with the probabilistic behavior
observed in different aspects of cognition.

\subsection{Permutation}\label{permutation}

Alongside \hyperref[bundling]{bundling} and \hyperref[binding]{binding},
permutation is a fundamental operation for manipulating sets. Permuting
a set is the primary mechanism for encoding its position within a
temporal sequence.

We use the notation \(\pi A\) for a permutation \(\pi\) (``pi'') applied
to a set \(A\). This operation replaces each element with a different
element using a fixed mapping. The set-theoretic equivalent of this
process is a bijection (or automorphism) \(\pi : U \to U\) that maps the
universal set \(U\) onto itself.

Permutation of sparse sets has the following properties:

\begin{itemize}
\tightlist
\item
  The permutation of a set is dissimilar to the original set.
\item
  Applying the same permutation to two sets preserves their mutual
  similarity.
\item
  Permutation distributes over bundling:
  \(\pi(A \cup B) = \pi A \cup \pi B\).
\end{itemize}

For practical purposes, we can set up permutations so that there are no
fixed points (elements mapped onto themselves) to maximize the
dissimilarity between sets and their permuted counterparts. A
computationally efficient approach is to increment each element by one
modulo the universal set's cardinality. This mapping is known as cyclic
shift, roll, or right-rotation.

In biologically plausible models, information processing must be
generally invariant against the natural permutation of signals as they
travel along neural pathways. The exception to this rule: by design,
recurrence inside the auto-associative architecture does not permute
information.

Permutation is an invertible operation, and the cyclic shift approach is
particularly convenient as its inverse is simply a corresponding
left-rotation. But computational models reliant on inverse permutations
assume a bidirectional mathematical symmetry that lacks a clear neural
analogue. To maintain full alignment with forward-propagating neural
pathways, this framework treats permutation as a one-way physical
shuffling of signals --- an invariable, hard-wired feature of the
network.

\begin{quote}
\textbf{Permutation} A hard-wired network operation that replaces each
element of a set with a different element using a fixed mapping. It is
primarily used to encode sequence positions and temporal ordering.
\end{quote}

If a circuit contains multiple pathways that permute their signals, each
of these permutation mappings must be unique to maintain the
orthogonality of representations. Therefore, this computational
framework avoids the cyclic shift shortcut in favor of proper, randomly
generated permutation maps.

Within our computational framework, permutation is an essential
mechanism for encoding \hyperref[sequences-and-pointer-chains]{pointer
chains} and temporal ordering via \hyperref[recurrence]{recurrent
networks}.

\section{Auto-Associative Memory}\label{auto-associative-memory}

Autoassociations are relations of the form \(A \to A\), mapping sets
onto themselves. Although superficially similar to an identity
operation, auto-associative memory is a feature-rich and versatile
mechanism. In particular, it serves as the engine of symbol
manipulation: a function that takes symbolic input (encoded as SHRs) and
returns symbolic output must preserve dimensionality --- the defining
property of autoassociations.

For context, the topological memory algorithm natively handles
heteroassociations \(A \to B\). This is the more general type, where
\(A\) and \(B\) are sets in distinct representational spaces.

\begin{quote}
\textbf{Auto-associative memory}\\
A recurrent memory topology where the input and output layers are
shared.
\end{quote}

To use the hetero-associative programming interface for autoassociations
\(A \to A\), we must configure it with identical hyperparameters
(dimensions and populations) for \(A\) and \(B\). Technically, this
setup still allows us to store heteroassociations where \(A\) and \(B\)
have the same dimension. To achieve strict auto-associative
functionality, we must ensure that only relations \(A \to A\) are stored
in memory.

Auto-associative memory can handle any type of set, regardless of its
semantic interpretation: sparse holographic representations
(\hyperref[sparse-holographic-representations]{SHRs}), their
\hyperref[bundling]{bundles}, partitioned sets resulting from disjoint
bundles, memory pointers, and general sparse distributed representations
(\hyperref[sparse-distributed-representations]{SDRs}). Notably, the
\hyperref[binding]{binding} operation for symbolic representations is
realized through storing SHR bundles in auto-associative memory.

\subsection{Irreducibility}\label{irreducibility}

Storing an autoassociation in topological memory is an irreducible
process. In other words, the memory state of a complete autoassociation
cannot be recreated by piecing together smaller, partial write
operations. Conversely, no sequence of auto-associative write operations
combines in memory to form a new autoassociation.

The irreducibility of autoassociations is due to the binary topological
storage model. If the system used continuous weights or integer
counters, the repeated auto-correlation of elements would cause
self-correlating links to inflate. The memory would collapse to an
identity operator: it would cease to perform pattern completion and
instead simply echoes any input pattern directly back as the output.
This underlines that binary memory storage, modeling the presence of a
topological connection, is not just an optimization but a prerequisite
for auto-associative memory functionality.

Here is a simple proof of the irreducibility of autoassociations.
Consider a decomposition of \(A\) into \(n\) subsets
\(\lbrace A_1, A_2, \ldots, A_n\rbrace\). We rewrite the memory's
combinatorial expansion, corresponding to the 2-subsets
\(\binom{A}{2}\), by taking all pairs of subsets \(\{A_i, A_j\}\) and
expanding their unions. For \(n=3\), this results in the identity:

\[\binom{A}{2} = \binom{A_1 \cup A_2}{2} \cup \binom{A_2 \cup A_3}{2} \cup \binom{A_3 \cup A_1}{2}\]

In the general case where \(A\) is split into \(n\) subsets, the
combinatorial expansion consists of \(n(n-1)/2\) pieces. Together, these
result in the same combinatorial expansion as \(A\). This shows that the
memory's latent state is reducible. But the memory storage process is
not. Sequential storage of the autoassociations
\(A_i \cup A_j \to A_i \cup A_j\) does not replicate the same memory
state that results from writing \(A \to A\): for instance, the hidden
layer expansion of \(A_1 \cup A_2\) never connects to the output
elements in \(A_3\).

Although the memory's hidden layer activation is reducible, the memory
storage process is irreducible. This property has non-trivial
consequences: a memory cannot be learned piece-by-piece. Learning a new
unit of information requires all of its elements to be present
simultaneously.

Irreducibility is a prerequisite for auto-associative memory to function
as an item store. Every distinct item creates a memory trace that can be
individually retrieved. Items do not morph into other items as the
memory is being populated. Borrowing from the terminology of Gestalt
psychology: in auto-associative memory, the whole is other than the sum
of its parts.

\begin{quote}
\textbf{Irreducibility of auto-associative memory}\\
The process of storing an autoassociation \(A \to A\) in memory cannot
be broken up into partial write operations. All elements of \(A\) must
be stored at once.
\end{quote}

Auto-associative topologies effectively store collections of sets. This
functionality can be simply modeled with a database where the retrieval
process computes the overlap between the query set and all stored
records. This is functionally identical to auto-associative memory
retrieval, but of course impractical for large datasets.

In contrast to autoassociations, heteroassociations can be
\hyperref[properties]{decomposed} into separate storage operations.
Overlapping associations merge in memory to form a new association which
absorbs its original constituents. This unique functionality cannot be
replicated with a simple database model.

Autoassociations have another important property, closely related to
their irreducibility: memories cannot be retrieved in pieces. Querying
the memory with a subset of a known item resolves to the full set. In
other words, the memory cannot separate an item from its subsets.

This is an immediate consequence of the underlying storage model. For a
proper subset \(A_1 \subset A\), storing both \(A_1 \to A_1\) and its
enclosing set \(A \to
A\) subsumes the association \(A_1\) --- regardless of the order in
which they are stored. When queried with \(A_1\), the core retrieval
algorithm cannot recall this subset as an independent item. Instead, it
completes the pattern and pulls up the enclosing set \(A\). We term this
property \emph{subset absorption}.

\begin{quote}
\textbf{Subset absorption}\\
The property by which a subset loses its independent identity when its
enclosing set is present in memory.
\end{quote}

A practical consequence of this property: the memory retrieval mechanism
cannot untangle symbolic token bundles (hyperassociations). A separate
memory that stores the tokens individually (an item store) is required
to decompose token bundles.

\subsection{Exact nearest-neighbor
search}\label{exact-nearest-neighbor-search}

Auto-associative memory functions as a database for sets. Although the
system does not keep records of individual write operations, memory
retrieval can isolate any distinct pattern stored in memory. This is a
direct consequence of the \hyperref[irreducibility]{irreducibility
property} of the storage process, as discussed above.

When queried with an input pattern, memory retrieval finds its nearest
neighbor among the patterns stored in memory. This mechanism is driven
solely by the coincidence (intersection) of elements, ignoring
non-overlapping elements: pattern matching maximizes the overlap metric.
The process does not minimize the Hamming distance.

This retrieval mechanism is independent of the overlap characteristics
and the size distribution of the sets stored in memory: auto-associative
memory can represent highly overlapping SDRs as well as SHRs or SHR
bundles.

Nearest-neighbor search systematically identifies the exact best match
rather than an approximation. This is a direct consequence of the
memory's discrete data representation.

\begin{quote}
\textbf{Exact nearest-neighbor search}\\
Topological auto-associative memory retrieves information by identifying
the \textgreater{} absolute best match, defining proximity by the
overlap metric.
\end{quote}

\subsection{Update rules}\label{update-rules}

Autoassociations require the input and output spaces to be identical,
which is realized through memory-internal recurrence. Output retrieved
from memory is routed back, updating the state of the memory's input
layer. How the output interacts with the input is governed by the
memory's update rule.

In this framework, hetero-associative memory cannot simulate
autoassociations because, as discussed earlier, an external recurrent
pathway would inevitably permute the signal. Internal recurrence within
the auto-associative topology is the only mechanism that leaves feedback
unscrambled.

The choice of the update rule defines the system's dynamics. Below are
the most common configurations for managing the relationship between the
memory's input layer (state) \(X\) and the computed output \(Y\).

{\def\LTcaptype{none} 
\begin{longtable}[]{@{}
  >{\raggedright\arraybackslash}p{(\linewidth - 6\tabcolsep) * \real{0.2430}}
  >{\raggedright\arraybackslash}p{(\linewidth - 6\tabcolsep) * \real{0.2150}}
  >{\raggedright\arraybackslash}p{(\linewidth - 6\tabcolsep) * \real{0.3551}}
  >{\raggedright\arraybackslash}p{(\linewidth - 6\tabcolsep) * \real{0.1869}}@{}}
\toprule\noalign{}
\begin{minipage}[b]{\linewidth}\raggedright
\textbf{Update Rule}
\end{minipage} & \begin{minipage}[b]{\linewidth}\raggedright
\textbf{Set Operation}
\end{minipage} & \begin{minipage}[b]{\linewidth}\raggedright
\textbf{Function}
\end{minipage} & \begin{minipage}[b]{\linewidth}\raggedright
\textbf{Mode}
\end{minipage} \\
\midrule\noalign{}
\endhead
\bottomrule\noalign{}
\endlastfoot
replacement & \(X \gets Y\) & pattern completion \& denoising, item
store & converging \\
augmentation & \(X \gets Y \cup X\) & incremental convergence &
converging \\
coincidence & \(X \gets Y \cap X\) & coincidence detection &
converging \\
residual & \(X \gets X \setminus Y\) & novelty detection & converging \\
complement & \(X \gets Y \setminus X\) & prediction & generative \\
difference & \(X \gets Y \Delta X\) & change detection, novelty filters
& generative \\
\end{longtable}
}

\emph{Replacement} update is the classic auto-associative setup, most
useful for pattern completion, denoising, and item stores. With this
configuration, the auto-associative memory converges to a stable state.
Stability is normally reached with just a single cycle because denoising
iterations are already encapsulated within the memory retrieval
algorithm.

The \emph{augmentation} update rule completes the input while retaining
non-matching elements. Used iteratively in conjunction with stochastic
subsampling, this mechanism converges incrementally to a stable state.

The \emph{coincidence} update mechanism retains only the matching
elements, without completing the pattern. This rule can also be applied
in hetero-associative memory. All other update rules discussed here are
unique to autoassociations.

The \emph{residual} update rule removes the retrieved pattern from the
memory's state, leaving only the novel elements.

The \emph{complement} update rule removes the current state (input) from
the retrieved pattern (output). This isolates the missing or predicted
elements, moving the system's state forward in generative tasks. This
mechanism can emulate heteroassociations and bidirectional memory. It is
also the basis of pattern replacement (``production rules'') operating
on bundled representations.

Closely related to the \emph{complement} rule is the \emph{difference}
update rule: This mechanism retains the symmetric difference between the
memory's input layer and the pattern retrieved from memory, removing
only the matching elements.

The functional dynamics of auto-associative topological memory with
distinct update rules map to well-known paradigms in computational
neuroscience and classical AI. The \emph{replacement} rule realizes the
attractor dynamics classically associated with Hopfield networks and
Kanerva's sparse distributed memory. The \emph{coincidence} logic is
functionally equivalent to the matching mechanisms of adaptive resonance
theory. The \emph{residual} and \emph{difference} rules inherently
execute the error-propagation and filtering concepts central to
predictive coding and Kohonen novelty filters, while the
\emph{augmentation} rule serves as a discrete counterpart to evidence
accumulation models. By simply swapping the set operator in the update
mechanism, this framework unifies the behavioral dynamics of these
diverse cognitive models under a single mathematical architecture.

\subsection{Auto-associative feedback
systems}\label{auto-associative-feedback-systems}

Auto-associative memory is inherently a feedback system because its
input and output layers are shared. The memory's dynamic behavior
depends on the specific update rule, the learning condition, stochastic
rate limiting and proportional decimation, and the interaction of the
feedback signal (excitatory or inhibitory) with new incoming inputs.

Depending on the update rule, two structurally different modes of
operation emerge: in the absence of new input, the auto-associative
feedback system either converges to a stable state, or generates a
temporal sequence.

Specifically, the \emph{complement} and \emph{difference} update rules
give rise to generative behavior: an initial signal can trigger
self-sustained update cycles, solely driven by the auto-associative
feedback loop. The exact dynamics depend on the overlap structure of the
query patterns and of the representations stored in memory. Generative
behavior can manifest in different shapes: toggling between two states,
short repeating cycles, deterministic self-terminating sequences, or
long random walks.

Beyond the initial trigger which starts a generative temporal sequence,
external events can interact with sequence generation at any cycle.
Depending on the merging mechanism, new input may influence the
direction of the generated sequence, or start a completely new sequence.

Auto-associative memory typically learns new information if memory
retrieval fails and the query population lies within certain boundaries.
In scenarios where any input is considered ground truth, the memory may
be configured to store any query term.

A generic auto-associative feedback algorithm consists of the following
steps:

\textbf{Configuration:} Auto-associative memory \(M\), stochastic
decimation factor \(D\), learning threshold \(L\), pattern matching
threshold \(T\).

\textbf{Input:} Sets \(A_t\) (external input) and \(X_{t-1}\) (feedback
from previous cycle \(t-1\))

\textbf{Output:} Set \(X_t\)

\begin{enumerate}
\def\labelenumi{\arabic{enumi}.}
\item
  \textbf{Merge feedback:} \(X \gets A_t \cup X_{t-1}\)
\item
  \textbf{Memory retrieval:} \(Y \gets M(D \cdot X)\)
\item
  \textbf{Learn:} If \(Y = \varnothing\) and \(\vert X\vert \ge L\),
  store \(X\to X\) in \(M\) and let \(Y \gets X\)
\item
  \textbf{State Update:} \(X_t \gets \texttt{updaterule}(Y, X)\)
\item
  \textbf{Output}: Return \(X_t\)
\end{enumerate}

This algorithm template can be adapted to different usage scenarios:
instead of being merged with external input, feedback may be processed
as an inhibitory signal. Memory retrieval may be repeated a few times
until a non-empty response emerges. The memory's learning dynamics may
be governed by the similarity or distance between the query and the
retrieved pattern.

\subsection{Taxonomy of
autoassociations}\label{taxonomy-of-autoassociations}

Topological associative memory can generally handle datasets with
different overlap characteristics, pattern size distributions, and
block-coding partitioning schemes.

Within the taxonomy of auto-associative datasets, we primarily
distinguish between SDR-based data (perception) and SHR-based data
(knowledge representation and reasoning).

An auto-associative memory that stores sparse distributed
representations (SDRs) is the equivalent of a \emph{vector database} for
sets. It typically processes encodings of real-world data. Set
populations may vary within the system's sparsity envelope. Patterns
within a dataset can have any degree of overlap, ranging from very small
(dissimilar items) to very high (similar items).

For a given input pattern, auto-associative memory retrieval finds the
exact nearest neighbor within the stored dataset. In typical
applications, memory retrieval does not require stochastic subsampling.

Data from disjoint domains (for example, colors and positions) are
represented by partitioned sets, typically with different dimensions.
Block coding within auto-associative memory correlates patterns across
partitions. This enables nearest-neighbor search within a continuum of
multi-dimensional data representations. Within a specific memory
instance, the same partitioning must be applied consistently to all
representations.

We use the type signature \((d_1,\ldots,d_n)\) for a representation
partitioned into \(n\) disjoint SDRs.

Autoassociations of sparse holographic representations (SHRs) are
classified by the partitioning structure and by the data type within
each block: individual tokens represented as SHRs, or token bundles. The
memory topology auto-correlates each individual block and
cross-correlates separate blocks.

A partition containing individual tokens functions as a block item
memory. Representations have consistent populations and are dissimilar,
achieved through randomized encoding with minimal overlap. Memory
retrieval cleans up input patterns, removing noise and completing
missing elements.

Beyond storing individual tokens, a partition may contain token bundles,
generally representing hyperedges. We term an auto-associative memory
that stores token bundles a
\hyperref[hyperassociations]{hyper-associative memory}.

Bundle sizes are roughly multiples of the individual SHR population. Two
bundles that share tokens approximately overlap by a multiple of the SHR
size (with small variations due to spurious interference). Stochastic
subsampling is essential when retrieving bundled tokens from memory.

We denote individual tokens as \(h\), token pairs as \(2h\), and
arbitrary-size token bundles as \(*h\). The type signature of a
composite data structure is given by a tuple, specifying the contents of
each block. For example, \((h)\) denotes an item memory, and \((h, *h)\)
the combination of a single token with a hyperedge.

The following table outlines the taxonomy of common auto-associative
memory configurations with sparse holographic representations.

{\def\LTcaptype{none} 
\begin{longtable}[]{@{}
  >{\centering\arraybackslash}p{(\linewidth - 2\tabcolsep) * \real{0.2807}}
  >{\raggedright\arraybackslash}p{(\linewidth - 2\tabcolsep) * \real{0.7193}}@{}}
\toprule\noalign{}
\begin{minipage}[b]{\linewidth}\centering
\textbf{Type Signature}
\end{minipage} & \begin{minipage}[b]{\linewidth}\raggedright
\textbf{Function}
\end{minipage} \\
\midrule\noalign{}
\endhead
\bottomrule\noalign{}
\endlastfoot
\((h)\) & knowledge entities, item memory for symbolic tokens \\
\((2h)\) & unordered bidirectional memory \\
\((h,h)\) & directed knowledge graph, bidirectional memory \\
\((3h)\) & unordered triples \\
\((h,h,h)\) & ordered triples \\
\((h,h,h,\dots)\) & \(n\)-ary knowledge graph \\
\((*h)\) & hyperedge, undirected hypergraph \\
\((h, *h)\) & labeled undirected hypergraph, nested hypergraphs \\
\((*h, *h)\) & directed hypergraph \\
\((h, *h, *h)\) & labeled directed hypergraph \\
\((*h, *h,\ldots)\) & \(n\)-ary hypergraph \\
\end{longtable}
}

Bridging data from perception to knowledge representation and reasoning,
a hybrid auto-associative memory may combine SDR partitions and SHR
partitions.

For instance, a representation of type \((d,h)\) maps similar SDRs \(d\)
to symbolic tokens \(h\), acting as a classifier when queried with an
SDR. But unlike a conventional classifier, this memory can also be
queried in reverse, returning the patterns associated with a specific
token.

\subsection{Bidirectional associative
memory}\label{bidirectional-associative-memory}

Bidirectional associative memory maps a set \(A\) to a set \(B\) and
simultaneously \(B\) to \(A\). The mapping and its reverse are stored in
memory as a single association, with no preferred direction. Memory
retrieval works, as usual, with partial or noisy inputs.

Unlike its neural-network-based counterpart known as BAM (introduced by
Bart Kosko in 1988), which requires an extra network layer,
auto-associative topological memory functions inherently in a
bidirectional mode.

There are two ways to program a bidirectional memory: using either
pairwise bundling for a symmetric (unordered) relation, or block coding,
which preserves the position of the two representations.

The pairwise bundling approach realizes an unordered bidirectional
association: we simply store the autoassociation of the bundle
\(A \cup B\) and use the \hyperref[update-rules]{complement update} rule
for memory retrieval. Querying the memory with \(A\) completes the
pattern \(A \cup B\) and yields \(B\) after applying the update rule.
Conversely, a query with \(B\) returns \(A\).

Because a bidirectional memory explicitly stores pairs, it cannot
simultaneously function as an item memory (storing single items) or
handle bundles of three or more items.

The pairwise bundling approach to bidirectional memory relies on the
dissimilarity of the two sides of the relation. If the sets \(A\) and
\(B\) slightly overlap (which is normal for random SHRs), a query with
\(A\) returns a slightly reduced version of \(B\). The update rule has
removed the overlapping elements. For SHRs, such small degradation is
negligible because any subset carries the same semantic meaning, and
subsequent associative memory operations tolerate minor noise.

Feeding the reduced version of \(B\) back into the memory reliably
recovers the full, noise-free version of \(A\) because the update rule
will not remove elements absent from \(B\). One more pass through the
memory returns the noise-free version of \(B\). A round trip through
bidirectional associative memory cleans up the input, much like a
regular item store.

Beyond simple pairwise bundling, a bidirectional associative memory can
also be realized via \emph{block coding}. In this scenario, an
auto-associative memory stores a partitioned set \((A,B)\).

\begin{figure}[H]
\centering
\includegraphics[width=0.333\linewidth,height=\textheight,keepaspectratio,alt={A bidirectional associative memory, realized as an auto-associative memory operating on a bipartite representation.}]{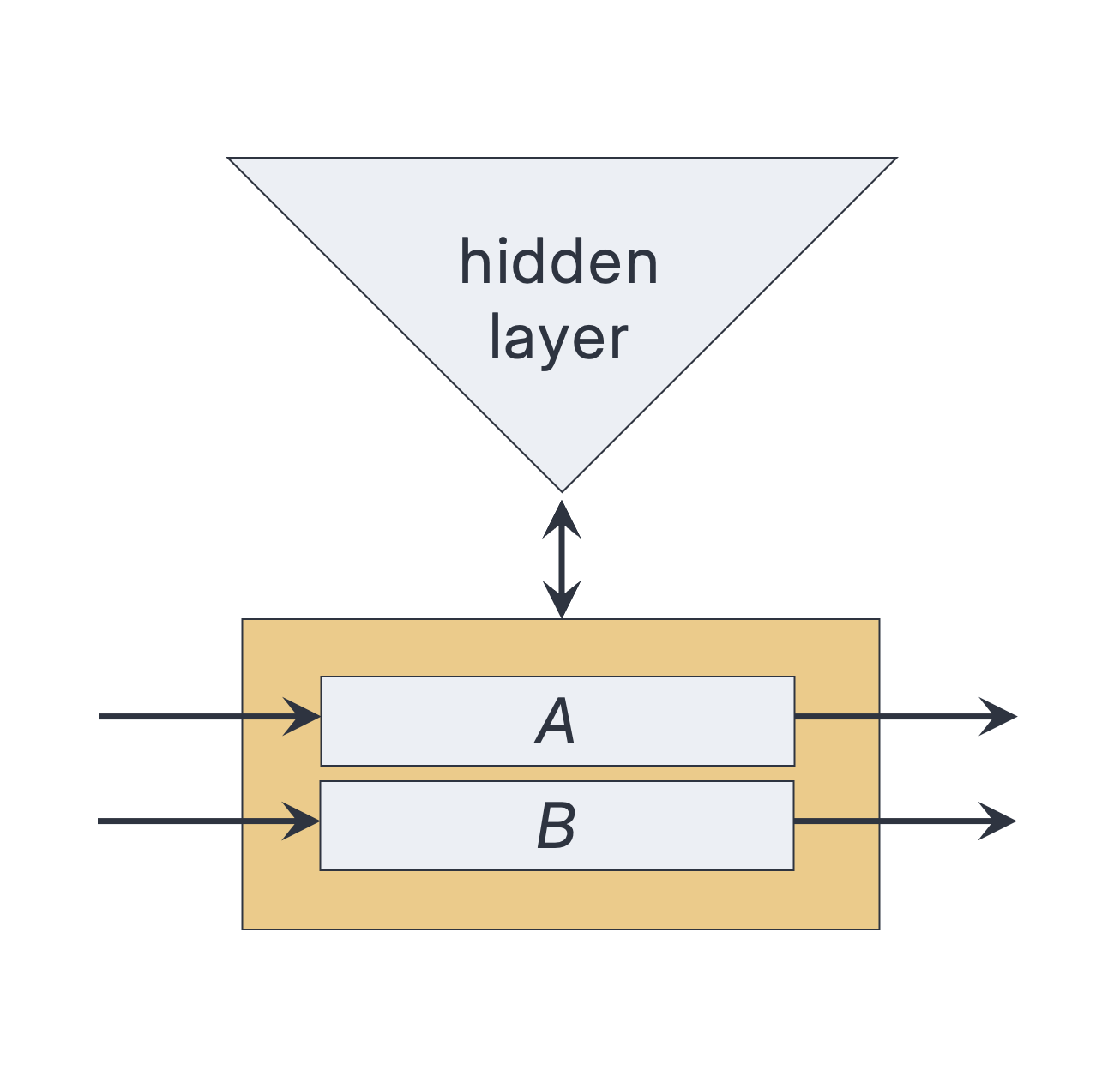}
\caption{A bidirectional associative memory, realized as an
auto-associative memory operating on a bipartite representation.}
\end{figure}

The two blocks may differ in dimensions, representing information from
distinct domains. This asymmetric bidirectional setup emulates a pair of
hetero-associative memories. The dynamics of this system are governed by
the memory's \hyperref[update-rules]{update rule}. Depending on the
choice of a particular recurrent update rule, the bidirectional memory
may exhibit reactive or generative behavior. By comparison, a
bidirectional memory composed of two hetero-associative networks is
always reactive, functioning as a cleanup memory.

With the \emph{replacement} update rule, bidirectional memory retrieval
functions via pattern completion. A query with either
\((A, \varnothing)\) or \((\varnothing,B)\) completes the association,
returning \((A,B)\). The patterns used in the memory query may be
incomplete or noisy. The block coding approach generates a noise-free
result within a single step, whereas the pairwise bundling method
discussed above requires a round trip to yield a clean result.

An auto-associative memory operating on a bipartite set can store
representations where one of the two blocks is empty. Mapping a set to
an empty block actively confirms it is known, without associating it
with other information. By comparison, regular memory retrieval maps
unknown sets to an empty result if no matching pattern is found --- this
mechanism cannot confirm that the query term exists.

Generally, bipartite autoassociations relate information from two
different domains, represented as sets with different dimensions and
populations. In the special case where both partitions share the same
dimension, the two blocks may represent data from a single domain.

A bipartite setup where each block stores a single SHR represents a
simple directed graph. If both blocks contain SHR bundles, the memory
represents a directed hypergraph. A labeled undirected hypergraph
corresponds to a setup where one block contains a single SHR (the label)
and the other block stores a bundle of SHRs (the hyperedge).

Within the same auto-associative topology, the memory's update rule
governs the dynamics of memory retrieval: clean-up and completion of
hyperedges (reactive update rules), or graph traversal (generative
update rules).

Partitioning auto-associative memory generally affects its capacity. The
capacity of a bidirectional memory, storing pairs of tokens \((A,B)\)
with dimension \(N\) and population \(P\), is given by

\[\texttt{capacity}(N, P) \approx \ln(2)\frac{N^2(N-1)}{P^2(P-1)}\]

This is derived as follows: the memory's storage space can be
parametrized as a prism with height \(2N\) whose base is a right
triangle with two equal sides of length \(2N\). This triangle describes
the expanded latent space, formed by the pair-wise combinations of the
elements in \((A,B)\). We divide this triangle into three parts: the
self-correlations within \(A\), the self-correlations within \(B\), and
the cross-correlations between the elements of \(A\) and \(B\).

\begin{figure}[H]
\centering
\includegraphics[width=0.292\linewidth,height=\textheight,keepaspectratio,alt={Memory space parametrization of a bipartite auto-associative memory.}]{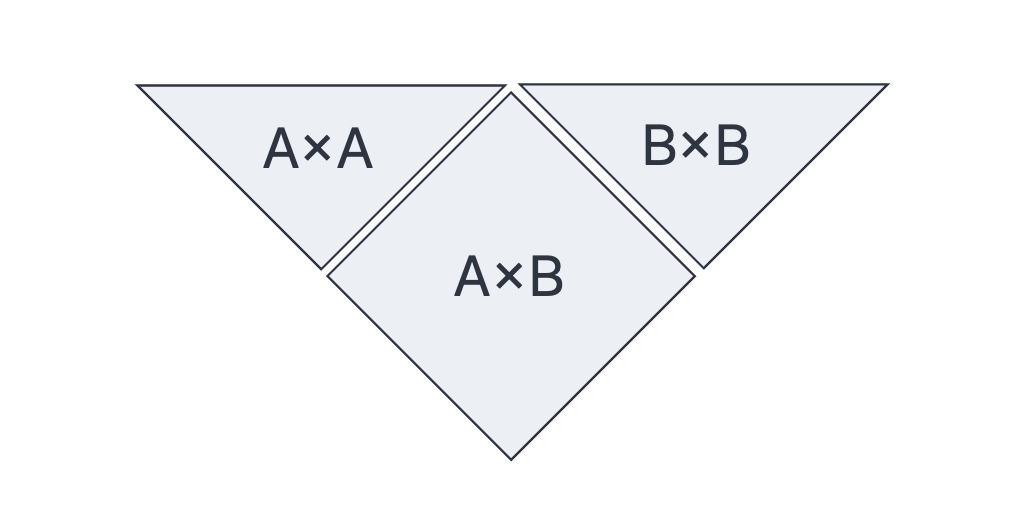}
\caption{Memory space parametrization of a bipartite auto-associative
memory.}
\end{figure}

When storing a collection of random autoassociations into this memory,
the cross-correlating division reaches its nominal capacity at memory
density \(\rho^* = 0.5\) first. The two self-correlating divisions fill
up more slowly because the underlying autoassociations require slightly
less storage. Therefore, we calculate the overall capacity solely on the
basis of the cross-correlating components.

Partitioning auto-associative memory with total dimension \(2N\) and
population \(2P\) into two equal-size blocks reduces its capacity by a
factor

\[\frac{C_{\text{bam}}}{C_{\text{auto}}}  = \frac{N(P-1/2)}{P(N-1/2)} \approx \frac{P-1/2}{P}\]

The approximation \((P-1/2)/P\) is valid for very large dimensions
\(N\).

A relatively small bipartite memory with two blocks of dimension
\(N = 1,000\) and population \(P = 10\) can store and accurately
retrieve nominally 769,393 bidirectional associations.

Note on the \hyperref[pattern-matching-threshold]{pattern matching
threshold} for partitioned autoassociations: calculating the threshold
based on an unpartitioned representation gives a mathematically
conservative upper bound. Since partitioning slightly reduces overall
capacity, this assumption safely protects against false-positive matches
without threshold adjustments. Therefore, the fact that the memory
algorithm is ``blind'' to partitioning does not impair its
functionality.

\subsection{Triplestore}\label{triplestore}

Ordered triples are a ubiquitous data structure for knowledge
representation and reasoning. A well-known application of ordered
triples is Minsky's \hyperref[frames]{frame model}, which organizes
knowledge in triple-layer objects composed of frames, slots, and
fillers.

In the context of the Resource Description Framework (RDF) and semantic
web technologies, semantic triples arrange information into subjects,
predicates, and objects. For instance, a semantic triple might state
``Italy is a country in Europe'', where ``Italy'' is the subject, ``is a
country in'' is the predicate, and ``Europe'' is the object.

The corresponding terminology in machine learning is
``head-relation-tail''.

Entity-attribute-value (EAV) models follow the same scheme, arranging
data into a triple of components: the entity (the object of interest),
the attribute (the property of the entity), and the value (the actual
data for that attribute).

We represent ordered triples with tripartite sets. The three blocks may
have different dimensions. A semantic triple, for instance, is encoded
via the disjoint union of three sets
\((\textit{subject}, \textit{predicate}, \textit{object})\). Here
\(\textit{subject}\) and \(\textit{object}\) must be sets with equal
dimensions, as the corresponding entities often flip roles. Depending on
the implementation, the \(\textit{predicate}\) set may be encoded using
a different dimension.

If we store a collection of ordered triples encoded as tripartite sets
in auto-associative memory, the result is a directed graph. Every triple
persisted in memory is unique because the memory always deduplicates
redundant data. Note that it is possible to write triples with an empty
block or two empty blocks into memory.

Querying a triplestore with partial data completes the pattern, removing
additive noise as usual within a single query. Memory retrieval works in
all three directions: if queried with any of the three blocks empty, the
missing information is filled in based on previously stored information.
This process may pull in information from multiple matching triples,
returning a superposition of possible answers. Assuming a sufficiently
low pattern matching threshold, memory retrieval also works if just one
of the three blocks is non-empty.

Using the core memory retrieval algorithm completes the input pattern.
But we can also impose one of the \hyperref[update-rules]{update rules}
of auto-associative memory for a modified behavior. For instance, using
the complement update rule turns memory retrieval into a dataflow
control structure.

Encapsulating a triplestore as a distinct component within a broader
cognitive architecture requires three input and three corresponding
output channels. These channels handle both training and inference
simultaneously. Rather than relying on an external toggle to switch
operating modes, the system achieves automatic online learning by
processing complete incoming triples as training data and partial inputs
as queries.

If the \hyperref[update-rules]{complement update} rule is applied,
complete inputs (whether novel or already known) always return an empty
output. Partial inputs, on the other hand, generate partial outputs and
simultaneously reverse the direction of the dataflow.

Standard relational databases and triplestores suffer from scaling
latency. A topological memory triplestore bypasses this: it performs
queries in constant time regardless of how many triples sit in memory.
New data is stored on the fly and becomes available immediately, without
preprocessing. Paired with the memory's high capacity, these
exact-match, zero-delay dynamics make this topological triplestore
attractive for high-throughput technical applications.

\subsection{Hyperassociations}\label{hyperassociations}

Under the \hyperref[taxonomy-of-autoassociations]{taxonomy} of
auto-associative memory setups, hyper-associative memory is defined by
the storage of bundled sparse holographic representations (SHRs),
modeling knowledge hypergraphs.

\begin{quote}
\textbf{Hyperassociation}\\
An autoassociation of bundled
\hyperref[sparse-holographic-representations]{sparse holographic
representations} (SHRs).
\end{quote}

A bundle of symbolic tokens, encoded as SHRs, is the equivalent of a
hyperedge. The tokens represent the hypergraph's vertices. A collection
of such hyperedges forms a simple undirected hypergraph. We call this
particular kind of hypergraph ``simple'' because it satisfies two
conditions: it has no empty edges, and no edge is a subset of another
edge (known as the Sperner property). Edges are non-empty because we
cannot write an empty set into associative memory. And the Sperner
property follows from the fact that storing subsets of known information
into memory has no effect on the memory state.

The following brief example demonstrates the mechanics of working with
hyperedges. First, we encode the symbolic tokens ``king'', ``man'',
``queen'', ``woman'', and ``person'' as random SHRs. Then we store the
following bundles (which act as semantic fields) into an
auto-associative memory:

\begin{verbatim}
> king man royalty -> king man royalty
> queen woman royalty -> queen woman royalty
\end{verbatim}

This builds two hyperedges in memory. A memory query using the
\emph{difference} update rule gives the union of the input and the
output, with the matching elements removed:

\begin{verbatim}
> king
man royalty
> man royalty
king
\end{verbatim}

The following query matches ``king'' and ``man'', leaving behind the
retrieved token ``royalty'', and the non-matching part ``woman''.

\begin{verbatim}
> king man woman
royalty woman
\end{verbatim}

Feeding this output back into memory generates ``queen'' as the final
result:

\begin{verbatim}
> royalty woman
queen
\end{verbatim}

Generalizing from undirected hypergraphs, we can build an \(n\)-ary
hypergraph by partitioning the overall representation into \(n\)
disjoint blocks, each containing bundled tokens.

As a common special case, a directed hypergraph is realized with two
partitions \((C,D)\), where \(C\) is the head and \(D\) is the tail of
the hypergraph.

We can add an extra block to the hyperedge representation to accommodate
a unique edge label, encoded as an SHR. Such labels function as unique
pointers to the data structure. This extends the object type from
hypergraph to multi-hypergraph. It allows the representation of empty
edges, edges that are subsets of other edges, nested hypergraphs, or
repeated edges that share the same nodes but are distinguished by their
unique labels. In the context of knowledge hypergraphs, the edge label
usually corresponds to a predicate.

Storing hyperedges in memory creates a strong topological linkage
between pairs of vertices occurring within the same hyperedge: the
autoassociation of a bundle \(A \cup B\) (two vertices encoded as SHRs)
self-correlates the elements within \(A\) and within \(B\), and also
cross-correlates \(A\) with \(B\). Each vertex is usually
cross-correlated to several other vertices, and every connection
produces roughly the same number of links within the memory topology.
When the memory retrieval algorithm encounters multiple such cross
connections, it returns their superposition. This behavior is the basis
for hypergraph traversal.

The pattern matching mechanism within hyper-associative memory is best
demonstrated with a small hypergraph:

\begin{figure}[H]
\centering
\includegraphics[width=0.5\linewidth,height=\textheight,keepaspectratio,alt={Undirected hypergraph with 6 nodes and 5 hyperedges.}]{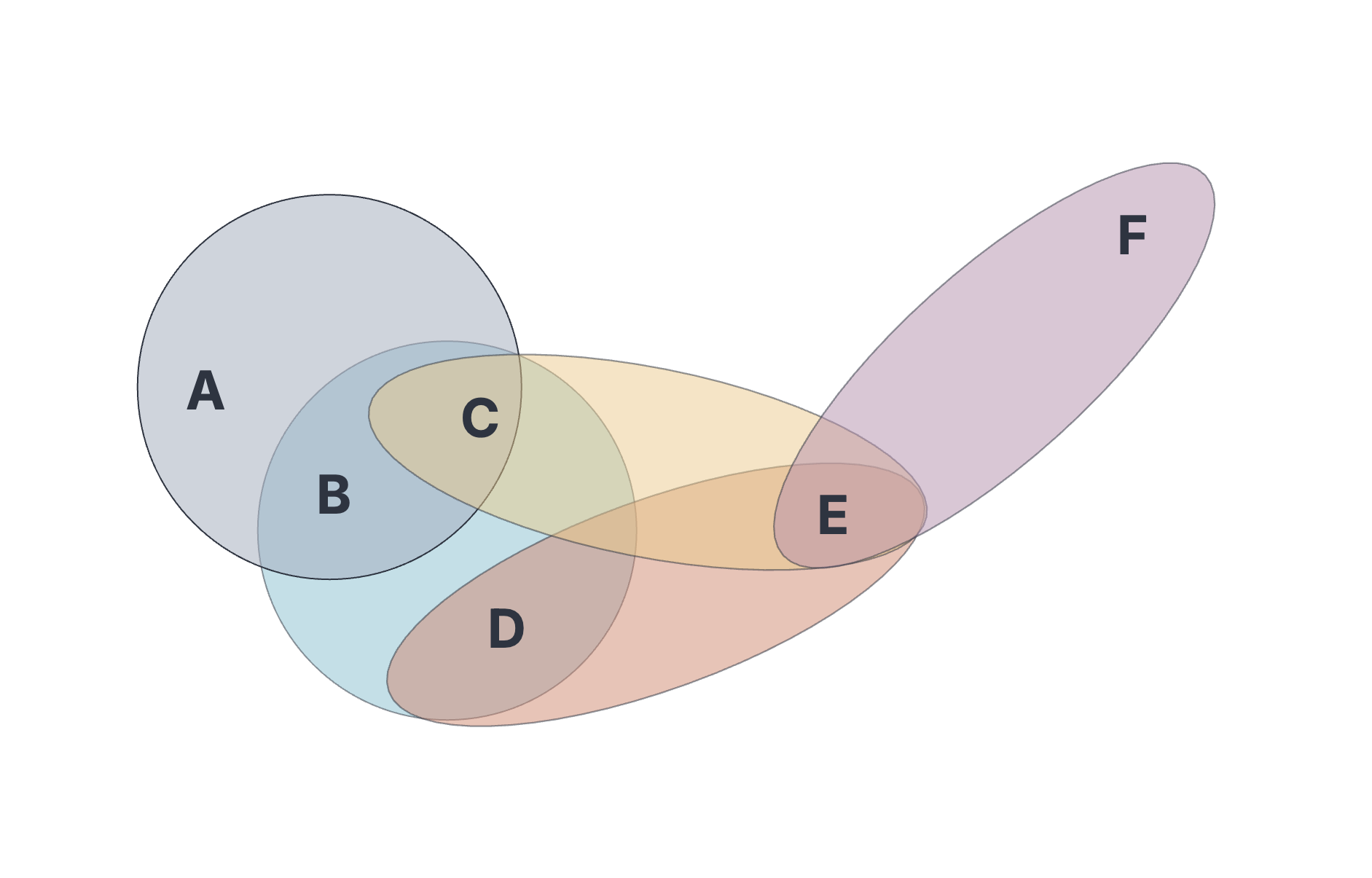}
\caption{Undirected hypergraph with 6 nodes and 5 hyperedges.}
\end{figure}

Vertices are encoded, as usual, via random SHRs with uniform population.
Apart from spurious overlaps, these representations are mostly
dissimilar.

Training the memory involves one write operation per edge:

\begin{verbatim}
> a b c -> a b c
> b c d -> b c d
> c e -> c e
> d e -> d e
> e f -> e f
\end{verbatim}

Querying the memory with just one vertex gives the superposition of all
directly connected vertices. With the default \emph{replacement} update
rule enabled, the query completes the pattern, and the result retains
the input:

\begin{verbatim}
> a
a b c
> b
a b c d
> c
a b c d e 
> d
b c d e
> e
c d e f
> f
e f
\end{verbatim}

If we query with a bundle of vertices that occur within the same edge,
the memory finds and combines all edges containing the input:

\begin{verbatim}
> a b
a b c
> b c
a b c d
> c e
c e
\end{verbatim}

Here the larger edge containing ``a'' and ``b'' wins over ``f'':

\begin{verbatim}
> a b f
a b c
\end{verbatim}

The memory retrieval behavior shown above is stable in the presence of
spurious overlaps, crosstalk from unrelated memory contents, and
subsampling of the input.

If we query with a combination of vertices that are not directly
connected, the retrieval behavior is unstable, depending on
representational overlaps and on the specific random subsample used as
input. Memory retrieval chooses among possible matches, each with a
certain probability that correlates with the number of matching
elements. Each outcome aligns with certain aspects of the input's
connectivity.

The following examples show possible outcomes for a query with the union
of ``a'' and ``e'', ranked by output size. (The probabilities of the
different outcomes depend on representation.) The largest match is the
cluster around ``e'':

\begin{verbatim}
> a e
c d e f
\end{verbatim}

followed by the edge that contains ``a'':

\begin{verbatim}
> a e
a b c
\end{verbatim}

This finds the common neighbor --- an interesting outcome:

\begin{verbatim}
> a e
c
\end{verbatim}

In some cases, no unique best match is found (a ``tie'' between equally
likely results): memory retrieval returns an empty output.

\begin{verbatim}
> a e
\end{verbatim}

Hyper-associative memory uses a vertex bundle to retrieve another bundle
linked to all or some of the inputs. For ambiguous queries, stochastic
subsampling selects one of several valid outcomes.

Walking a hypergraph vertex by vertex normally requires extracting a
single item from a retrieved bundle using a separate item store, then
feeding it into the next query.

Extracting a specific vertex at every step is unnecessary: we can
instead use the entire output from the previous step to compute the next
step. This process doesn't just move to one neighbor; it captures the
entire context. This mirrors how we can jump from a specific memory to a
broad category instantly. The random walk fluidly transitions between
entire groups of nodes based on their participation in overlapping
hyperedges.

Simulations of \hyperref[random-walks-on-graphs]{random walks} on larger
graphs are discussed further below to illustrate these dynamics.

\subsection{Iterative decomposition}\label{iterative-decomposition}

The ability to bundle sparse holographic representations would be of
very limited use without ``unbundling'' --- the mechanism to extract
items from a bundle, which is the inverse of the bundling operation.

We represent a bundle as a flat union of sets, so its elements cannot be
directly traced back to the distinct items from which the bundle was
formed. Furthermore, due to spurious interference some elements may
originate from multiple items, causing a small amount of data
compression.

The prerequisite for unbundling SHRs is an item memory that stores all
items involved in the bundle. The item memory must contain only
individual SHRs, but no bundles of SHRs.

Querying the item memory with a bundle of non-overlapping SHRs with
equal populations results in a pattern matching tie: memory retrieval
returns an empty result. We apply
\hyperref[stochastic-subsampling]{stochastic subsampling} to raise the
probability of a unique match emerging during memory retrieval. For
small bundles, one or two trials are normally sufficient to retrieve one
SHR.

Once a bundled SHR is matched with the item store, we delete it from the
bundle using the \hyperref[update-rules]{complement update} rule and
iterate the process. This iterative decomposition process extracts the
components of a bundle one after another.

\begin{quote}
\textbf{Iterative decomposition (unbundling)}\\
Step-by-step extraction of individual tokens from a bundle, using
stochastic subsampling to break representational symmetries and isolate
unique matches.
\end{quote}

This generates a \emph{random temporal sequence}, assuming uniform
populations of the items within the bundle. Each item has an
approximately equal chance of being recalled first (spurious overlaps
may introduce a small bias).

If the bundle to be decomposed consists of items with different
populations, larger items are more likely to emerge as the best unique
match. This allows us to manipulate the retrieval order, creating a
priority-based decoupling mechanism by modulating the size of the items
during bundling.

For instance, a \emph{last-in-first-out} queue can be realized by
scaling down older items every time a new item is added. This implements
temporal sequencing through magnitude decay.

Another interesting possibility: if we subsample (or transcode) items
with different populations that encode a ranking, iterative
decomposition functions as a \emph{natural sorter}: the most relevant
items pop out of the bundle in the first few iterations of the recovery
process.

Similarly, an attention mechanism might encode information that requires
\emph{priority processing} (for instance ``food'' or ``danger'') with a
higher-than-average set population.

We can experimentally measure the average number of iterations needed to
extract one item from a bundle:

\begin{figure}[H]
\centering
\includegraphics[width=0.708\linewidth,height=\textheight,keepaspectratio,alt={Average number of iterations needed to extract one token from a bundle, as a function of the number of bundled tokens. This is simulated with hyperparameters N=2,000 and P=10. The results are consistent with Miller's magical number seven, plus or minus two.}]{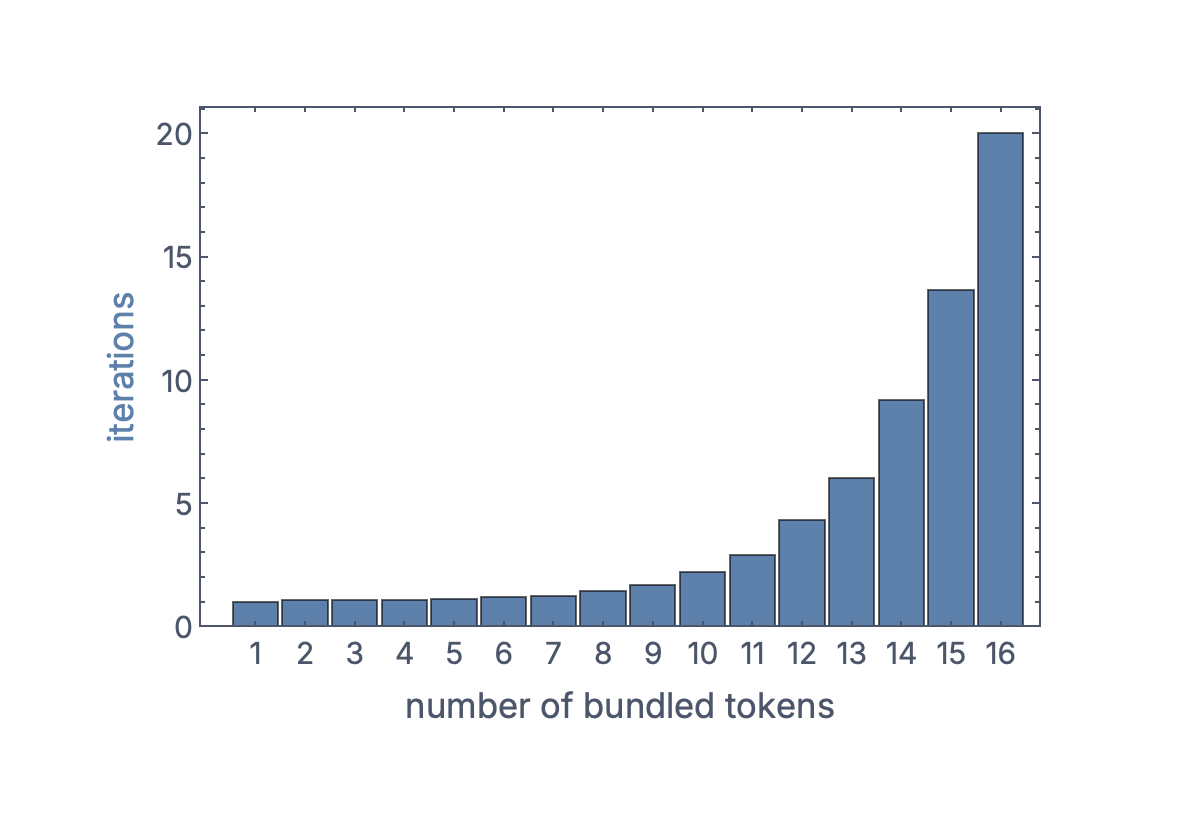}
\caption{Average number of iterations needed to extract one token from a
bundle, as a function of the number of bundled tokens. This is simulated
with hyperparameters \(N=2,000\) and \(P=10\). The results are
consistent with Miller's magical number seven, plus or minus two.}
\end{figure}

Extracting one item from a small bundle usually succeeds on the first
try, and rarely a second try is needed. This works reliably for bundles
composed of fewer than 10 items. For larger bundles, the number of
attempts needed grows rapidly. The process becomes impractical for
bundles larger than about 12 items. We observe this consistent behavior
within a wider range of hyperparameter configurations.

The number of items practically extractable from a bundle coincides with
George \emph{Miller's magical number seven, plus or minus two} (1956).
Miller discovered that the average human working memory has a capacity
of roughly \(7
\pm 2\) items. This correlation suggests that human working memory makes
use of stochastic bundle decomposition.

The bundle decomposition mechanism bears a striking similarity to a
quantum-mechanical measurement process. A quantum system exists in a
superposition of multiple basic states until a measurement forces it to
reduce to a single state (wavefunction collapse). Similarly, we can
interpret a bundle of SHRs as a superposition of multiple symbols.
Repeatedly measuring subsampled versions of this bundle against an item
store eventually reduces it to one of its basic constituents. Spurious
overlaps between items act like quantum interference, skewing the
probability of which symbol emerges. This aspect of the topological
memory model aligns with the concept of quantum cognition: it mimics the
quantum-mechanical formalism of superposition, interference, and
collapse --- but without requiring a physical quantum substrate.

\subsection{Binding}\label{binding}

Binding is one of the three fundamental operations used to compose and
manipulate high-dimensional distributed representations (the others
being \hyperref[bundling]{bundling} and
\hyperref[permutation]{permutation}). Binding is the glue that
associates distinct pieces of information, such as a key and its value,
or a role and its filler. This concept was introduced in 1990 by Paul
Smolensky in the context of tensor product representations --- arguably
the common ancestor of hyperdimensional computing, vector-symbolic
architectures, and the topological memory model discussed here.

To represent a structured object such as ``redcircle'' we use the
binding operation ``redcircle'' = ``red'' \(\otimes\) ``circle'', which
creates a compact higher-level representation of the two items. But the
same mechanism can create many other semantic structures, for example
semantic triples and knowledge graphs.

Some fundamental properties of the binding operation:

The binding ``redcircle'' is dissimilar (orthogonal) to both ``red'' and
``circle''. This is important because a ``redcircle'' should not be
confused with just the color ``red'' or the shape ``circle''. But while
the output is dissimilar to each of the inputs, it preserves relative
relationships. If ``oval'' is similar to ``circle'', then ``redcircle''
should be similar to ``redoval''. Binding also preserves the
dimensionality of the underlying representations.

Binding is invertible: If we have the pair and know one of its elements,
we can retrieve the other element.

\begin{quote}
\textbf{Binding}\\
An invertible set operation (\(A \otimes B\)) that links distinct items
(like a key and value) into a set that is dissimilar to its
constituents.
\end{quote}

Another defining property of the binding operation is distributivity
over bundling:
\(A \otimes (B \cup C) \approx (A \otimes B) \cup (A \otimes C)\). This
relation allows us to bundle multiple bound pairs into a single complex
data structure, and to perform queries on the compound expression.

Binding shares algebraic properties with multiplication. But defining an
operation that satisfies these properties and simultaneously preserves
sparsity proved challenging. Hyperdimensional computing with dense
representations, by comparison, does not have this problem.

To tackle the challenge of sparse binding, we remember that storing a
set in topological memory binds its elements. The memory algorithm
functions through pairwise binding, creating a web of links between all
input elements. Consequently, inverse binding is achieved naturally
through memory retrieval with partial inputs.

This mechanism is best illustrated in the common case
\(A_1 \otimes A_2 = B\), where two tokens are bound to a third. Storing
the bundle \(A_1 \cup A_2 \cup B\) in auto-associative memory binds the
three tokens to each other. Technically, writing
\(A_1 \cup A_2 \cup B \to A_1 \cup A_2 \cup B\) into memory links all
pairs of input elements to every element of the output, which is in this
case identical to the input. This operation is fully symmetric between
the involved tokens.

Memory retrieval inverts binding: Querying with \(A_1 \cup B\) extracts
\(A_2\), and a query with \(A_2 \cup B\) gives \(A_1\). Of course we can
also retrieve \(B\) through a query with \(A_1 \cup A_2\): memory
retrieval works in all three directions.

(Note that ``raw'' memory retrieval from auto-associative memory
completes the input pattern. The query mechanism as shown here uses the
\emph{complement} update rule.)

In the context of topological memory, the \emph{distributivity} property
materializes as follows. Consider two bindings \(A_1 \otimes B_1 = C_1\)
and \(A_2 \otimes B_2 = C_2\), both being stored in memory. We can think
of this as keys \(A_i\) bound to values \(B_i\). These key-value
bindings can now be combined into one dynamic data structure
\(C_1 \cup C_2\), which is dissimilar to the original keys and their
values. To extract the original values \(B_i\) from this compound data
structure, we simply query the memory with \(A_i \cup
(C_1 \cup C_2)\). And similarly, we can also extract a key from its
value.

Here is a practical example. We bind two items ``red'' and ``circle''
(both encoded as SHRs) by creating a higher-level object ``redcircle'',
and write their union into auto-associative memory. (The pseudocode used
here implies a union operation between adjacent tokens.)

\begin{verbatim}
> red circle redcircle -> red circle redcircle;
\end{verbatim}

Similarly for ``blue'' and ``square'':

\begin{verbatim}
> blue square bluesquare -> blue square bluesquare;
\end{verbatim}

Forming a dynamic representation \(A\) containing both chunks:

\begin{verbatim}
> A := redcircle bluesquare;
\end{verbatim}

The set \(A\), a bundle of two symbols, is a compressed representation
of the original information which consisted of four symbols. To extract
the underlying relations, we query the memory (using the
\emph{complement} update rule) with the union of \(A\) and any of the
entities:

\begin{verbatim}
> A red
circle
> A square
blue
\end{verbatim}

This mechanism accurately retrieves the original, noise-free keys and
values within a single pass. By comparison, the corresponding operations
in traditional hyperdimensional computing or vector-symbolic
architectures only give approximate results which require further
cleanup steps.

The same binding mechanism can be applied to numerous items at once,
only limited by the system's sparsity constraints:
\(A_1 \otimes A_2 \otimes \dots
\otimes A_f = B\) simultaneously binds \(f\) elements. And \(A_1 = B\)
binds a single item.

Conventional vector-symbolic architectures treat binding as a strict
dyadic (two-input) operator, requiring chaining for multiple objects.
Topological memory executes \(n\)-ary binding natively in a single
operation. Binding is fundamentally a memory-dependent operation. It is
a common process to replace two (or more) items with their binding,
which compresses information and maintains sparsity. In the context of
perception, this mechanism is known as \emph{chunking}: the grouping of
inputs into cohesive perceptual units.

To chunk two perceptions ``red'' and ``circle'', a memory query tests if
this combination is already known. By now we have probably encountered
many examples of red circles, so we can assume that the query returns
the binding ``redcircle''. But in case of a novel perception, for
instance an orange circle, the corresponding memory query fails: a new
random SHR representing the binding ``orange circle'' must be deployed.
This way, higher-level representations are generated on the fly. This is
a form of \hyperref[heteroencoder]{heteroencoding}.

Depending on the context, binding can be linked to long-term memory or
to a short-term associative memory where information decays quickly.

The ability to construct and deconstruct complex data structures through
various combinations of bundling and binding is a prerequisite of
symbolic reasoning. For instance, \hyperref[frames]{semantic frames} are
realized within auto-associative memory through a grid-shaped binding
structure.

\subsection{Sequences and pointer
chains}\label{sequences-and-pointer-chains}

Biological networks do not appear to dynamically allocate and persist
variable-length arrays. To be consistent with this constraint, we model
sequences as temporal rather than spatial processes.

Within the topological memory framework, there are several ways to
realize temporal sequences, either based on hetero-associative or on
auto-associative structures. Common mechanisms are based on pointer
chains, forming a skeleton sequence to which items are associated. This
skeleton can be driven by two distinct mechanisms: memory-linked chains
or permutation-linked chains.

In a memory-linked pointer chain, the sequence progression is stored in
the associative memory. A memory query with a given pointer retrieves
the next pointer in the sequence. The sequence naturally terminates when
querying the memory with the final pointer gives an empty result.

Alternatively, a pointer chain can be formed through repeated
permutations. Here, a single SHR represents the entry point to the
sequence and serves as the first pointer. Accessing the \(n\)-th element
of the sequence requires \(n-1\) permutations of the initial address.
With this mechanism, advancing the pointer to follow the sequence
without extracting the item payload is a highly efficient operation that
does not involve memory queries. This type of sequence terminates when a
memory query to extract a pointer's payload returns an empty result.

Sequence items (the payload) can be either bundled with the pointer, or
separated via block coding, using a bipartite representation. In either
implementation, separating an item from its pointer requires a memory
query.

The following diagram shows a bipartite auto-associative memory where
one block stores the items and the other block stores the pointers. An
external feedback line routes the permuted pointer back to the memory's
input. Each feedback cycle extracts one item from the sequence.

\begin{figure}[H]
\centering
\includegraphics[width=0.333\linewidth,height=\textheight,keepaspectratio,alt={Auto-associative memory storing a temporal sequence. Recurrent permutations create a pointer chain.}]{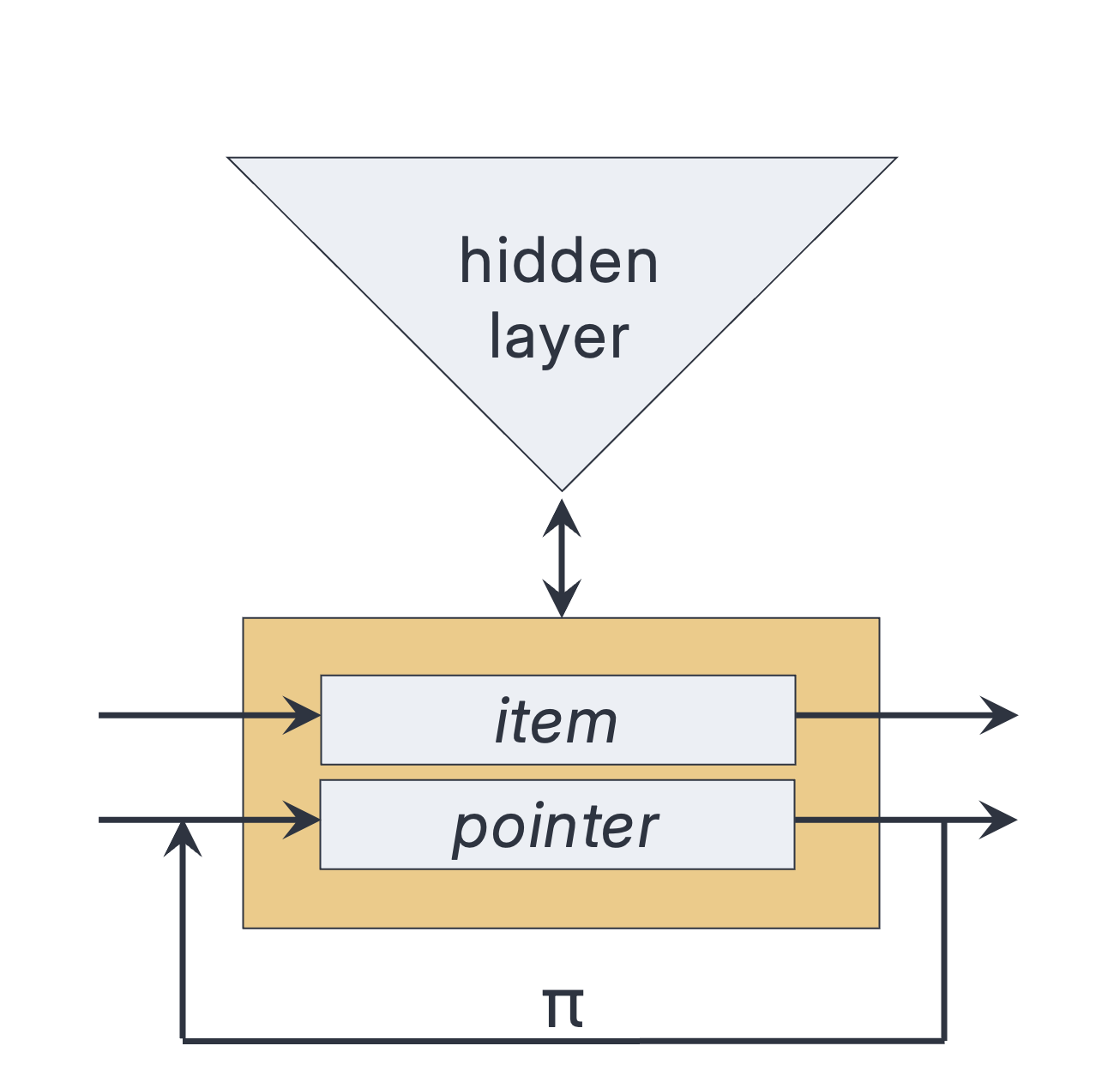}
\caption{Auto-associative memory storing a temporal sequence. Recurrent
permutations create a pointer chain.}
\end{figure}

This auto-associative system automatically learns the associations
between the items and the pointer chain when receiving a sequence of
items. To trigger inference, we prime the sequence with the initial
token while leaving the item block empty. A memory query completes the
item block, generating noise-free representations for the item and the
pointer.

The length of pointer chains, as the backbone of the sequence, is
practically unlimited. The achievable sequence length depends on the
lexicon size (the number of unique items): a larger lexicon distributes
information across a greater number of memory locations, resulting in a
higher capacity to store sequences. This mirrors how human memory can
remember long texts consisting of hundreds of words, but struggles to
remember a sequence of a hundred digits without chunking.

An alternative sequence memory based on a
\hyperref[reservoir-architectures]{reservoir architecture} and a
hetero-associative readout mechanism is discussed further below.

\section{Hetero-Associative Memory}\label{hetero-associative-memory}

Hetero-associative memory for discrete sets emerges from topologies with
a combinatorially expanded hidden layer. This network architecture maps
patterns from the input layer to a physically separate output layer. The
information flow within this network topology is purely feed-forward.

\begin{quote}
\textbf{Hetero-associative memory}\\
A feed-forward memory topology with separate input and output layers.
\end{quote}

The input layer and the output layer generally have different dimensions
and hold differently encoded information, often from distinct domains.

We denote heteroassociations by \(A \to B\), where \(A \subset U\) and
\(B \subset V\) are small subsets of their respective universal sets
\(U\) and \(V\).

Storing a heteroassociation in memory requires both \(A\) and \(B\) to
be specified. The presence of a secondary input for training (the
``teaching signal'') distinguishes hetero-associative memories from the
auto-associative architecture.

Memory retrieval takes an input pattern \(A\) and produces an output
\(Y\). The retrieval algorithm identifies the unique best match within
the aggregate information stored in memory: the largest subset
\(X \subseteq A\) in which all element pairs contribute to all elements
within \(Y\). This mechanism, functioning via partial pattern matching,
resolves degraded or noisy versions of known patterns into a clean
output.

For a successful match, \(X\) and \(Y\) must exceed the memory's minimum
pattern matching thresholds. Otherwise, an empty result is returned.
This mechanism filters known information from potential false-positive
matches based on the patterns' overlap probabilities.

A tie occurs if there are multiple possible matches that meet the above
criteria. When this occurs, the retrieval algorithm returns (by design)
an empty result.

\subsection{Properties}\label{properties}

\hyperref[irreducibility]{Irreducibility} is a defining property of
autoassociations: storing a set into memory cannot be replaced by a
sequence of partial write operations. Heteroassociations, by contrast,
can be built up incrementally. A hetero-associative memory query does
not necessarily recall a relation \(A \to
B\) exactly as it had been stored in memory. Instead, it retrieves the
best match based on all contributing relations previously stored in
memory. This means the same query may return different results as the
memory evolves over time. By comparison, auto-associative memory
maintains the integrity of stored items --- a prerequisite for handling
stable symbolic representations.

The following properties define the reducibility of heteroassociations:

Storing \(A \to B_1\) and \(A \to B_2\) is equivalent to storing
\(A \to B_1 \cup
B_2\): one-to-many associations result in a superposition.

But storing \(A_1 \to B\) and \(A_2 \to B\) separately is different from
storing \(A_1 \cup A_2
\to B\), because the separate operations do not cross-correlate the
elements of \(A_1\) with \(A_2\). This property allows the memory to
accurately represent many-to-one associations.

Storing \(A_1 \cup A_2 \to B\), \(A_2 \cup A_3 \to B\), and
\(A_1 \cup A_3 \to B\) is the same as storing
\(A_1 \cup A_2 \cup A_3 \to B\), because these three overlapping pairs
collectively cover all possible 2-subsets (cross-correlations) of the
complete union. Similar properties hold if the left-hand side of the
association is decomposed into four or more subsets.

Subset absorption is a property that hetero-associative memory shares
with auto-associative memory:

For a subset \(A_1 \subseteq A\), storing \(A_1 \to B\) and \(A \to B\)
is equivalent to storing \(A \to B\).

Similarly, for a subset \(B_1 \subseteq B\), storing \(A \to B_1\) and
\(A \to B\) is equivalent to storing \(A \to B\).

\subsection{Supervised learning}\label{supervised-learning}

A hetero-associative memory system configured for supervised learning
has two fundamental modes: training and inference. It stores an
association \(A \to B\) into memory if it receives non-empty signals
\(A\) and \(B\) (training). If \(B\) is an empty set, the system
retrieves and returns the pattern associated with the input \(A\)
(inference).

Storing either \(A \to \varnothing\) or \(\varnothing \to B\) does not
alter the memory's state.

The system learns new information in one shot --- there is no need to
repeatedly cycle through a training dataset.

A useful variant of this scheme combines training and inference: here,
the system always stores \(A \to B\) into memory and subsequently
queries the memory with \(A\). Note that this does not necessarily
return \(B\), because a combination of previously stored relations may
contribute to the retrieved pattern. This model continuously learns from
every incoming signal. Combined with a stochastic decay mechanism, the
memory refreshes frequent signals while fading out rarely occurring
ones.

Querying a hetero-associative memory system with an unknown pattern
\(A\) returns an empty result. This property may be used to modify the
system's learning dynamics so that it only stores new information if a
prior query returns an empty result. Here the boundary between inference
and learning is fluid: the system continuously evaluates its internal
model against reality, dynamically switching to a learning state
whenever it detects a knowledge gap.

\subsection{Predictive learning}\label{predictive-learning}

Predictive hetero-associative learning is the temporal equivalent of
supervised learning, which instead processes simultaneous spatial
associations.

In a predictive associative memory, cause and effect are separated by
one execution cycle. At every clock tick \(t\), the memory stores the
association \(A_{t-1} \to B_t\) and retrieves a prediction \(B_{t+1}\)
by querying the memory with the current input \(A_t\).

A variant of this model only learns if the prediction was incorrect, as
defined by a common distance or similarity metric. This mechanism is an
analogue to predictive processing (or predictive coding) in the brain.

Implemented as a self-contained system component, a hetero-associative
predictor simply carries its input \(A_t\) over to the following cycle.

\subsection{Heteroencoder}\label{heteroencoder}

In the topological memory framework, the transition from continuous
perception data to discrete symbolic tokens uses a dedicated bridging
mechanism: the heteroencoder. A heteroencoder is an unsupervised
learning process that translates overlapping perceptual data (SDRs) into
stable symbolic tokens (SHRs). Rather than extracting continuous hidden
features from an input, a heteroencoder acts as a discrete classifier.
It operates dynamically on the boundary between the known and the
unknown.

When a heteroencoder receives an input \(A\), it attempts to retrieve an
output value \(B\) from memory. The query fails if \(A\) is a novel
pattern. In this case, the encoder generates a new random SHR \(B\) on
the fly and stores the association \(A \to B\) in memory. This
autonomous learning mechanism guarantees that any input, as long as it
is above a certain size, returns an SHR.

\begin{quote}
\textbf{Heteroencoding}\\
An unsupervised learning process that maps sets (SDRs or SHRs) to stable
symbolic tokens (SHRs) on the fly.
\end{quote}

Heteroencoding is the bridge from perceptual information (SDRs) to
symbols, represented as SHRs. Similar SDRs, for example, encodings of
nearby color hues, are intrinsically mapped to the same SHR, subject to
the memory's pattern matching threshold. This mechanism functions as an
unsupervised classifier or clustering algorithm for sets.

Traditional neurosymbolic architectures require complex pipelines to
translate continuous dense embeddings into discrete tokens. The
heteroencoder does this using only sparse representations. This natively
grounds symbols to physical reality.

Apart from mapping SDRs to SHRs, heteroencoding can be used to transcode
symbols between spaces with different dimensions (\(N\)) or populations
(\(P\)). This enables the transfer of SHRs between circuits that operate
under distinct hyperparameter constraints, preserving their semantic
meaning while converting their mathematical footprint.

Applied to a partitioned set where SHRs are separated into
non-overlapping blocks, heteroencoding can consolidate such arrays into
new, more compact symbolic representations.

\section{Circuits}\label{circuits}

Intelligent systems, whether natural or synthetic, emerge not from
monolithic scaling, but from the integration of discrete functional
components. Even complex cognitive functions like episodic or working
memory may arise from the interaction of just a few associative memory
instances. We model such architectures by assembling specialized units,
each performing a specific associative function.

A topological memory unit may consist of thousands of neurons, forming
millions of connections within their vast hidden layer. But rather than
individually modeling these neurons, we encapsulate their high-level
aggregate function: the storage and associative retrieval of population
codes. This abstracts away the messy details of individual neural
interactions as well as non-essential details of their connectome,
leaving us with an idealized topological blueprint. We work with a
digital approximation that captures the core algorithms of intelligence
and allows cognitive computation at scale.

Different types of memory components can be created on the basis of the
core \hyperref[the-core-algorithm]{topological memory algorithm}. There
are a variety of auto-associative memory types, with either reactive or
generative behavior (depending on their recurrent update mechanism),
storing individual patterns or symbols, or composite data such as
hypergraphs. Hetero-associative memories differ in how they acquire new
information: via supervised learning, predictive coding, or unsupervised
heteroencoding. Memory units can be configured for storing perceptual
data (SDRs) or symbolic information (SHRs), or for translating
information between these domains.

Besides memory components, there are various signal flow components that
do not embed an associative memory. These control the flow of neural
codes through delays, limit throughput, stochastically decimate neural
populations, or integrate data over time. In particular, performing a
\hyperref[k-winners-take-all]{majority vote} among competing
representations is a prominent function not associated with learning.

Information flows between system components via pathways: wide
connections that transport entire population codes in parallel. Each
lane within a pathway represents a single element within a
hyperdimensional space --- meaning a single pathway may consist of
thousands of parallel, one-way lanes. Overall traffic on a pathway is
consistently low at all times: according to our principle of sparse
coding, only one percent or less of the capacity is used.

Assembling these distinct components visually and functionally resembles
electronics --- though instead of routing single bits or scalar
currents, these systems route hyperdimensional sparse population codes.
This analogy with electronic circuits should not come by surprise, as
neural systems ultimately communicate via electrical pulses. For this
reason --- and to explicitly avoid confusion with conventional neural
nets or spiking neural networks (SNNs) --- we term our topological
memory networks \emph{circuits}.

\begin{quote}
\textbf{Circuits}\\
Functional modules composed of topological memory components and
components that control signal flow, connected by pathways which
transport population codes as discrete pulses.
\end{quote}

Designing special-purpose, functional modules is a stepping stone on the
path towards creating intelligent systems. Taking the biological
blueprint as a guide, we can design building blocks that handle specific
cognitive tasks. These include self-contained models of working memory,
episodic memory, pattern separation and completion, sequence learning
and generation, and hierarchical temporal processing.

Following this approach, known as ``synthetic neuroanatomy'', we
integrate functional modules into system-level architectures ---
creating intelligence that may simulate human cognition, or solve
non-human, technical challenges.

\subsection{Dataflow dynamics}\label{dataflow-dynamics}

The execution within topological memory circuits is driven by the flow
of data rather than a sequence of control instructions. There is no
instruction set --- a radical departure from von Neumann architectures.

Here, the program is coded as a dataflow graph where the nodes (the
circuit components) represent independent actors, and the directed edges
(the pathways) represent the flow of data between those actors. The
framework allows entire subgraphs to be logically encapsulated as nodes
without incurring runtime overhead. To facilitate the design of nested
and hierarchical architectures, this encapsulation is best applied to
cohesive subnets that expose only a few, clearly defined inputs and
outputs.

Consistent with our \emph{middle-up} approach to cognitive computation,
these dataflow graphs are in the realm of macro-cognitive computing
architectures. They model the computational principles of intelligent
systems (recurrence, sparsity, distributed representations) using
high-level software abstractions. The actors within dataflow graphs
capture the collective simultaneous behavior of thousands of cells,
rather than simulating individual neurons. We directly prototype
high-level cognitive functions via purposely-designed, explainable
circuits which are inspired by biological blueprints.

The data payload passed by the edges of the dataflow graph consists of
discrete sets. Typically, these are either sparse distributed
representations (SDRs representing perceptual data) or sparse
holographic representations (SHRs representing symbols and data
structures). The system also supports external input in the form of
dense sets and provides the necessary sparsification functions
(\hyperref[k-winners-take-all]{kWTA} or sparse hashing).

Because memory storage and retrieval are discrete, indivisible
operations, our model relies on a structured, discrete abstraction of
time. Although memory components are typically activity-driven, the
recurrent loops and generative components depend on predictable clock
ticks or execution cycles to manage feedback. Time is a scheduled rather
than a physical parameter in this computational framework.

This choice of a synchronous dataflow model over an asynchronous spiking
model is not mandatory. Although it deviates from biological
plausibility, we chose it for the ease of software and hardware
implementations. Future versions of topological memory models may well
adopt purely event-driven, or hybrid paradigms.

Synchronous dataflow models typically implement edges as first-in,
first-out (FIFO) queues: an upstream node pushes its output token to the
tail of the queue, and the downstream node consumes the token from the
head of the queue. But for the purpose of our circuits, a simple
two-phase clocking mechanism combined with a separation of input and
output buffers is sufficient.

Compute cycles are governed by two-phase clocking: a tick-tock mechanism
alternates between a read phase and a compute/write phase within each
cycle. This ensures that no computation overwrites data mid-transit.

The first global signal triggers data transfer. Every node copies the
stable output state of its upstream neighbors into its local input
buffer. In cases where pathways fork, the upstream output is broadcast,
duplicating the data into multiple downstream input buffers. Once this
transfer is complete, the upstream output buffers are freed.

A second global signal triggers the execution. Operating in parallel,
every node processes the contents of its input buffer and updates its
output buffer with the result of its computation.

Most components discussed here trigger a computation only when their
input token meets a threshold (generative auto-associative memories
being the notable exception). In the absence of this input, such
activity-driven components remain idle, expending almost no
computational power. This is a feature shared with typical neuromorphic
systems. Energy efficiency is also driven by the sparsity of
computation: topological memory components activate only a small
fraction of their vast internal circuitry when storing or retrieving
data.

Recurrent architectures, once activated by an external input, can
sustain output generation over multiple cycles without further inputs.
Depending on the characteristics of the processed data, generated
temporal sequences may die out, converge into repeating patterns, or
continue running without repetition due to stochastic subsampling
occurring in every cycle. At every cycle, external input may influence
or reset generation.

The nodes within a dataflow graph generally maintain a dynamic state
between execution cycles. For instance, a node may integrate pulses over
time, acting like an electric capacitor. Similarly, auto-associative
memory components maintain a dynamic state by blending external input
with internal recurrent feedback. Beyond their dynamic state, memory
components obviously persist memory traces within their topology.

In essence, topological memory circuits operate as fully stateful,
clock-driven machines that process streams of perceptual and symbolic
information.

\subsection{Schematics and components}\label{schematics-and-components}

We visualize topological memory circuits as schematics, similar to the
standard circuit diagrams used in electronics. Specific symbols indicate
the different component types, and arrows represent the connections
between the elements. Dataflow is always directed and governed by the
two-phase \hyperref[dataflow-dynamics]{clocking dynamics} discussed
above.

Once the library of components and their parameterizations is
standardized, system design ultimately reduces to editing human-readable
dataflow specifications, such as the JSON schema included in this
software framework. Alternatively, visual drawing tools may be used to
wire together and run prototype circuits. This format is also conducive
to generative algorithms that scan the space of possible circuits in
search of intelligent behaviors.

A necessary step towards this vision of low-code AGI development is the
standardization of circuit components and data transport mechanisms. The
following is a high-level overview of this framework. A detailed
documentation of the programming interface is available at this
project's companion repository.

In circuit schematics, we depict topological associative memory units as
square symbols:

\begin{figure}[H]
\centering
\includegraphics[width=0.367\linewidth,height=\textheight,keepaspectratio,alt={Generic schematic symbol of a topological associative memory component.}]{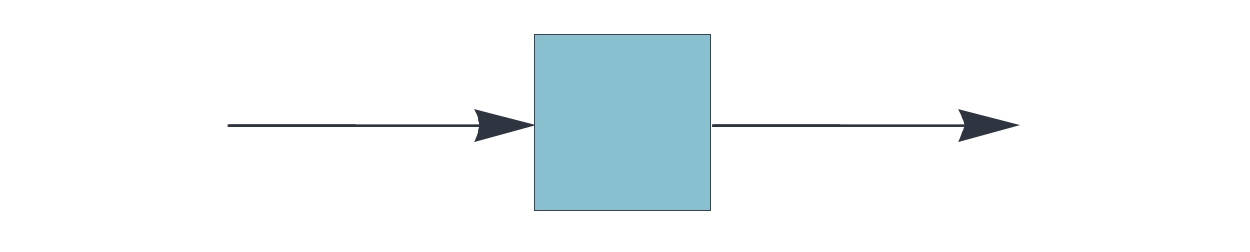}
\caption{Generic schematic symbol of a topological associative memory
component.}
\end{figure}

Signal flow components --- any node that does not encapsulate a memory
--- are shown as circles:

\begin{figure}[H]
\centering
\includegraphics[width=0.367\linewidth,height=\textheight,keepaspectratio,alt={Generic schematic symbol of a signal flow component.}]{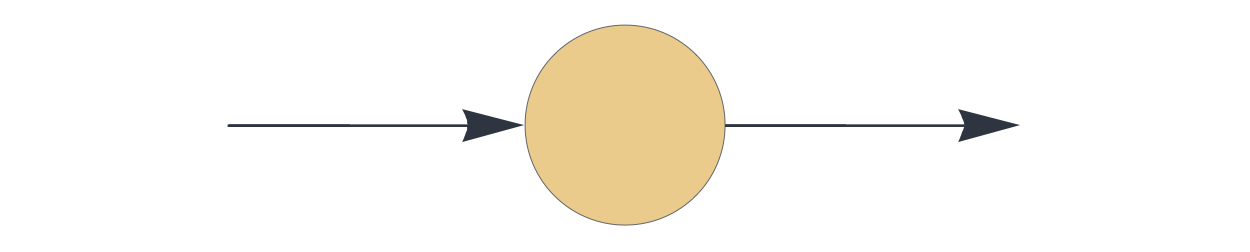}
\caption{Generic schematic symbol of a signal flow component.}
\end{figure}

Pathways, drawn as one-directional arrows, connect the circuit
components. At the \emph{tick} phase of each two-phase execution cycle,
data propagates down the pathways to the receiving components where it
is processed (the \emph{tock} phase). Therefore, a circuit component
delays the signal flow by one cycle.

Whereas circuit components generally maintain a state between evaluation
cycles, pathways are strictly stateless conduits. They do not remember
what signal flowed through them in the previous cycle. (Even the
temporal gating mechanism is based on an external clock, not the data
payload.)

We can modify pathways to transform data: apply a rate limit, a
population threshold, a temporal gate to support multiplexing, or a
permutation. In circuit diagrams, such transformations are indicated via
edge labels:

\begin{figure}[H]
\centering
\includegraphics[width=0.208\linewidth,height=\textheight,keepaspectratio,alt={A pathway on which a permutation is applied to the signal.}]{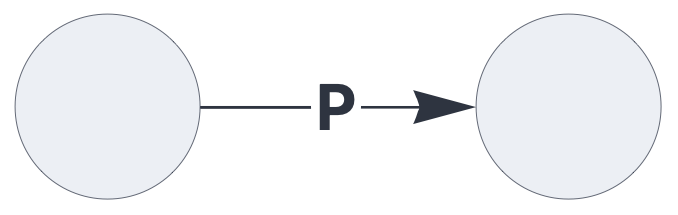}
\caption{A pathway on which a permutation is applied to the signal.}
\end{figure}

In this computational framework, pathways normally transport pure sets
(population codes), but can be extended to carry multisets. This is by
no means a departure from our fundamental premise of representing
cognitive information as sets: Multisets, as an \emph{analog} data
model, are formed naturally via temporal or spatial integration of
population codes and resolve to \emph{digital} sets through
\hyperref[k-winners-take-all]{kWTA} algorithms.

Allowing downstream components to dynamically suppress specific patterns
or mediate competing signals, multiset pathways may also convey
inhibitory signals. These are represented as the negative-valued
counterparts of regular, excitatory set elements.

Here is an example of a multiset that includes both excitatory and
inhibitory elements:

\begin{verbatim}
{-732, -641, -284, -86, 5, 5, 225, 235, 307, 307, 610}
\end{verbatim}

As excitatory and inhibitory elements cancel out, a multiset never
contains a positive element and its negative counterpart at the same
time.

\begin{figure}[H]
\centering
\includegraphics[width=0.208\linewidth,height=\textheight,keepaspectratio,alt={Multiset pathway are displayed as thicker, colored lines.}]{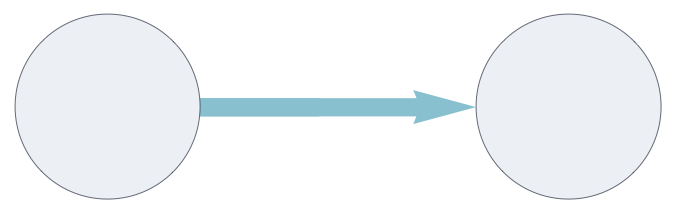}
\caption{Multiset pathway are displayed as thicker, colored lines.}
\end{figure}

Pathways do not alter the dimensionality of the transported data. In the
dataflow specification, this condition is enforced by assigning
hyperparameters (dimension \(N\) together with the reference population
\(P\)) not to the circuit components but to the pathways.

Circuit components can bundle separate pathways with non-overlapping
information. This realizes the disjoint union of sets, known as
\hyperref[block-coding]{block coding}. For instance, an auto-associative
memory component may handle a triple of incoming and a triple of
outgoing pathways, performing memory storage and retrieval on a single
set formed by concatenating the triples.

\begin{figure}[H]
\centering
\includegraphics[width=0.367\linewidth,height=\textheight,keepaspectratio,alt={A memory component that processes a triple pathway via block coding.}]{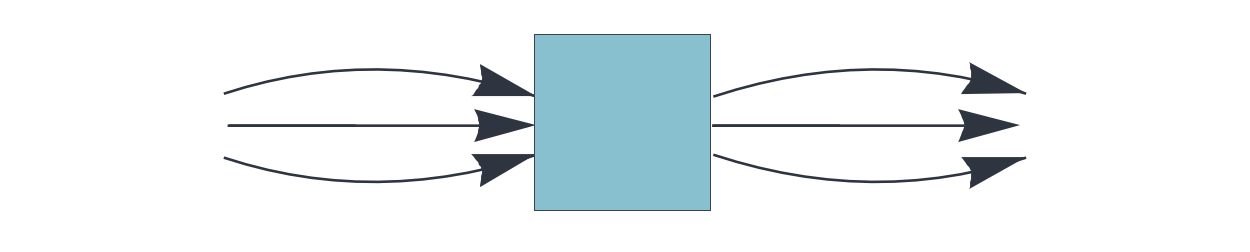}
\caption{A memory component that processes a triple pathway via block
coding.}
\end{figure}

When multiple pathways converge at a component's input, their signals
are aggregated into a multiset before processing. In this computational
framework, the pathway merging mechanism supports expressive control
flow management via logical gating applied to the connections. A pathway
may act as a mandatory prerequisite (blocking all input if empty), an
absolute veto (clearing all signals if active), a priority override, or
a fallback. With this design, any stateless routing functionality is
handled by the circuit's networking layer (\emph{white matter}), while
core associative memory functions --- including adaptive semantic
routing --- are encapsulated in memory components (\emph{gray matter}).

Input components represent the gateway for external data to enter a
circuit. There can be multiple input components, each representing one
block of data with distinct hyperparameters. The software framework
supports SDR/SHR encoders as well as preprocessing functions to be
plugged into input components.

\begin{figure}[H]
\centering
\includegraphics[width=0.22\linewidth,height=\textheight,keepaspectratio,alt={Schematic symbol of an input component.}]{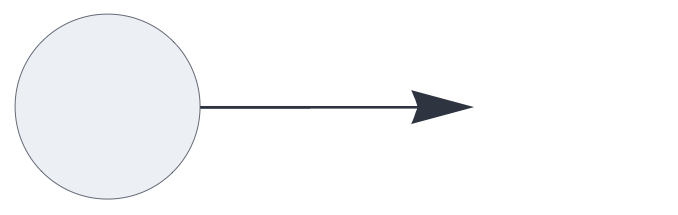}
\caption{Schematic symbol of an input component.}
\end{figure}

Output components form the circuit's outbound gateway. Each output line
corresponds to one block of data. Postprocessing and decoding functions
can be plugged into individual output components.

\begin{figure}[H]
\centering
\includegraphics[width=0.22\linewidth,height=\textheight,keepaspectratio,alt={Schematic symbol of an output component.}]{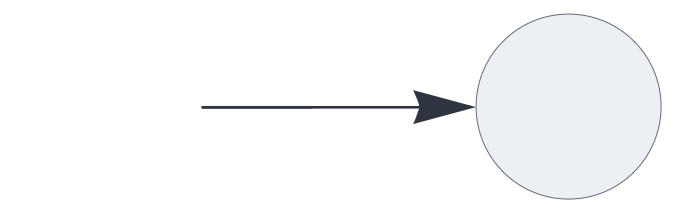}
\caption{Schematic symbol of an output component.}
\end{figure}

Unlike any other component types, inputs and outputs do not delay the
flow of data.

A circuit defines a stateful function \(f\) of \(m\) input blocks that
returns \(n\) output blocks.

\[ f: (A_1, \ldots, A_m) \to (B_1, \ldots, B_n)\]

Each function call corresponds to a circuit evaluation cycle in which
signals propagate downstream from one component to the next. Encoding,
decoding, preprocessing and postprocessing functionality does not
consume extra processing cycles. Note that the \(m\) input blocks may be
empty sets, or external data encoded into empty sets: in this case, the
circuit may still generate non-empty outputs through processing internal
feedback.

Each individual circuit component corresponds to a function \(f\) as
defined above. Consequently, we can express an entire circuit as a
logical component and use it as a modular building block in nested and
recursive architectures.

\begin{figure}[H]
\centering
\includegraphics[width=0.417\linewidth,height=\textheight,keepaspectratio,alt={A circuit, encapsulated as a modular circuit component.}]{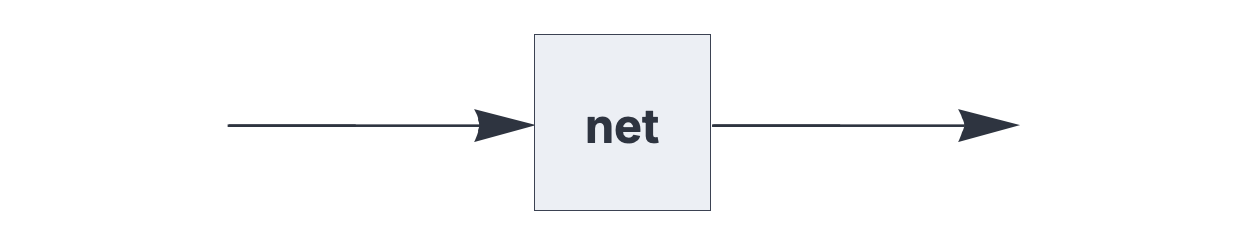}
\caption{A circuit, encapsulated as a modular circuit component.}
\end{figure}

We can set up an embedded circuit to operate at a different speed from
its enclosing circuit. For instance, a high-frequency perception
processor may run at a higher clock speed than a symbolic module at the
top of the cognitive hierarchy. The software framework supports such
multi-frequency configurations via temporal gating and multiplexing
options.

Delay components affect the signal flow in different ways. Apart from
delaying signals by one execution cycle, delay components may apply
stochastic subsampling through proportional decimation and rate
limiting. This functionality is analogous to the role of resistors and
constant-current diodes in electronic circuits. In addition, delay
components may be configured to temporally integrate signals up to a
maximum population --- the direct equivalent of an electronic capacitor.

Block coding, multisets and inhibitory signals are handled transparently
by delay components.

\begin{figure}[H]
\centering
\includegraphics[width=0.367\linewidth,height=\textheight,keepaspectratio,alt={Schematic symbol of a delay component.}]{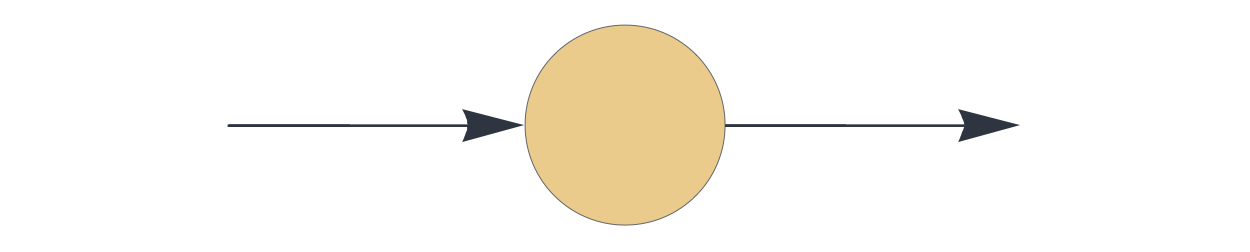}
\caption{Schematic symbol of a delay component.}
\end{figure}

In this framework, the simplest recurrent network consists of a delay
component with an external feedback line:

\begin{figure}[H]
\centering
\includegraphics[width=0.383\linewidth,height=\textheight,keepaspectratio,alt={A basic recurrent architecture.}]{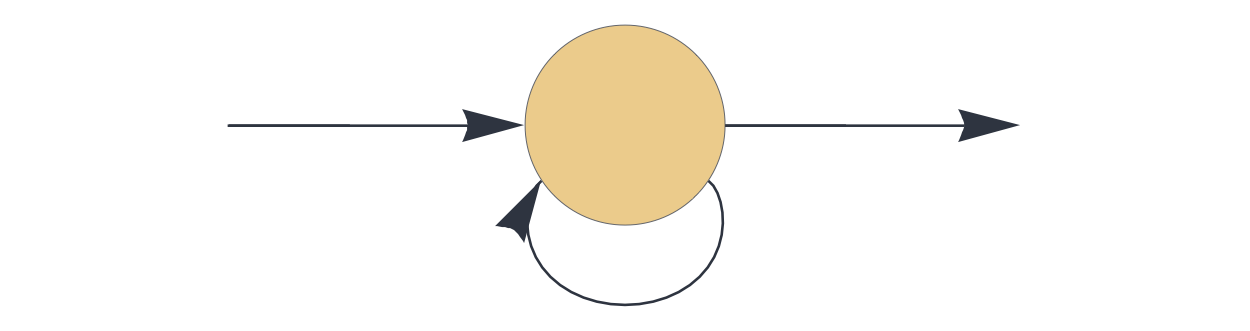}
\caption{A basic recurrent architecture.}
\end{figure}

Here the incoming signal is bundled with the delayed feedback. This
system requires stochastic subsampling at every step to prevent runaway
signal populations.

If this feedback system stops receiving inputs, it continues to emit the
last non-zero signal. A similar ``sample-and-hold'' function is
encapsulated in the \emph{latch} component, which repeats the last
signal for a number of cycles while receiving sub-threshold inputs.

\begin{figure}[H]
\centering
\includegraphics[width=0.383\linewidth,height=\textheight,keepaspectratio,alt={Schematic symbol of a latch component which holds and repeats the last above-threshold signal.}]{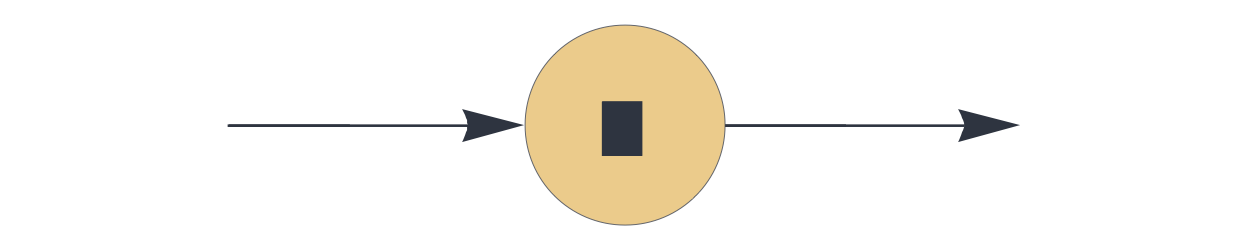}
\caption{Schematic symbol of a \emph{latch} component which holds and
repeats the last above-threshold signal.}
\end{figure}

A dedicated k-winners-take-all (\hyperref[k-winners-take-all]{kWTA})
component integrates incoming signals, bounding time (a sliding window
of execution cycles), space (a population capacity), or a combination of
both. At each cycle, the top-\(k\) elements of the multiset state are
calculated and emitted: the kWTA circuit component effectively functions
as a high-pass filter.

\begin{figure}[H]
\centering
\includegraphics[width=0.367\linewidth,height=\textheight,keepaspectratio,alt={A kWTA component calculates the top-k signal over a temporally aggregated multiset.}]{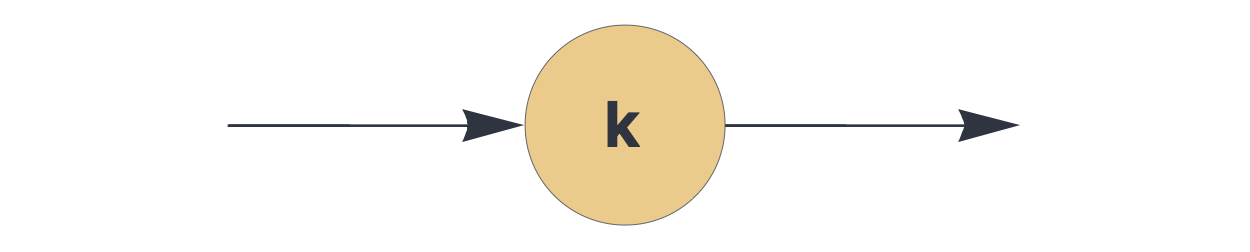}
\caption{A kWTA component calculates the top-\(k\) signal over a
temporally aggregated multiset.}
\end{figure}

Complementing stochastic subsampling as a model for ``leaky'' pathways
or the lossy memory components, we model additive noise through a
circuit component that emits a random pattern at every cycle. Like the
input component, the random noise component has no incoming pathway.

\begin{figure}[H]
\centering
\includegraphics[width=0.22\linewidth,height=\textheight,keepaspectratio,alt={Random noise generator.}]{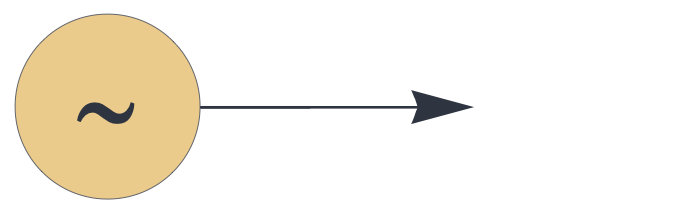}
\caption{Random noise generator.}
\end{figure}

Auto-associative circuit components incorporate a topological memory
instance with a shared input and output layer and an internal feedback
loop. Inputs and outputs have the same overall dimensions and can be
divided into non-overlapping partitions (block coding). As purely
``digital'' actors, memory components operate on proper sets, resolving
multiset and inhibitory inputs and concatenating partitioned inputs
before executing the core memory algorithm.

There is no external switch between training and inference modes.
Auto-associative memory learns new information on the fly, provided it
falls into a configurable population range. Memory retrieval can be
modulated through proportional stochastic decimation and a configurable
pattern matching threshold. Optional rate limits, capping the signal
size received per cycle, protect the memory algorithm from excessive
combinatorial activation.

The behavior of an auto-associative circuit component fundamentally
depends on its internal \hyperref[update-rules]{update mechanism}. Here
is an overview of the six quintessential types of autoassociations,
which use elementary set operations to update their input/output layer
with the pattern retrieved from memory. These form the basis for
purpose-built memory components with customized update and learning
dynamics.

\begin{figure}[H]
\centering
\includegraphics[width=0.367\linewidth,height=\textheight,keepaspectratio,alt={Auto-associative circuit component based on the replacement update rule. Typically used as item memory. Completes the query pattern while removing noise.}]{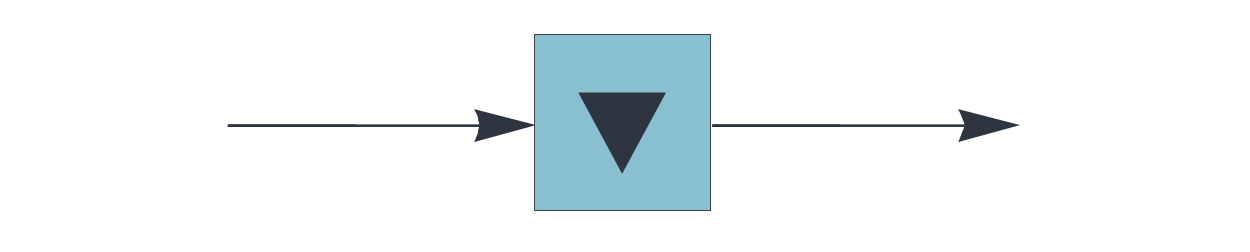}
\caption{Auto-associative circuit component based on the
\emph{replacement} update rule. Typically used as item memory. Completes
the query pattern while removing noise.}
\end{figure}

\begin{figure}[H]
\centering
\includegraphics[width=0.367\linewidth,height=\textheight,keepaspectratio,alt={A memory component based on the residual update rule uses inhibitory feedback, removing the retrieved pattern from the query pattern and retaining only the novel part of the input.}]{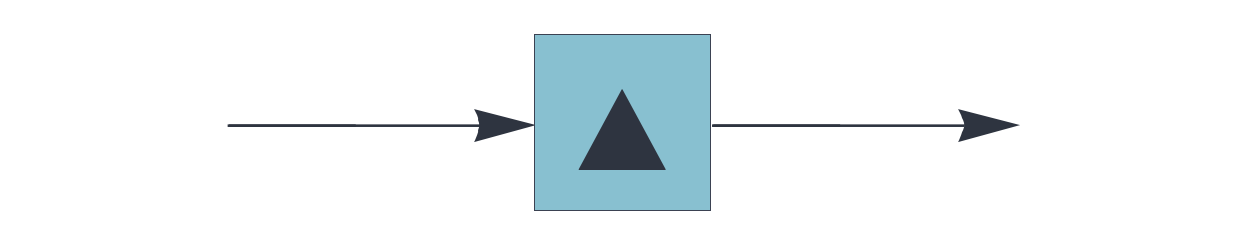}
\caption{A memory component based on the \emph{residual} update rule
uses inhibitory feedback, removing the retrieved pattern from the query
pattern and retaining only the novel part of the input.}
\end{figure}

\begin{figure}[H]
\centering
\includegraphics[width=0.367\linewidth,height=\textheight,keepaspectratio,alt={Auto-associative memory using the augmentation update rule forms the union of the query and the retrieved pattern.}]{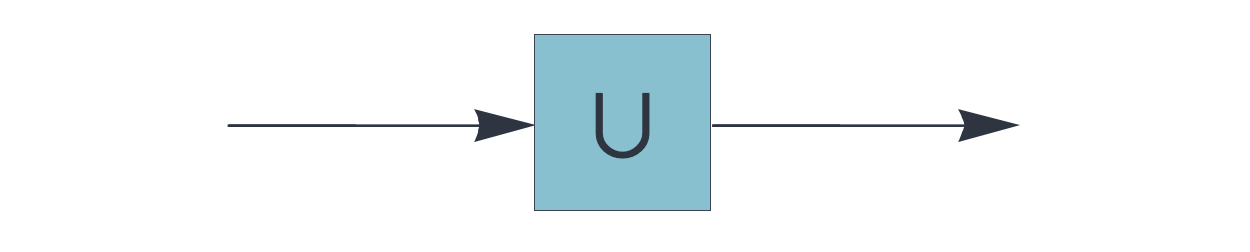}
\caption{Auto-associative memory using the \emph{augmentation} update
rule forms the union of the query and the retrieved pattern.}
\end{figure}

\begin{figure}[H]
\centering
\includegraphics[width=0.367\linewidth,height=\textheight,keepaspectratio,alt={The coincidence update rule retains the matching subset of the query pattern by computing the intersection with the retrieved memory.}]{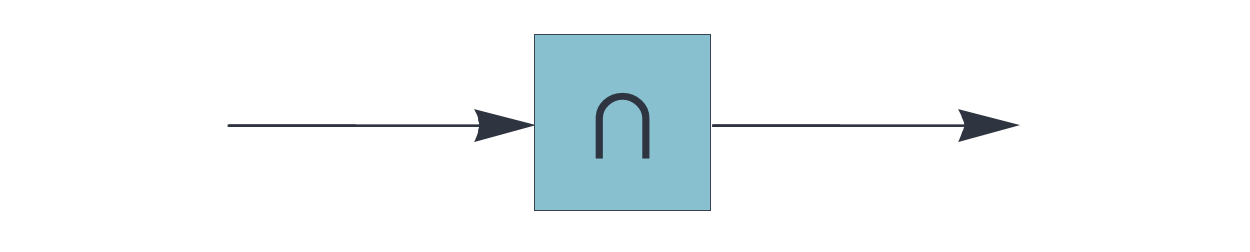}
\caption{The \emph{coincidence} update rule retains the matching subset
of the query pattern by computing the intersection with the retrieved
memory.}
\end{figure}

\begin{figure}[H]
\centering
\includegraphics[width=0.367\linewidth,height=\textheight,keepaspectratio,alt={Auto-associative memory based on the complement update rule returns the retrieved pattern after removing the query elements. Generative behavior emerges from this circuit component when its output is routed back to the input.}]{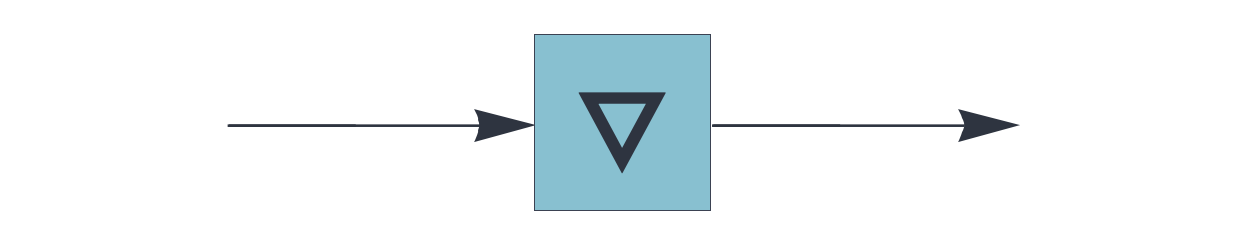}
\caption{Auto-associative memory based on the \emph{complement} update
rule returns the retrieved pattern after removing the query elements.
Generative behavior emerges from this circuit component when its output
is routed back to the input.}
\end{figure}

\begin{figure}[H]
\centering
\includegraphics[width=0.367\linewidth,height=\textheight,keepaspectratio,alt={The difference update rule retains the symmetric difference between the query and the retrieved pattern. This gives rise to generative functionality similar to the complement update rule.}]{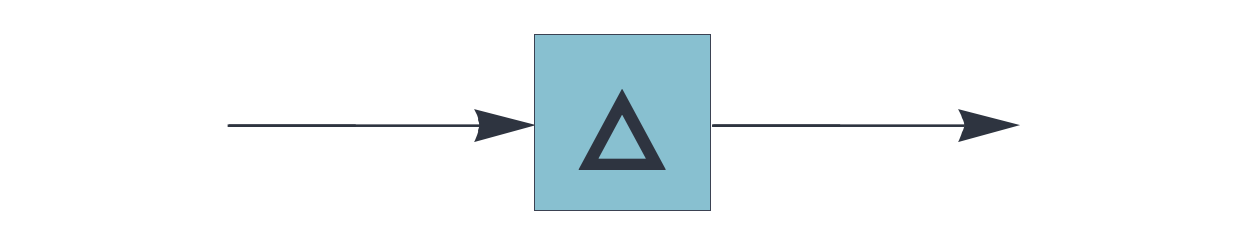}
\caption{The \emph{difference} update rule retains the symmetric
difference between the query and the retrieved pattern. This gives rise
to generative functionality similar to the \emph{complement} update
rule.}
\end{figure}

Hetero-associative circuit components can be categorized according to
their learning dynamics. Supervised learning is realized either
spatially by providing the teaching signal simultaneously with the
context signal, or temporally through predictive learning.

The context signal can be partitioned into blocks and carry multiset
patterns including inhibitory elements. Memory components systematically
resolve such inputs to proper sets.

\begin{figure}[H]
\centering
\includegraphics[width=0.367\linewidth,height=\textheight,keepaspectratio,alt={An associator circuit component that learns and retrieves heteroassociations of a context signal with a teaching signal.}]{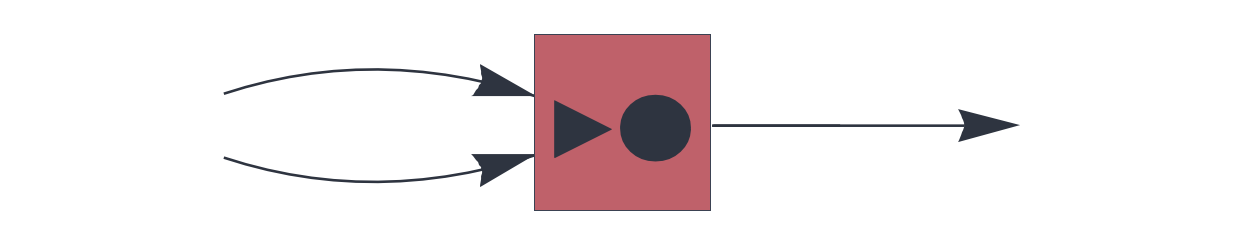}
\caption{An \emph{associator} circuit component that learns and
retrieves heteroassociations of a context signal with a teaching
signal.}
\end{figure}

\begin{figure}[H]
\centering
\includegraphics[width=0.367\linewidth,height=\textheight,keepaspectratio,alt={A predictor circuit component predicts the upcoming teaching signal based on the current context signal, realizing temporal hetero-associative learning.}]{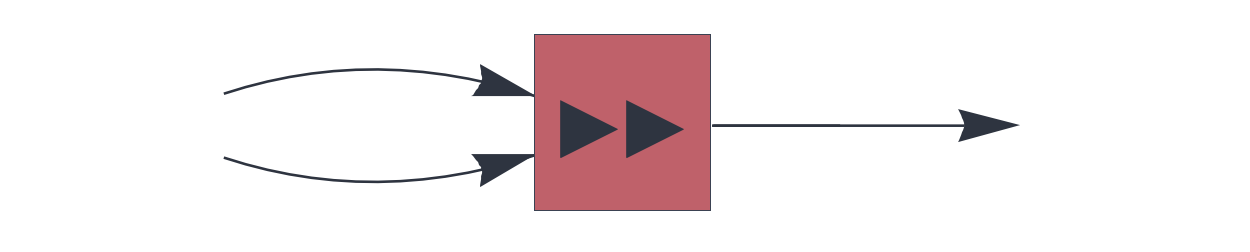}
\caption{A \emph{predictor} circuit component predicts the upcoming
teaching signal based on the current context signal, realizing temporal
hetero-associative learning.}
\end{figure}

Unsupervised hetero-associative learning requires no teaching signal.
Instead, it generates and learns a random pattern representing each
cluster of similar inputs.

\begin{figure}[H]
\centering
\includegraphics[width=0.367\linewidth,height=\textheight,keepaspectratio,alt={A heteroencoder circuit components works as a classifier, categorizing input patterns and generating a unique token for each detected cluster.}]{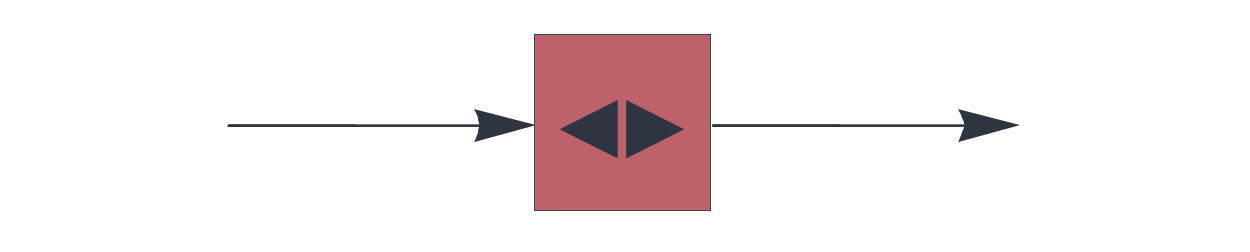}
\caption{A \emph{heteroencoder} circuit components works as a
classifier, categorizing input patterns and generating a unique token
for each detected cluster.}
\end{figure}

\section{Recurrence and Temporal
Memory}\label{recurrence-and-temporal-memory}

\subsection{Recurrence}\label{recurrence}

As a defining property of cognitive architectures, recurrence is
characterized by cyclic connections in the circuit graph, integrating
data across synchronized clock cycles through feedback loops. Any such
feedback system requires a sparsity-controlling mechanism to prevent
runaway feedback. This computational framework supports multiple methods
to achieve this: stochastic rate limiting (located in the data transport
logic or within delay nodes), k-winners-take-all ranking (applied in
spatial or temporal data accumulation), signal flow logic, and specific
update rules that govern the auto-associative memory components.

The following circuit schematic shows the most basic recurrent circuit,
consisting of a single delay node with direct feedback from its output
to its input state. This mechanism merges the two incoming pathways into
a multiset and subsequently applies stochastic rate limiting to the
merged data.

\begin{figure}[H]
\centering
\includegraphics[width=0.383\linewidth,height=\textheight,keepaspectratio,alt={A basic recurrent circuit. Rate limiting is applied by the circuit component.}]{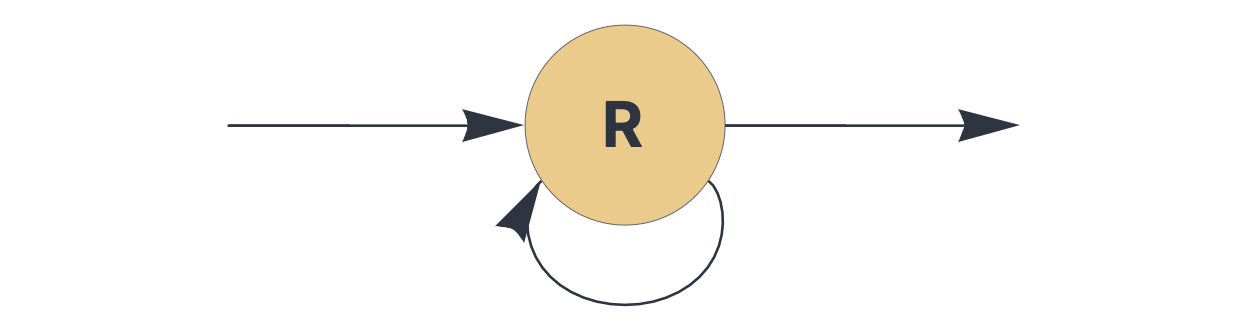}
\caption{A basic recurrent circuit. Rate limiting is applied by the
circuit component.}
\end{figure}

When processing a data stream, the state of this system is dominated by
the most recent inputs, while older signals gradually fade.

We can instead limit the feedback signal \emph{prior} to merging it with
the incoming data stream. This results in a subtly different behavior
compared with the above circuit: the incoming signal is fully manifest
in the component's state for one cycle before it fades through the
feedback decimation.

\begin{figure}[H]
\centering
\includegraphics[width=0.383\linewidth,height=\textheight,keepaspectratio,alt={A basic recurrent circuit with rate-limited feedback, indicated by a pathway labeled R.}]{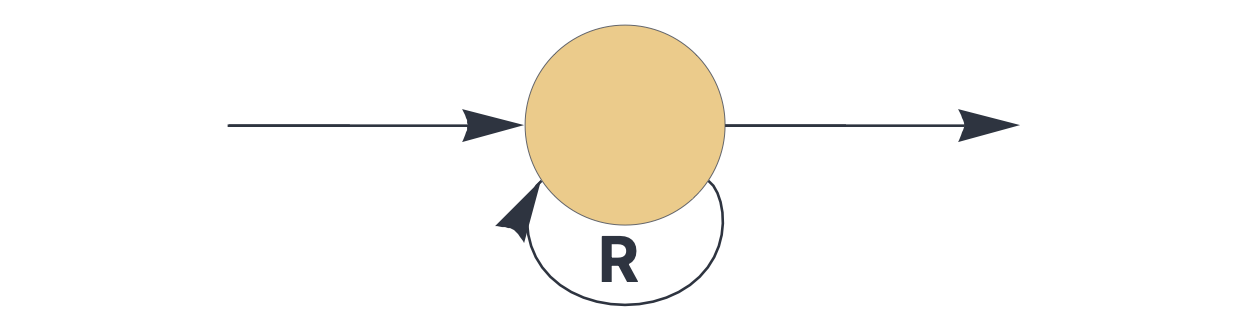}
\caption{A basic recurrent circuit with rate-limited feedback, indicated
by a pathway labeled \emph{R}.}
\end{figure}

Applying a permutation to the feedback encodes each signal's temporal
position within the input sequence.

\begin{figure}[H]
\centering
\includegraphics[width=0.383\linewidth,height=\textheight,keepaspectratio,alt={Recurrent architecture with a permuted feedback line, indicated by a pathway labeled P.}]{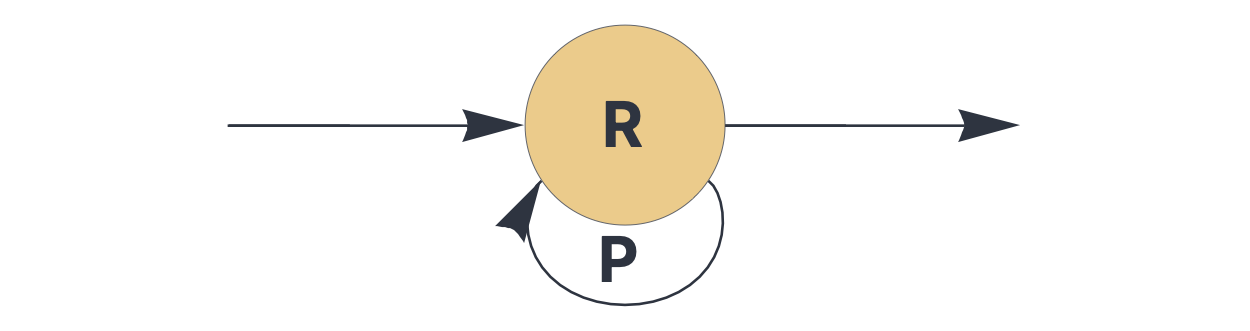}
\caption{Recurrent architecture with a permuted feedback line, indicated
by a pathway labeled \emph{P}.}
\end{figure}

Here is a recurrent circuit formed by two delay components. The feedback
signal is permuted and delayed by one execution cycle.

\begin{figure}[H]
\centering
\includegraphics[width=0.383\linewidth,height=\textheight,keepaspectratio,alt={Recurrent architecture with delayed feedback.}]{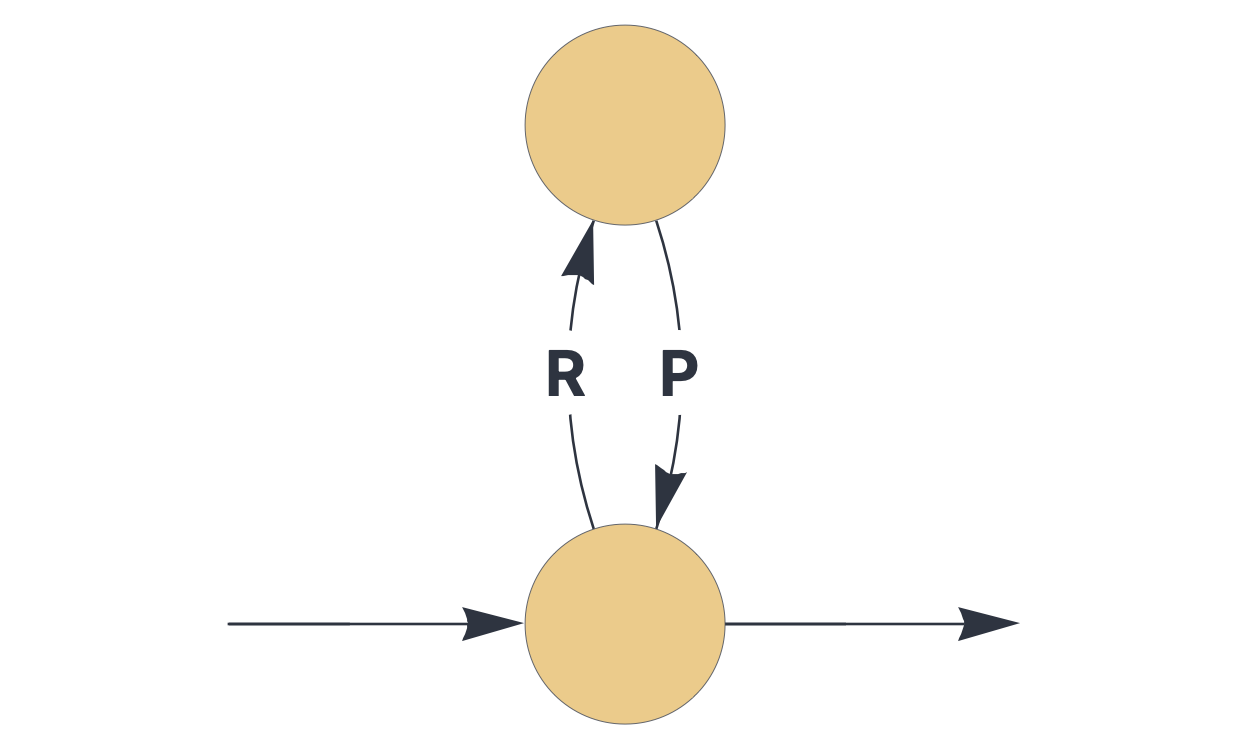}
\caption{Recurrent architecture with delayed feedback.}
\end{figure}

\subsection{Reverberatory circuits}\label{reverberatory-circuits}

Modeling dynamic circuits that hold data without leaking information at
each cycle requires a sparsity-controlling mechanism beyond the
stochastic subsampling discussed above. Priority gating provides such an
alternative: when merging a forward signal with a feedback line,
feedback is suppressed unless the forward signal is empty --- an
equivalent to biological feedforward inhibition. This model is shown in
the following circuit diagram.

\begin{figure}[H]
\centering
\includegraphics[width=0.383\linewidth,height=\textheight,keepaspectratio,alt={Elementary reverberatory circuit. The forward signal (tagged with an exclamation mark) takes priority over the feedback.}]{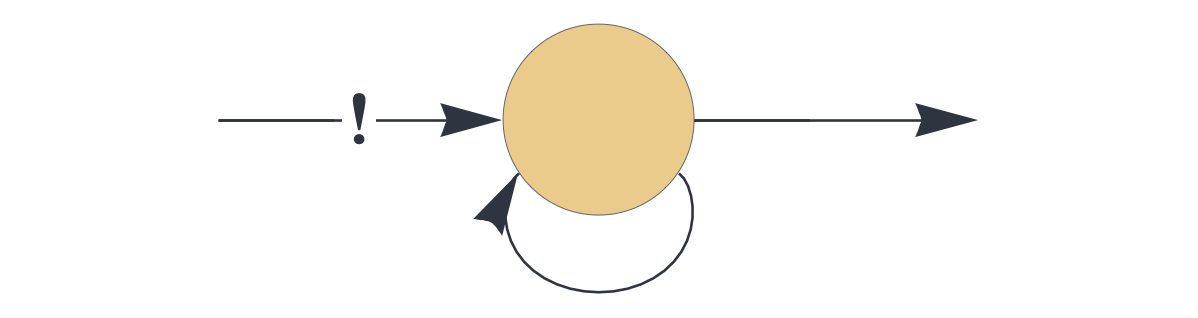}
\caption{Elementary reverberatory circuit. The forward signal (tagged
with an exclamation mark) takes priority over the feedback.}
\end{figure}

This circuit samples the input and holds it without loss of information
until new input replaces the system's state.

The following diagram shows a reverberatory circuit where the feedback
is delayed by one cycle. In this system, signals bounce between two
delay nodes, toggling two states in the absence of new input. Again, the
incoming pathway takes priority over the feedback line.

\begin{figure}[H]
\centering
\includegraphics[width=0.383\linewidth,height=\textheight,keepaspectratio,alt={A circuit with delayed reverberation, bouncing two stable states between delay nodes.}]{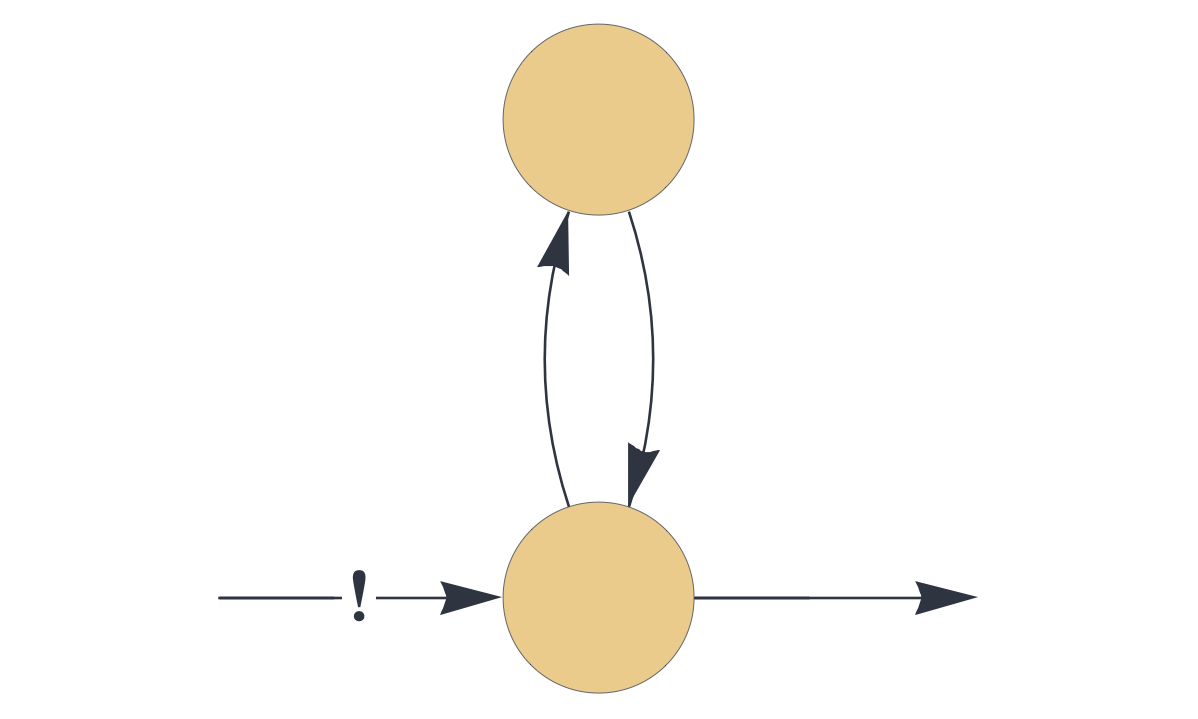}
\caption{A circuit with delayed reverberation, bouncing two stable
states between delay nodes.}
\end{figure}

A reverberatory circuit feeding into an item memory decomposes bundles
into their constituents. An inhibitory feedback line removes the
extracted items, one after another, from the recurrent subnet. The
circuit outputs noise-free tokens in random order, interspersed with
empty representations reflecting cycles where memory retrieval failed
due to pattern matching ties.

\begin{figure}[H]
\centering
\includegraphics[width=0.383\linewidth,height=\textheight,keepaspectratio,alt={A reverberatory circuit combined with an item memory extracts items from a bundle.}]{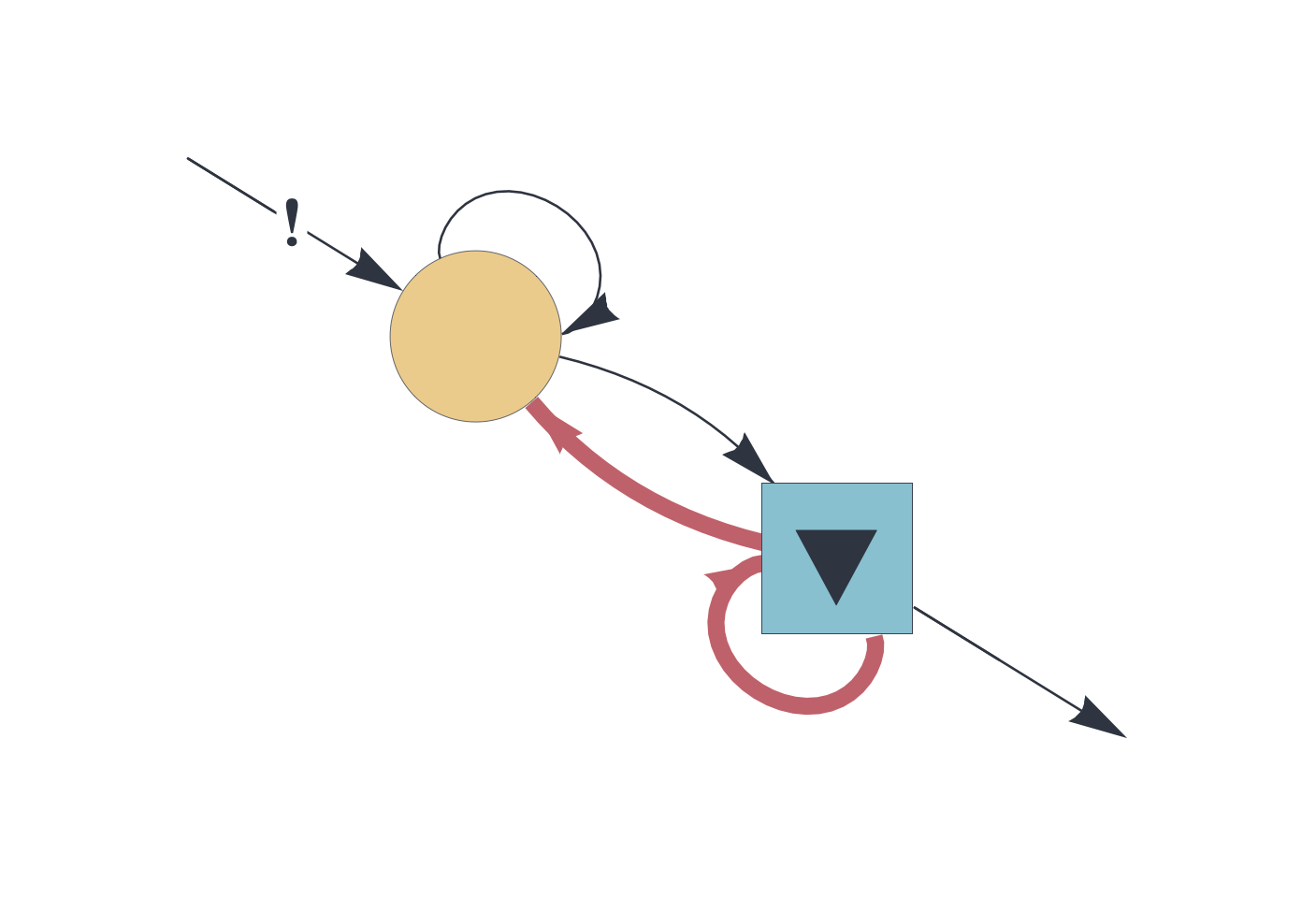}
\caption{A reverberatory circuit combined with an item memory extracts
items from a bundle.}
\end{figure}

\subsection{Reservoir architectures}\label{reservoir-architectures}

Temporal memory emerges from the combination of recurrent architectures
and topological associative memory. A system that learns sequences on
the fly while predicting the next token can be implemented via a
reservoir architecture. Here, the reservoir is a recurrent network with
permuted feedback, building up a state that represents the recent inputs
as an ordered sequence. The most recent input pattern dominates the
state, while older inputs fade due to recursive stochastic subsampling.
A \hyperref[predictive-learning]{hetero-associative predictor} node
reads out the reservoir's state at every cycle and learns its
association with the subsequent input.

The following figure shows a basic temporal sequence memory with a
single-delay reservoir:

\begin{figure}[H]
\centering
\includegraphics[width=0.433\linewidth,height=\textheight,keepaspectratio,alt={Reservoir architecture, representing a temporal sequence memory.}]{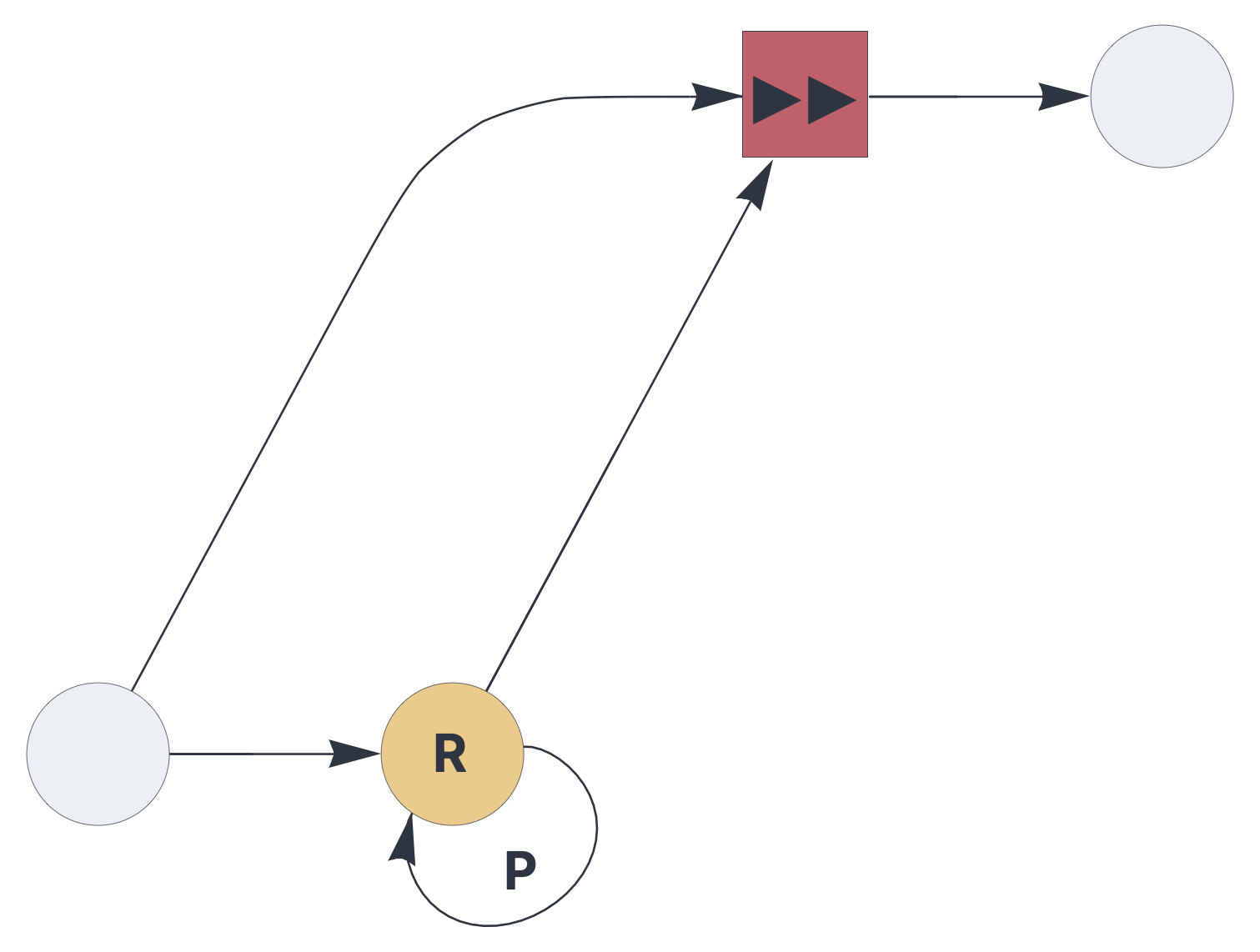}
\caption{Reservoir architecture, representing a temporal sequence
memory.}
\end{figure}

Taking this basic design as a starting point, we now build a more
advanced reservoir architecture, shown in the following circuit diagram.

\begin{figure}[H]
\centering
\includegraphics[width=0.833\linewidth,height=\textheight,keepaspectratio,alt={A reservoir architecture incorporating bigram encoding and a dual-delay reservoir.}]{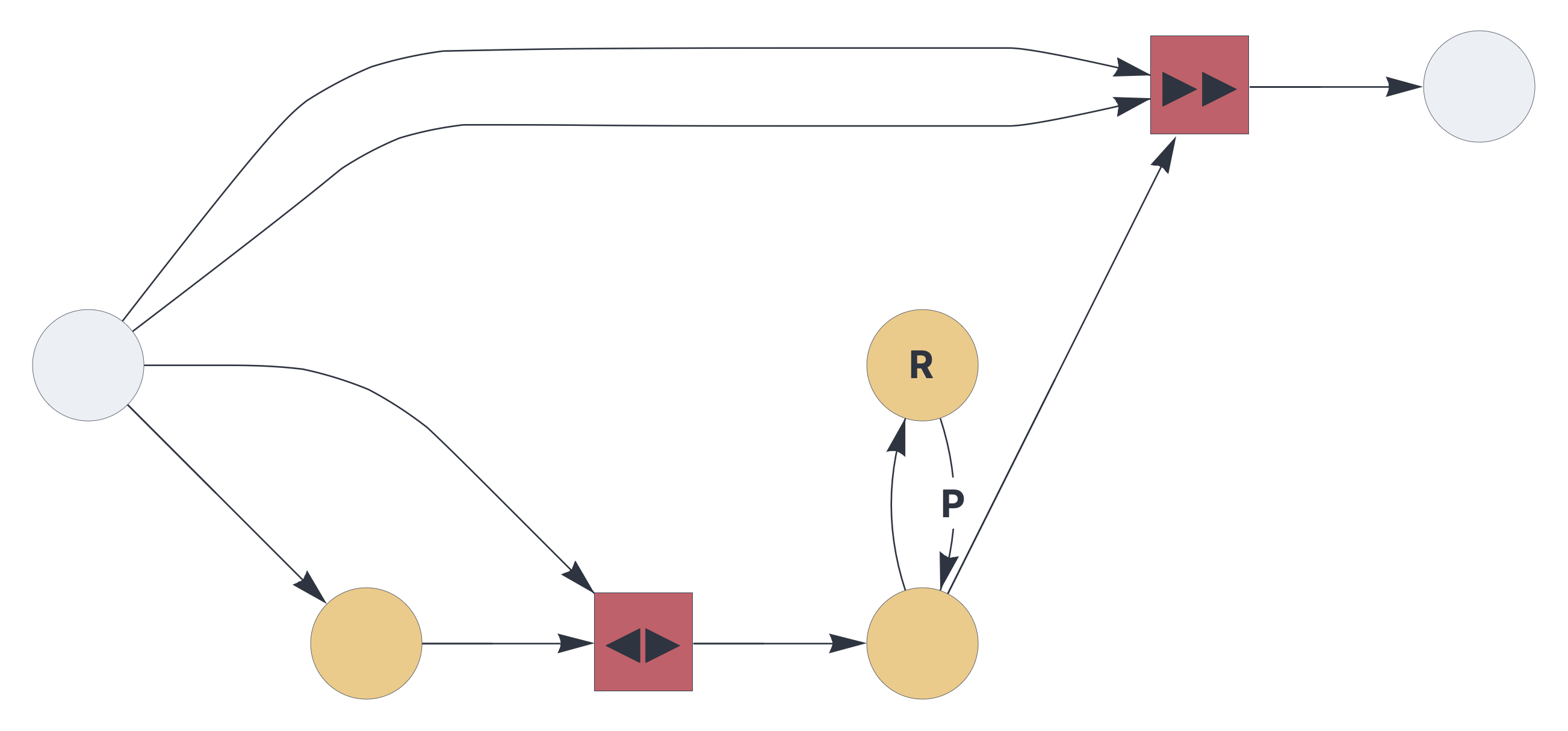}
\caption{A reservoir architecture incorporating bigram encoding and a
dual-delay reservoir.}
\end{figure}

Here, the reservoir at the center of the circuit is formed by a delayed
feedback loop. This increases the reservoir's temporal depth, allowing
it to hold information over a longer time window.

Instead of filling the reservoir with raw incoming tokens, this circuit
transcribes each incoming token pair into a bigram. The bigram encoding
mechanism shown here uses a \hyperref[heteroencoder]{heteroencoder}
memory component that generates a new sparse holographic representation
(SHR) for every unique token pair. Bigram encoding squares the
representational space, vastly increasing the circuit's learning
capacity. For example, a data stream consisting of digits 0 to 9 (a
10-token lexicon) expands to 100 different bigram tokens, activating a
much larger section of the topological memory's latent space.

In this architecture, the readout node associates predictions not only
with the reservoir's state, but also directly with the current input.
This shortcut improves the reaction time of the learning mechanism.

The circuit as shown above is capable of learning long sequences, such
as random strings consisting of thousands of digits.

\section{Perception and
Classification}\label{perception-and-classification}

\subsection{Classification of handwritten
digits}\label{classification-of-handwritten-digits}

Hetero-associative topological memory maps perceptual data, encoded as
sparse distributed representations (SDRs), directly to sparse
holographic representations (SHRs) that encode symbolic categories, or
classifications.

Unlike conventional neural networks, which achieve generalization by
compressing data through a continuous dimensional bottleneck during
training, this classifier learns in a purely additive fashion. It does
not extract or compress generalized features from the training data.
Instead, generalization takes place during memory retrieval, where the
nearest-neighbor search maps a continuum of noisy data to the correct
symbolic category.

This capability is easily demonstrated using the canonical challenge of
classifying handwritten digits.

\begin{figure}[H]
\centering
\includegraphics[width=0.667\linewidth,height=\textheight,keepaspectratio,alt={Sample digits from the MNIST Handwritten Digit Database}]{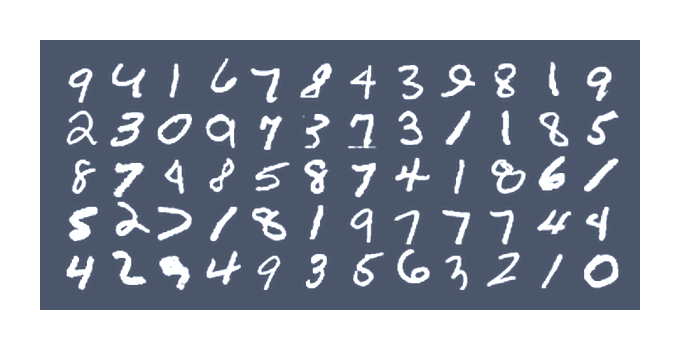}
\caption{Sample digits from the MNIST Handwritten Digit Database}
\end{figure}

The dataset (LeCun 1998) consists of 70,000 handwritten digits,
rasterized as \(28\times28\) pixel grayscale images.

We map the flattened image data, after applying a 1-pixel-radius
Gaussian blur, to SDRs via a random projection onto a hyperdimensional
space, followed by a kWTA selection of the top-ranked elements. The
random projection is achieved by multiplication with a sparse binary
\(784 \times N\) matrix with sparsity \(\sqrt{N}\). The SDR dimension
\(N\) must be large enough to capture the dataset's features.
Empirically, dimensions between 10,000 and 16,000 and a population
sparsity of 1\% work well.

The categories (digits 0 to 9) are encoded as non-overlapping SHRs with
populations \(P = 15\).

The following circuit diagram shows the hetero-associative memory
component receiving two inputs: the SDR-encoded image data and the
SHR-encoded categories. The associative memory learns when both inputs
are non-empty. If the category input is empty, the component retrieves a
value from memory.

\begin{figure}[H]
\centering
\includegraphics[width=0.417\linewidth,height=\textheight,keepaspectratio,alt={Hetero-associative memory component set up for supervised learning.}]{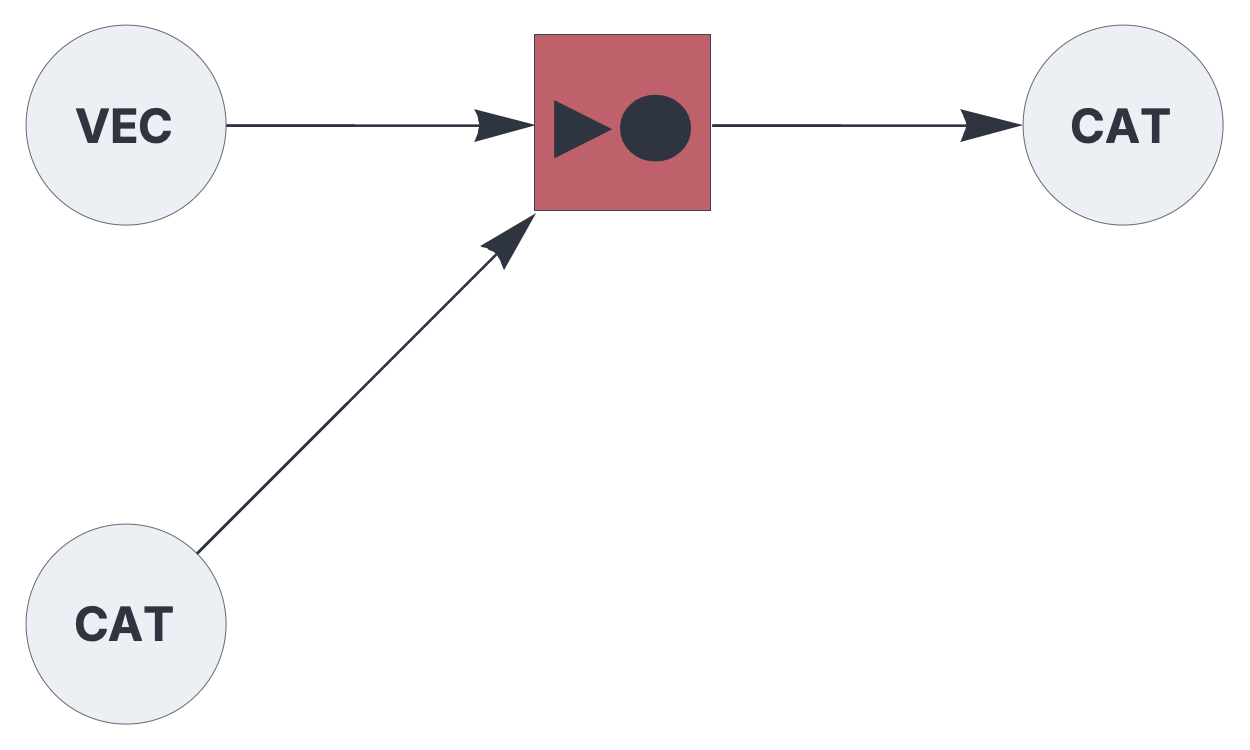}
\caption{Hetero-associative memory component set up for supervised
learning.}
\end{figure}

We train this classifier with 60,000 images in a single pass. Testing
with 10,000 images results in an accuracy score of about 96\%, out of
the box, without any optimization. This demonstrates that topological
associative memory is perfectly capable of handling machine learning
classification tasks.

Test scores can be improved by using an adaptive image encoder that
suppresses the ``center blob'' that dominates the images in this
dataset.

\section{Knowledge Representation}\label{knowledge-representation}

\subsection{Visual scene graphs}\label{visual-scene-graphs}

Our memory does not store raw noisy perceptual data. Instead, it
transforms perceptions into higher-level stable abstractions, encoded as
sparse holographic representations, or symbols. For instance, various
perceptions of the color red, differing in hue, brightness, or
saturation, are clustered into a symbolic category ``red''. Similarly,
we can encode an object's shape and relative size and position as stable
symbols.

The \hyperref[binding]{binding} mechanism, based on storing SHR bundles
in auto-associative memory, links features to objects and objects to an
entire scene graph. The memory component in the circuit shown below uses
the \emph{replacement} update rule, which completes patterns based on
incomplete information --- a simple cognitive model of \emph{seeing what
we know, not what we see}.

\begin{figure}[H]
\centering
\includegraphics[width=0.367\linewidth,height=\textheight,keepaspectratio,alt={Auto-associative memory component storing a visual scene graph.}]{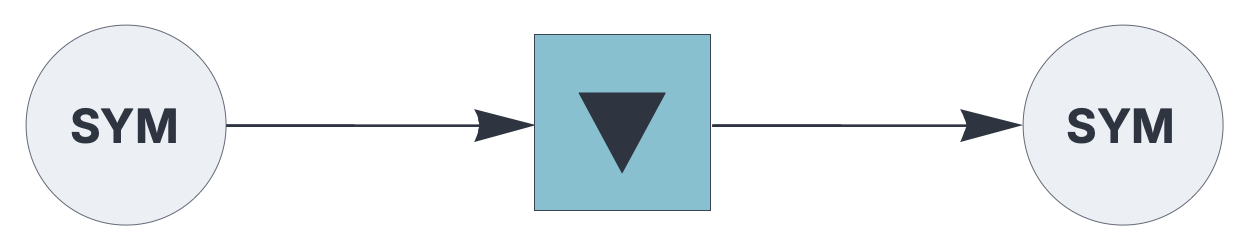}
\caption{Auto-associative memory component storing a visual scene
graph.}
\end{figure}

We demonstrate this mechanism using the following picture as an example.

\begin{figure}[H]
\centering
\includegraphics[width=0.3\linewidth,height=\textheight,keepaspectratio,alt={A scene with three colored features.}]{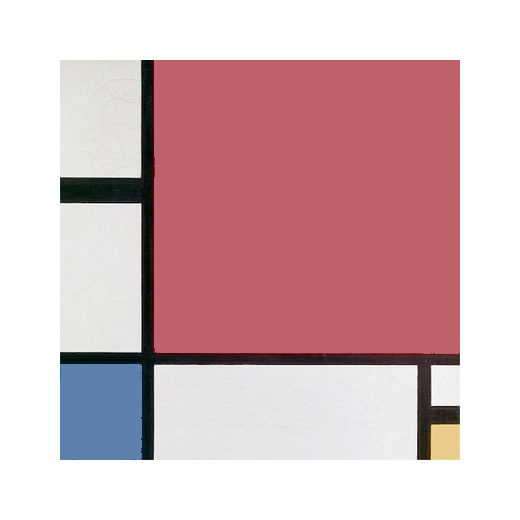}
\caption{A scene with three colored features.}
\end{figure}

The features in this image --- colors, sizes, relative positions --- are
encoded as symbolic SHRs. Storing the entire scene in memory involves
binding the features to three objects (encoded as latent symbols O1, O2,
O3), and binding these objects to a latent symbol that points to the
entire scene:

\begin{verbatim}
> O1 large red top right;
> O2 medium blue bottom left;
> O3 small yellow bottom right;
> Scene O1 O2 O3;
\end{verbatim}

Now we can query the scene graph top down:

\begin{verbatim}
> Scene
O1, O2, O3, Scene
> O1
O1, large, red, top, right, O2, O3, Scene
\end{verbatim}

Note that the query with O1 returns every directly connected\\
hypergraph node, not only its associated features as one might expect.

Queries based on specific features:

\begin{verbatim}
> red
O1, large, red, top, right
> small
right, bottom, O3, small, yellow
> bottom
right, O2, medium, blue, bottom, left, O3, small, yellow
> bottom right
right, bottom, O3, small, yellow
\end{verbatim}

When queried with ``bottom'', the system pulls all information from the
two objects at the bottom of the scene. A more specific query with the
bundle ``bottom'' \(\cup\) ``right'' narrows the result to object O3.

These examples show how memory queries, simulating partial perceptions,
complete the picture. Using an auto-associative memory with the
\emph{residual} instead of the \emph{replacement} update rule would turn
the system into a novelty filter: the queries shown above would return
empty results, but novel query patterns would pass through.

\subsection{Semantic similarity}\label{semantic-similarity}

In the context of AI and semantics, a symbol is a discrete unit (such as
a word or token) that represents a high-level human-readable object or
concept. Symbols are the building blocks for knowledge representation
and reasoning. A symbol acts as a pointer to something else: real-world
entities (such as ``apple''), abstract concepts (``wisdom''),
relationships (``is part of''), logical operations (if-then), linguistic
phrases or grammatical structures (verb-phrase), or mathematical symbols
and expressions (\(x + 1\)).

In models based on sparse hyperdimensional representations, different
symbols are encoded as sparse holographic representations (SHRs),
regardless of their underlying semantic similarity or differences. Like
symbols, SHRs serve as abstract and unique identifiers, pointing to
discrete items.

Contrary to common belief, semantically related symbols, such as ``cat''
and ``dog'', are not represented by sparse representations where the
overlapping elements denote shared features, such as ``fur'' and
``ears''. The number of features shared between cats and dogs far
exceeds the typical size of a sparse representation. For symbols to
function as pointers, their representations must be nearly orthogonal:
this ensures that subsampled versions point to the same thing, rather
than a subset of its features. The same principle also enables the
decomposition of bundles into distinct items.

Semantic similarity between symbols emerges from shared associations
within knowledge graphs, either through direct or spreading (cascading)
activation.

To explore this concept, we use a dataset of word associations that was
specifically developed for topological memory prototyping.

The dataset consists of roughly 1.9 million word associations,
representing an English speaker's mental lexicon. In this semantic
network, every word or concept is a node, and the associations between
them are the links. Here is a data sample, showing all nodes linked to
``freedom'':

\begin{verbatim}
freedom -> agency assembly autonomy capability choice consent emancipation
 equality escape exemption expression freewill impunity independence
 information liberation liberty license movement nonconformity openness
 opportunity press privilege release religion rights self-determination
 self-government self-rule sovereignty speech thought unfettered unhampered
 unrestricted volition
\end{verbatim}

This dataset includes paradigmatic as well as syntagmatic associations.
\emph{Paradigmatic} associations (for example ``freedom'' \(\to\)
``liberty'') are words that belong to the same grammatical class and
often share semantic features. These are driven by mental categorization
and synonymy. \emph{Syntagmatic} associations (for example ``freedom''
\(\to\) ``speech'') are words that frequently appear next to each other
in a sequence or sentence. Additionally, this dataset reflects
\emph{schematic} associations which are linked to the organization of
knowledge.

Cognitive information is normally represented via triples, not direct
symbol-to-symbol associations. In fact, the dataset represents triples
in disguise, omitting predicate information.

We can store symbolic associations directly in a hetero-associative
memory after encoding each word as a sparse holographic representation
(SHR). This schematic shows the corresponding circuit:

\begin{figure}[H]
\centering
\includegraphics[width=0.417\linewidth,height=\textheight,keepaspectratio,alt={Hetero-associative memory for symbolic information.}]{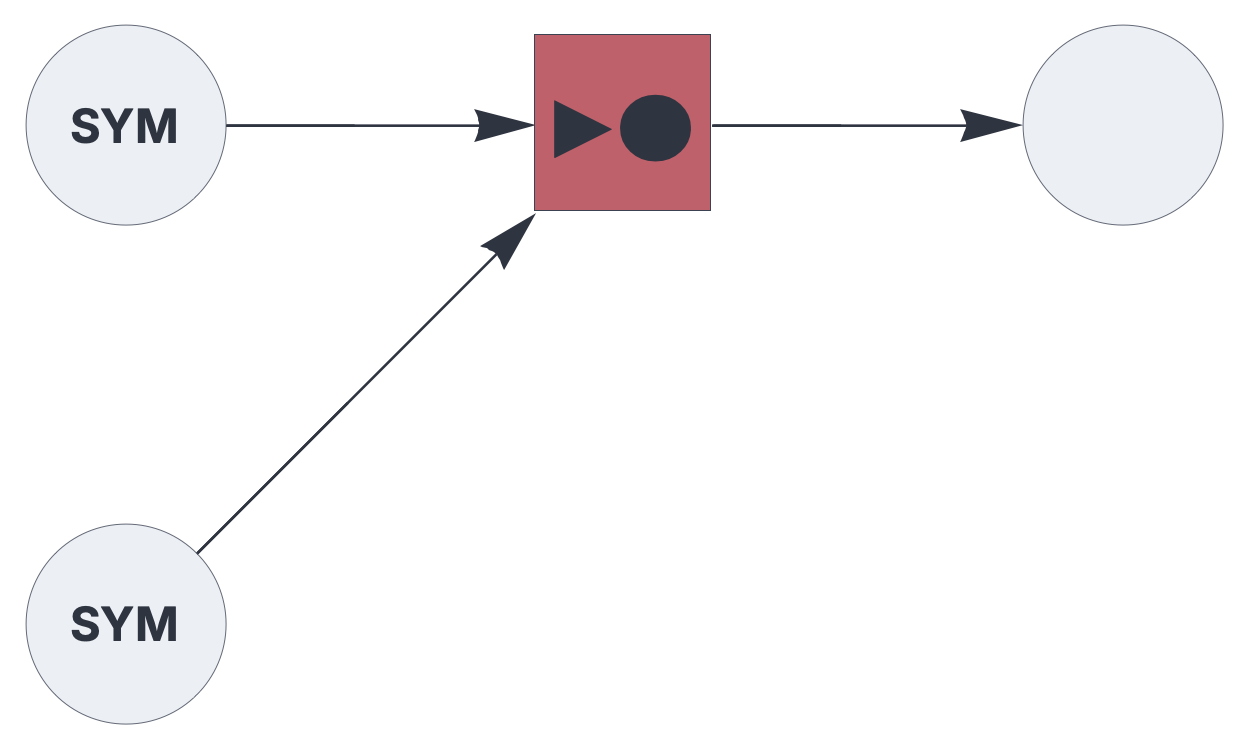}
\caption{Hetero-associative memory for symbolic information.}
\end{figure}

Storing 1.9 million associations requires a memory of dimension
\(N = 3,000\) if tokens are encoded with a population \(P = 10\).

The memory inherently merges associations that share the same key into a
single association. Retrieving such a one-to-many association typically
results in a non-sparse representation. For example, a query with the
word ``freedom'' returns a dense set with 350 active elements (a
sparsity of about 12\%), encoding the semantic field of the query term.

Dense representations are not suitable for direct associative memory
queries, such as iterative decomposition which is restricted to bundles
of approximately 10 tokens. Instead, we must reduce dense intermediate
activations for further processing using a
\hyperref[k-winners-take-all]{kWTA} mechanism. The similarity of two
words, for example, can be analyzed by activating their semantic fields
and decomposing their overlap into separate tokens.

A similar approach extends to text paragraphs: accumulating the
activations of every word into one single multiset and then reducing it
via kWTA gives a sparse representation of the paragraph's semantic
field. This can be subsequently used for supervised classification or
unsupervised clustering with a separate hetero-associative memory
instance.

We can measure the semantic similarity between words directly through a
comparison of their dense activation patterns. This requires a
similarity metric that compensates for the probability of random
overlaps among dense representations --- a straight overlap or Jaccard
distance calculation does not account for this. The adjusted Rand index
(ARI), defined in the formula below, is a suitable measure for the
similarity of dense distributed representations:

\[\texttt{ARI}(A, B, N) = \frac{|A \cap B| - \frac{|A| \cdot |B|}{N}}{\min(|A|,|B|) - \frac{|A| \cdot |B|}{N}}\]

Experiments using the word associations dataset validate the feasibility
of this approach. For example, the ARI score between the words ``ship''
and ``boat'' is 0.614, while ``ship'' and ``cheese'' score at -0.025,
indicating dissimilarity.

The following figure shows a similarity matrix computed for a few words
via the ARI among their densely activated word associations.

\begin{figure}[H]
\centering
\includegraphics[width=0.5\linewidth,height=\textheight,keepaspectratio,alt={A similarity matrix of 9 words, computed as the ARI between their semantic fields.}]{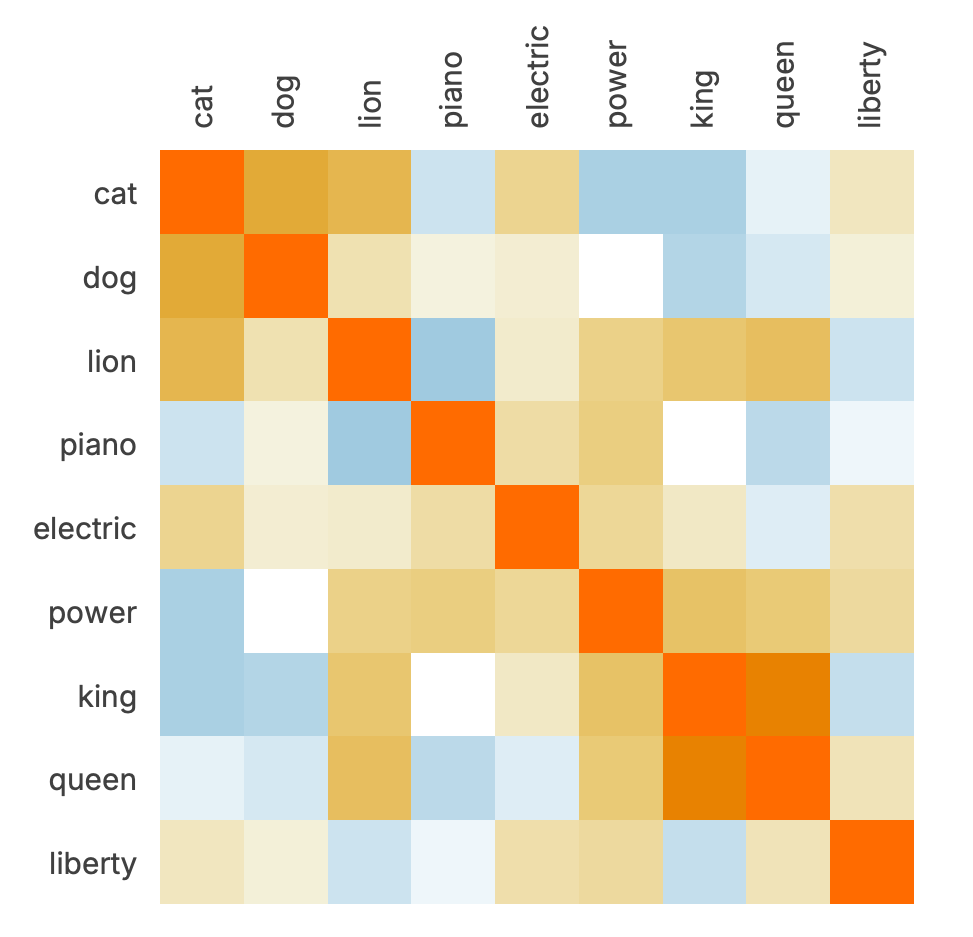}
\caption{A similarity matrix of 9 words, computed as the ARI between
their semantic fields.}
\end{figure}

A similar result would be obtained using a topological
hetero-associative memory algorithm that returns the raw activated
hidden layer as a multiset, skipping the kWTA reduction to sparse binary
sets. The neural equivalent of the ARI computation is a process located
within the memory's expanded hidden layer. Such mechanisms are a natural
extension of our computational framework which does not expose the
combinatorial state of the latent space.

\subsection{Frames}\label{frames}

Marvin Minsky's frame theory, introduced in his 1974 MIT AI Memo, models
knowledge as a collection of ``frames'': modular data structures
representing stereotyped situations. Frames function as interconnected
grid-shaped subgraphs of the mind's knowledge hypergraph. This nicely
aligns with auto-associative topological memory which represents
undirected hyperedges as bindings of sparse holographic representations
(SHR).

A frame contains top-level nodes for fixed, invariant properties and
lower-level terminals (``slots'') for specific instance data
(``fillers''). These terminals carry default assignments that the system
replaces when new information becomes available. This architecture
enables pattern matching and common-sense reasoning, allowing a system
to infer context from pre-stored templates.

Here is an example for a ``country'' frame. The first row contains the
top-level nodes, the second row indicates the specific instance of the
frame, and the fillers for their respective slots are in the third row:

\begin{figure}[H]
\centering
\includegraphics[width=0.583\linewidth,height=\textheight,keepaspectratio,alt={An instance of a frame.}]{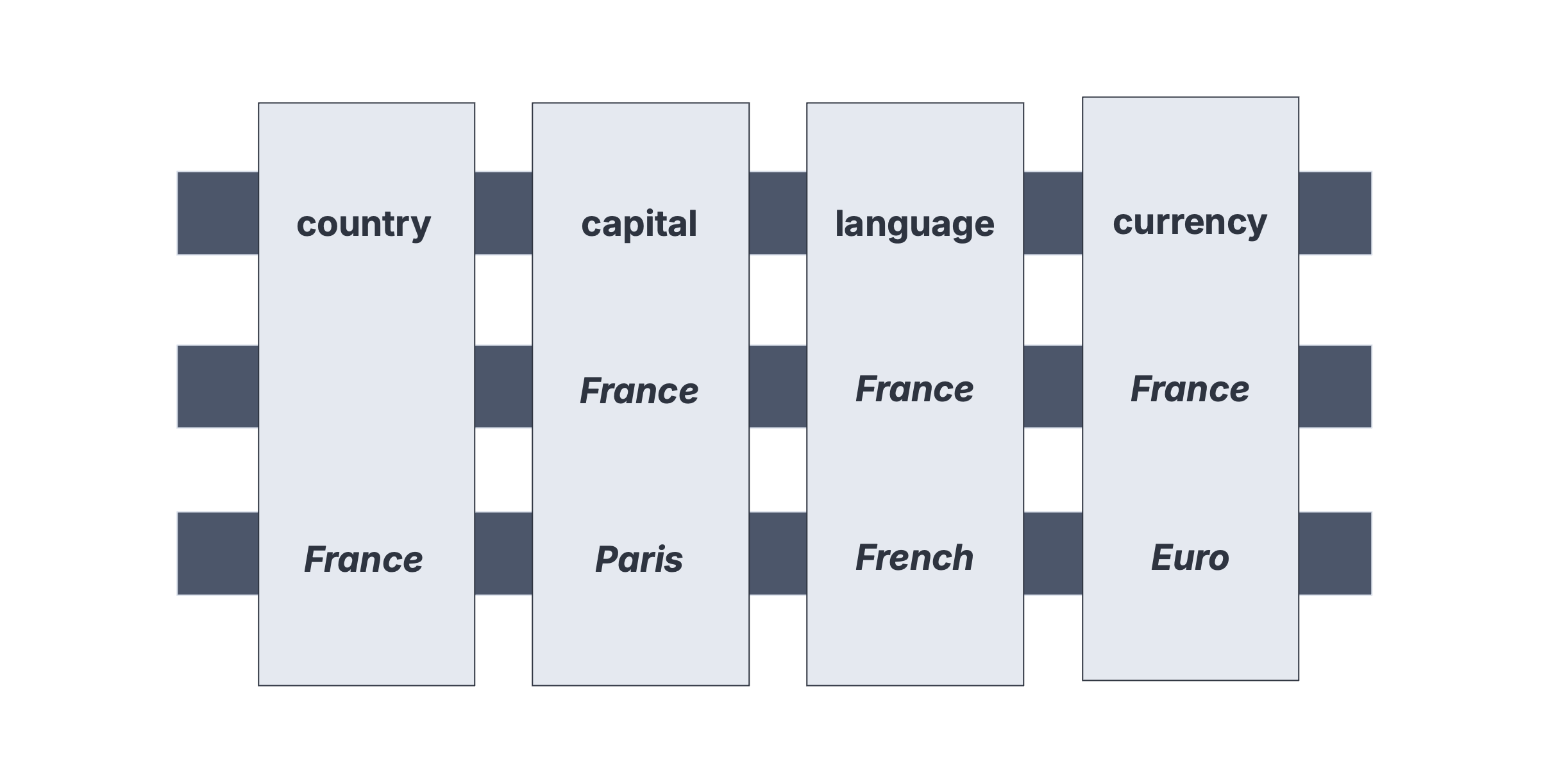}
\caption{An instance of a frame.}
\end{figure}

Each column represents a slot, binding a top-level node (for instance
``capital'') with a filler value (``Paris'') in the context of the
specific frame instance (``France''). The first column plays a special
role: It binds the class (``country'') to the instance (``France''). In
this bare-bones model, a frame always has three rows, while the number
of columns varies. In more elaborate versions, additional information
such as procedural instructions or type constraints, often called
``facets'', can be attached via an extra row. And it might be useful to
incorporate a unique pointer to serve as the frame's universal
identifier and locator. In the grid shown above, such a pointer would be
placed in the empty cell at the left.

A column where the middle cell is empty binds a slot to a generic
default value. This default will be used by any instance that has no
filler (the bottom cell) specified. This behavior is an essential part
of Minsky's theory.

To realize a frame in auto-associative memory, we simply store each row
and each column as bundles in memory. All elements (the intersections of
rows and columns) are encoded as sparse holographic representations
(SHRs). Storing rows or columns as bundles in memory binds their
elements.

(Unordered columns are used here for simplicity. A more rigorous
implementation would structure columns as non-overlapping
\hyperref[triplestore]{triples}.)

Columns can be queried in three directions: Memory retrieval with a
bundle of two tokens adds the missing third token to complete the triad.
In our example, querying with the bundle ``capital France'' returns
``capital France Paris''. But we can also query with ``capital Paris'',
or ``Paris France''. Similarly, querying the memory with elements of a
row completes the row.

Minsky defined the process of completing a frame via horizontal and
vertical queries as ``matching''. This involves several cognitive
sub-tasks which transition a general blueprint into a specific
understanding of a situation: Slot-filler binding is the cognitive
operation that binds a specific entity (``Paris'') to a known structural
role (``capital''). If real-world data is missing, the mind
automatically completes the frame using default values: typical
expectations that stay in place until contradicted. This process is
known as ``default reasoning''. Minsky also argues that perception is
driven by expectation --- we use frames to make conjectures and fill in
details that are not currently visible.

Frames are linked if they share an element, or several elements. Shared
elements may occur in different rows. A class or instance name in one
frame can serve as a filler value in another. This allows frames to be
nested and to form hierarchies. But also cyclic structures are possible,
as well as links to subgraphs that are not shaped like frames.

Integrating a new piece of knowledge into an existing frame is a
multi-step operation --- an aspect of \emph{memory consolidation}.

A new item must be added to two bundles: the row and the column. This
process involves reading an existing row, bundling it with the new item,
and storing the resulting expanded row back into memory. The same steps
are applied to the column.

The necessity to retrieve a bundle before storing an extended version of
it back into memory follows from the mathematical irreducibility of
autoassociations: There is no atomic memory operation that would append
information to auto-associative memory.

Knowledge retrieval from frames works via simple associative memory
queries, using either individual tokens (SHRs) or token bundles as
input.

The following simple example, based on the ``country'' frame,
demonstrates this mechanism. We populate the memory with the frame for
``France'' (as shown in the above figure), and also with the
corresponding frames for ``UK'' and ``Netherlands''.

In the interactive examples shown below, every token is implicitly
encoded as an SHR, and adjacent tokens are bundled. The results of
memory queries are implicitly
\hyperref[iterative-decomposition]{unbundled} via an item memory and
displayed in decoded form.

First we query the memory with the class name. The output is a list of
its top-level nodes and its instances:

\begin{verbatim}
> country
country capital language currency France UK Netherlands
\end{verbatim}

The class and the name of an instance are paired:

\begin{verbatim}
> country UK
country UK
\end{verbatim}

The name of an instance is bound with all cells in the frame:

\begin{verbatim}
> France
country capital language currency France Paris French Euro
\end{verbatim}

Querying with a top-level node pulls information from all rows and
columns where it occurs:

\begin{verbatim}
> language
country capital language currency France French UK English 
    Netherlands Dutch
\end{verbatim}

And the same happens with a filler token as input:

\begin{verbatim}
> English
language UK London English Pound

> Paris
capital France Paris French Euro
\end{verbatim}

This joins information from two frames:

\begin{verbatim}
> Euro
currency France Paris French Netherlands Amsterdam Dutch Euro
\end{verbatim}

``Vertical'' retrieval including the slot's top-level node acts like a
common database query:

\begin{verbatim}
> capital UK
capital UK London

> language France
language France French

> currency Netherlands
currency Netherlands Euro

> currency Pound
currency UK Pound

> currency Euro
currency France Netherlands Euro
\end{verbatim}

As expected, the last example joins data from both frames where
``currency'' is bound to ``Euro''.

In a ``horizontal'' query, where the input consists of filler values,
the memory attempts to complete the row with known information:

\begin{verbatim}
> Euro Dutch
Netherlands Amsterdam Dutch Euro

> Euro Paris
France Paris French Euro
\end{verbatim}

Here the ``Netherlands'' frame wins against ``France'' with 3-to-2
matching elements:

\begin{verbatim}
> Euro Paris Amsterdam Dutch
Netherlands Amsterdam Dutch Euro
\end{verbatim}

The following query results in a 2-to-2 tie between ``France'' and
``UK''. In such an unstable situation, the outcome is influenced by
possible random overlaps between the underlying SHRs, causing the two
competing inputs to have slightly different populations. Also randomized
subsampling of the input may nudge the system into a nearby stable
state.

\begin{verbatim}
> France Paris UK London
UK London
\end{verbatim}

Similarly, ``diagonal'' queries are generally unstable: the result
depends on small representational details.

Sometimes we get reasonably meaningful answers:

\begin{verbatim}
> language Amsterdam
language Netherlands Dutch

> currency Paris
capital France Euro 

> language London
capital UK English

> language Euro
France Paris French Euro

> currency English UK
UK Pound
\end{verbatim}

But it also happens that a diagonal query bounces with a generic result
or the verbatim input:

\begin{verbatim}
> currency Paris
country capital language currency

> language currency France
language currency France
\end{verbatim}

Unstable queries are a well-known phenomenon in Minsky's frame theory,
and an expected behavior within our independent implementation. Rather
than a limitation, this instability provides a plausible model of the
mind attempting to make sense of incoherent perceptions. In real-world
applications there is normally a dense stream of inputs, and
intermittent ambiguities dissolve as new information arrives.

The topological memory implementation of the frame model offers a simple
mechanism to suppress unstable outcomes: increasing the pattern matching
threshold to a value slightly above the SHR population will reliably
generate an empty result in response to an ambiguous query.

\subsection{Random walks on graphs}\label{random-walks-on-graphs}

Random walks on knowledge graphs are an essential mechanism of
cognition, exploring mutual relationships and semantic similarities.
With graphs stored in auto-associative memory, a random walk unfolds
simply by chaining memory queries, using the subsampled output from one
step as the input for the next. This is easily set up using an
auto-associative memory and the \emph{complement} update rule:

\begin{figure}[H]
\centering
\includegraphics[width=0.383\linewidth,height=\textheight,keepaspectratio,alt={Generative auto-associative memory circuit, using the complement update rule and an external feedback path.}]{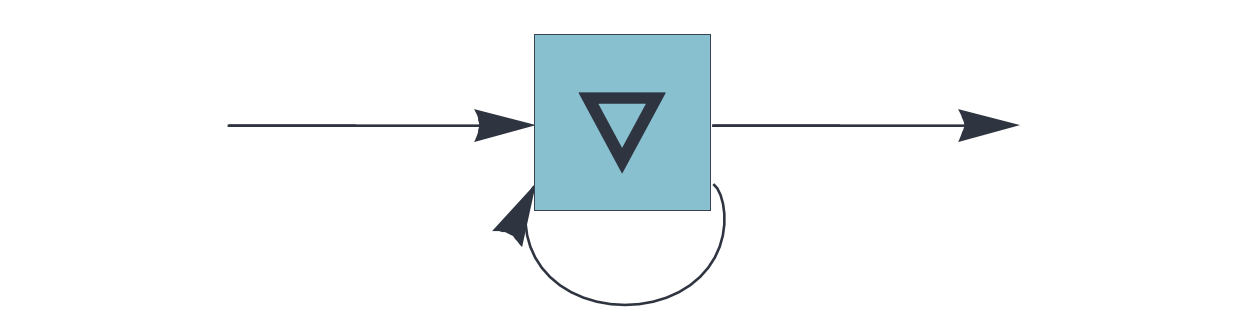}
\caption{Generative auto-associative memory circuit, using the
\emph{complement} update rule and an external feedback path.}
\end{figure}

Once an initial signal from the memory's input channel kicks off the
process, the walk continues independently. If external input is received
during a walk, it acts as a guide, nudging the walking direction towards
the specified destination.

Querying a graph-like structure in auto-associative memory usually
returns a bundle of vertices linked to the input. This output, a bundle
of multiple nodes, serves as subsequent input. Resolving the
superposition to a single node is not necessary. Stochastic subsampling
ensures that each step follows a trajectory weighted by graph
connectivity. This mechanism transforms the retrieval process into a
Markovian walk across the memory landscape. Graph traversal is an
inherent property of auto-associative memory retrieval. It requires no
additional program logic or architectural overhead.

This process works equally for undirected graphs, for directed graphs
(bipartite representations), and for labeled directed graphs (triple
partitions). It also extends to hypergraphs where blocks contain bundles
of vertices.

To experiment with this behavior, we generate a pseudorandom directed
graph using a uniform graph distribution with 300 vertices and 600
edges. Edges are encoded as bipartite sets. The auto-associative memory
applies the \emph{complement} update rule and proportional stochastic
decimation by a factor \(D = 0.8\). The process starts from a single
random node and runs for 300 steps.

\begin{figure}[H]
\centering
\includegraphics[width=0.625\linewidth,height=\textheight,keepaspectratio,alt={Random walk on a knowledge graph using the complement update rule.}]{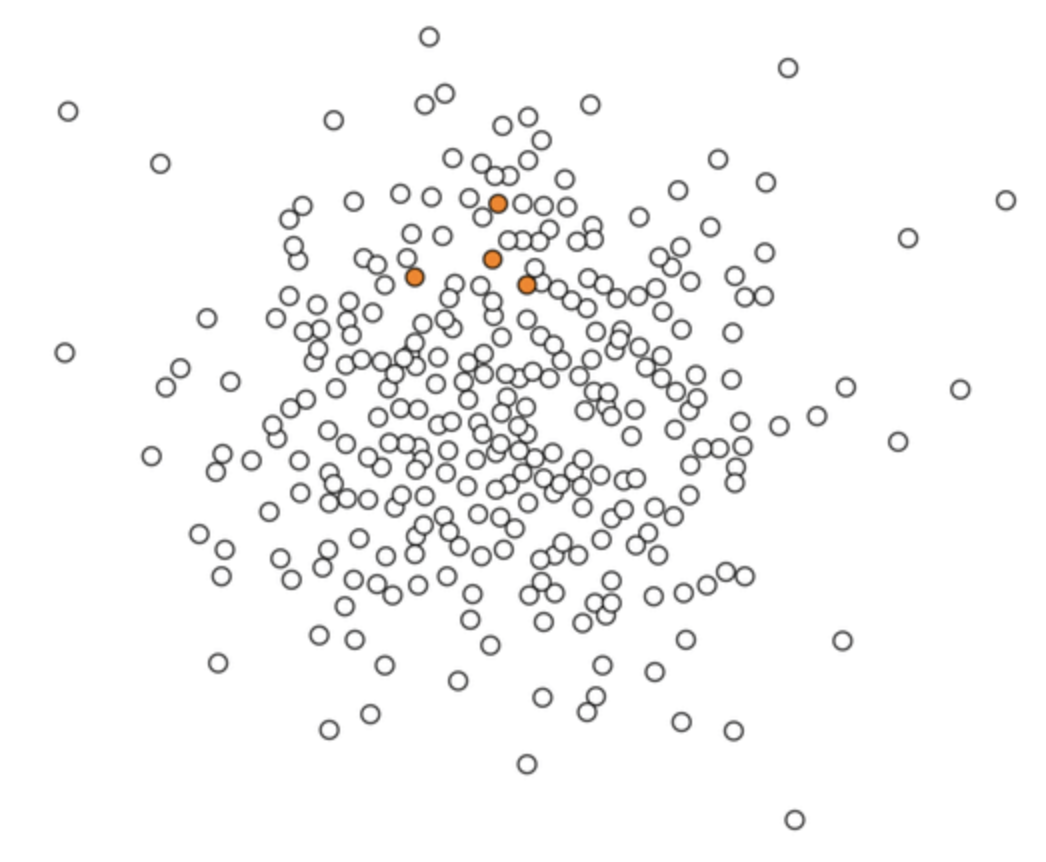}
\caption{Random walk on a knowledge graph using the \emph{complement}
update rule.}
\end{figure}

During this run, about 100 different combinations of vertices get
activated. As a consequence of the \emph{complement} update rule,
consecutive vertex constellations do not overlap.

Applying the \emph{difference} update rule changes the behavior: now
consecutive activations overlap, slowing down the walking pace and
increasing the average size of active clusters. In this run (again with
300 steps) most of the generated states were distinct.

\begin{figure}[H]
\centering
\includegraphics[width=0.625\linewidth,height=\textheight,keepaspectratio,alt={Random walk on a knowledge graph using the difference update rule.}]{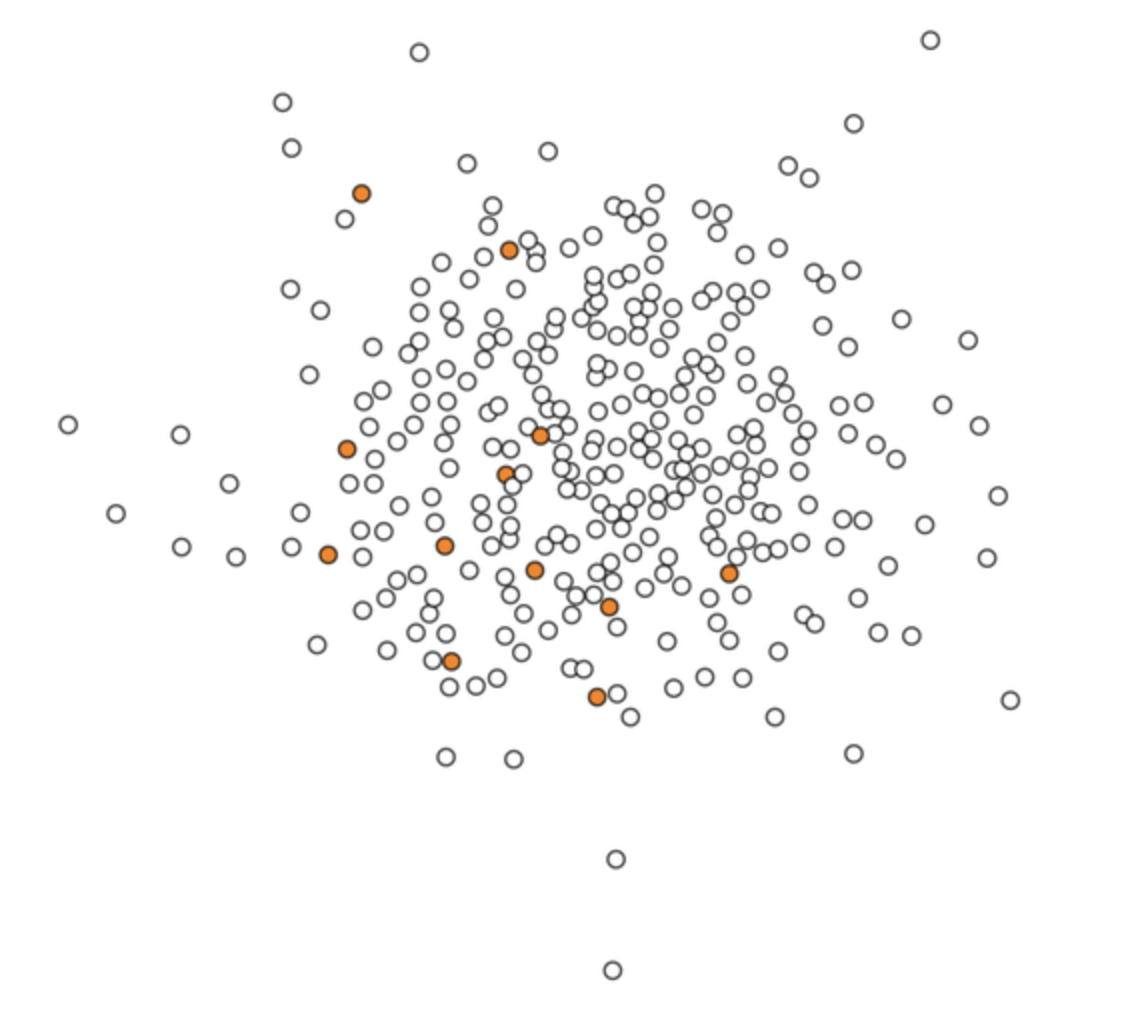}
\caption{Random walk on a knowledge graph using the \emph{difference}
update rule.}
\end{figure}

(Animations of these random walk experiments are available on this
project's website.)

The setup described above implements an undirected random walk on a
directed graph, because querying with a single vertex returns its
neighborhood. State transitions happen between node clusters rather than
individual nodes. Realizing a directed node-to-node walk requires two
configuration changes: reducing the bundle retrieved from
auto-associative memory to a single token by routing it through an item
memory, and swapping the positions within the bipartite representation
via crossing feedback lines.

\section{Software Implementations}\label{software-implementations}

\subsection{Standard C reference
implementation}\label{standard-c-reference-implementation}

The topological memory kernel, along with a library for set manipulation
and validation testing, is available on
\href{https://github.com/PeterOvermann/CreatingIntelligence}{GitHub}. It
provides a unified programming interface for both auto-associative and
hetero-associative memory.

This reference implementation is written strictly in standard C. It has
zero dependencies on external libraries and is fully portable across
platforms. This allows direct execution on resource-constrained
hardware. The reference implementation also serves as the baseline for
porting the architecture to other languages and environments. To ensure
consistency, all ports should include the provided test program to
verify parity with the reference version.

Performance-critical subroutines, specifically memory access and
bit-level operations, are optimized for single-threaded execution.

A typical AI development environment uses a high-level interpreted
language, usually Python, as the frontend for scripting application
logic and prototyping experimental models. All the performance-critical
operations are delegated to low-level libraries. This C-language
implementation can be easily integrated as a high-performance library
with a variety of frontends.

Some notes on performance:

Storing an association into memory has a time complexity of
\(\mathcal{O}(P^2)\), where \(P\) is the number of elements in the input
pattern. This is independent of the amount of information already
stored.

The standard C implementation, benchmarked on a 2023 laptop computer,
achieves a single-threaded write speed of about 600,000 random
associations per second, using hyperparameters \(N = 1,000\) and
\(P = 10\). Given this high throughput, the code is well-suited for
applications requiring real-time model updates.

Data retrieval also follows \(\mathcal{O}(P^2)\) complexity if the input
is noise-free, remaining independent of memory density. Under the same
testing configuration, the system retrieves roughly 100,000 items per
second.

When retrieving information with a noisy input pattern, read performance
is linked to the number of iterations needed to extract the best subset
match. Each successive iteration accelerates as the number of active
inputs drops. Empirical testing shows that doubling the input size by
adding random input elements results in an approximate 4x decrease in
retrieval speed.

\subsection{Mathematica version}\label{mathematica-version}

The topological memory algorithm was initially prototyped in
\emph{Mathematica}, a high-level interpreted programming language. The
native \emph{Mathematica} version is included in the repository,
alongside a package for designing high-level circuits. Consisting of
just 50 lines of code, the \emph{Mathematica} implementation is highly
compact.

The standard C library can be dropped into \emph{Mathematica} via a
foreign function interface, replacing the native implementation. It
should be straightforward to replicate this kind of setup with
alternative (and probably more familiar) development environments.

\section{Post-GPU hardware}\label{post-gpu-hardware}

The massive efficiency gap between an autonomous robot and a biological
brain is not a hardware problem. Hardware simply mirrors the underlying
data structures. The root problem lies in the computational foundation:
the cost of dense matrix arithmetic compared to sparse, discrete
operations.

Hardware optimized for discrete associative memory fundamentally
diverges from contemporary AI accelerators. Graphics processing units
(GPUs) and tensor processing units (TPUs) specialize in matrix
multiplication. They lack the specialized circuitry required for
bit-level operations on extended binary arrays --- the core
computational mechanism underlying topological memory.

Traditional computer architectures separate memory and processing,
inevitably creating a data-transfer bottleneck. Because topological
memory relies heavily on large-scale bitvector arithmetic, this
separation severely limits overall performance. Scaling these systems
requires moving beyond the current trend of computing in or near memory.
Instead, the hardware must mimic the brain by computing directly with
the memory itself, unifying storage and logic into a single mechanism.

Topological memory is based on binary logic and bit-plus-integer
addition, lending itself to a direct integration with existing memory
technologies. Neural networks rely on complex arithmetic that resists
integration into high-density storage. This fundamental mismatch creates
the \emph{memory wall} bottleneck in modern AI accelerators.

Post-GPU hardware, optimized for AI applications based on topological
memory, can take different shapes, as outlined in the following.

\subsection{Processing with DRAM}\label{processing-with-dram}

Dynamic random-access memory (DRAM) technology is a promising starting
point for the design of a native topological memory processor. Standard
DRAM row sizes (typically 8, 16, or 32 Kbits) align with the necessary
dimensions of associative memory sets.

A DRAM-based accelerator augments high-density memory with specialized
logic to store and retrieve associations. The unit requires very low
external bandwidth because it only receives and transmits small,
serialized sets. Internal logic expands these incoming sets into full
bitvectors.

Some architectural details:

\begin{itemize}
\tightlist
\item
  The triangular prism of the rank-3 tensor maps cleanly onto the DRAM
  structure. A dedicated row decoder translates every pair of input
  elements \((i,j)\) into a logical memory row.
\item
  To store a new association, the target data is temporarily loaded into
  a bitvector buffer. The system then updates all activated memory rows
  simultaneously via a logical OR operation, setting relevant bits to 1.
  DRAM can perform this logic across an entire row in a single step: The
  usual column decoders are not needed here.
\item
  Memory retrieval uses analog processing. The system simultaneously
  opens the target DRAM rows, allowing electric charges in the active
  memory cells to accumulate along each column (the bitline). An analog
  electronic circuit then rapidly computes
  \hyperref[k-winners-take-all]{kWTA}, using for instance the mechanism
  proposed by Lazzaro (1988).
\item
  Additional circuitry manages iterative noise removal.
\item
  A conventional data interface allows model data to flow to and from
  external bulk storage.
\end{itemize}

DRAM-based associative memory accelerators can be adapted into several
distinct hardware formats:

\begin{itemize}
\tightlist
\item
  System-on-Chip (SoC) integration or mainboard-mounted memory chips
\item
  Computational RAM (C-RAM) modules
\item
  Neuromorphic processors
\item
  Discrete accelerator cards for servers and workstations
\item
  Functional co-processors deployed alongside standard GPUs and TPUs
\end{itemize}

This architecture computes directly in memory. That makes it ideal for
continuous, real-time model updates under strict power constraints.
Prime applications include battery-powered autonomous robots, drones,
mobile devices, IoT hardware, and brain implants.

\subsection{NAND implementation}\label{nand-implementation}

Certain applications, particularly autonomous robotics and edge devices,
require their model memory to persist without continuous power. An
associative memory unit built from NAND flash modules fulfills this
need. We couple flash memory arrays with a custom bitvector processing
unit. This ensures the system retains learned associations across power
cycles and reboots.

NAND flash reads and writes data in large, contiguous pages. This
architectural trait aligns perfectly with the demands of topological
memory, which relies on processing extensive bitvectors simultaneously.
The specialized controller executes the necessary logical operations
directly alongside the storage modules, matching the wide data buses of
the flash memory to the wide vector requirements of the associative
algorithm.

Similar to DRAM-based designs, a NAND implementation drastically reduces
the data transfer bottleneck. The heavy computational lifting occurs
internally within the flash subsystem. The external interface is
reserved exclusively for transmitting sparse data sets in a highly
compressed, serialized format, ensuring rapid communication with the
rest of the system using minimal bandwidth.

\subsection{An associative connection
machine}\label{an-associative-connection-machine}

A powerful intelligent machine may consist of hundreds or thousands of
circuit components with self-contained associative memory instances.
These nodes are sparsely interconnected, fetching their inputs
(serialized small sets) from the output registers of directly connected
nodes. Nodes are individually programmed to carry out specific memory
functions, as explored earlier in the discussion of neural circuit
components. The connection topology is programmable and could even
evolve at runtime in order to achieve a degree of plasticity.

Such a system can be built from memory units proposed in the previous
section, combined with a configurable chip-to-chip interconnect. A
low-bandwidth interconnect is adequate for peer-to-peer networking
between memory units, as only sparse data is exchanged. It might even be
feasible to build a system from geographically distributed units
connected via wide-area networks, creating a form of hive intelligence.
On the other end of the design spectrum, an entire self-contained system
could be integrated onto a single silicon wafer.

The machine can be designed to tolerate small manufacturing faults and
sporadic memory or transmission glitches, considering the associative
memory's resilience against systematic and statistical errors. This
should significantly reduce production costs.

A von Neumann computer relies on a central processor. An associative
connection machine has none: computation is fully distributed across
decentralized memory units.

Some nodes can receive external inputs, which are already preprocessed
and encoded as small sets. Different input nodes specialize in various
data types, such as images, audio, text, or physical sensors. Similarly,
specialized output nodes connect to decoders that transform small sets
to external data formats. Different regions within the machine may be
more or less active, depending on the computational task being
performed. The system is always learning and adapting to changing
external inputs, continuing processing even in their absence.

Conventional computers have two bottlenecks that lead to high-pressure
data flow: small word sizes (up to 64 bits) and narrow memory
interfaces. An associative connection machine uses hyperdimensional
words (such as 8, 16 or 32 Kbits). Neighboring nodes exchange only
sparse data.

Eliminating these two bottlenecks minimizes the data pressure, allowing
the system to run at comparatively low clock speeds. Operating at
megahertz instead of the usual gigahertz frequencies, it would consume
only a tiny fraction of the energy required by today's AI accelerators.

Assuming a 10,000-fold increase in parallelization due to
hyperdimensional word sizes, the system can operate at a proportionally
lower clock speed. Because power consumption scales quadratically with
frequency, an associative connection machine requires merely 1/10,000
the power of a conventional computer performing the identical task. This
energy efficiency directly mirrors the human brain, which relies on
massive spatial parallelization to compensate for the fundamentally slow
operating speeds of biological neurons.

\section{Algorithm Variants}\label{algorithm-variants}

\subsection{Non-binary memory storage}\label{non-binary-memory-storage}

The baseline topological memory algorithm uses a binary storage model.
This approach is significantly more space-efficient than storing integer
counters or floating-point values.

Each storage bit represents the topological connectedness between an
input pair and an output element. A 0 means no link exists, and a 1
means at least one link exists. This model does not count parallel
connections, as redundant routes do not alter the transmitted binary
signal.

An alternative storage model might use non-binary memory values,
tracking the number of topological connections, or serving as an
abstract model of connection strength.

While much less space-efficient than a binary model, this approach has
an immediate advantage: it enables the exact reversal of a write
operation without degrading overlapping, unrelated information. This
precise rollback feature is useful for technical applications, even if
it diverges from biological memory decay models.

The baseline topological memory algorithm stores information in memory
sets, each linked to a node within the expanded hidden layer. In a model
based on integer counters, those memory sets are replaced with memory
multisets. The multiplicity of an element within a memory multiset
indicates the number of output patterns to which a specific hidden layer
node contributes. Storing new information increments those counters, and
decrementing the same counters reverses the storage operation.

Here is the modified write algorithm for a memory model with integer
counters. Step 2 uses multiset aggregation where the original algorithm
uses the union operation:

\textbf{Input:} Sets \(A\) and \(B\)

\begin{enumerate}
\def\labelenumi{\arabic{enumi}.}
\item
  \textbf{Expansion coding:} Map 2-subsets \(\binom{A}{2}\) to memory
  multisets \(M_{ij}\)
\item
  \textbf{Memory update:} \(M_{ij} \gets M_{ij} \uplus B\) (multiset
  summation)
\end{enumerate}

The corresponding deletion process uses multiset subtraction, stopping
at zero because an element cannot appear a negative number of times.

\textbf{Input:} Sets \(A\) and \(B\)

\begin{enumerate}
\def\labelenumi{\arabic{enumi}.}
\item
  \textbf{Expansion coding:} Map 2-subsets \(\binom{A}{2}\) to memory
  multisets \(M_{ij}\)
\item
  \textbf{Memory deletion:} \(M_{ij} \gets M_{ij} \setminus B\)
  (truncated multiset subtraction)
\end{enumerate}

Memory retrieval aggregates the overall response multiset by combining
the memory multisets, accurately adding up each element's
multiplicities. Here are the steps for a single-pass memory query.

\textbf{Input:} Query set \(A\)

\textbf{Output:} Unique best match \(Y\), or an empty set
\(\varnothing\)

\begin{enumerate}
\def\labelenumi{\arabic{enumi}.}
\item
  \textbf{Threshold check:} If \(|A| < T\) return \(\varnothing\)
\item
  \textbf{Expansion coding:} Map 2-subsets \(\binom{A}{2}\) to memory
  multisets \(M_{ij}\)
\item
  \textbf{Aggregate:} \(R \gets \sum_{ij} M_{ij}\) (overall response
  multiset)
\item
  \textbf{Select}: \(Y \gets R_{\top P_B}\) (computing
  \hyperref[k-winners-take-all]{kWTA})
\item
  \textbf{Threshold check:} If \(|Y| < P_B\) return \(\varnothing\)
\item
  \textbf{Output:} return \(Y\)
\end{enumerate}

This algorithm can store and retrieve heteroassociations much like the
core algorithm of topological memory. Using integer counters instead of
bits does not increase the memory's storage capacity.

Unlike topological memory, this model does not deduplicate information:
storing the same association twice increases its weight during memory
retrieval.

However, neither this counter-based algorithm nor variants with
continuous weights are suitable for autoassociations. When storing
autoassociations, the values corresponding to the self-correlations
quickly inflate because they increase at every step. When
self-correlation dominates cross-correlation, the memory collapses into
a useless identity operation: a memory query with any input pattern,
whether known or unknown, returns exactly the input. A similar
phenomenon has been observed in classical auto-associative memories
based on continuous weights.

This demonstrates that the binary model of topological memory
connections is not just a space optimization. Instead, binary memory
storage is a functional necessity for auto-associative functionality to
emerge.

\subsection{Memory decay}\label{memory-decay}

Forgetting is a natural behavior of memory: it is necessary to prevent a
loss of functionality as the memory fills up. Topological memory
reliably functions up to a density \(\rho^* = 0.5\), and smoothly
deteriorates above this mark. We model information decay as a discrete
process that removes connections within the memory topology, much like
learning opens such connections.

Random removal of topological connections causes information to degrade
gracefully, rather than instantly erasing entire patterns. This is
because topological memory stores information redundantly: sets
distribute information across multiple elements, and the memory
distributes the information for each element pair across its vast latent
space. An association remains accessible as long as the number of its
remaining intact connections stays at or above the pattern matching
threshold (\(T\)). Forgetting is the transition where the activation of
a pattern, thinned by stochastic decay, finally falls below this
threshold: the information becomes inaccessible to the retrieval
algorithm, and is categorized as ``unknown''.

This mechanism differs from the catastrophic forgetting found in deep
neural networks. In connectionist models, global weight adjustments
originating from a single gradient can inadvertently overwrite
previously stored information. In topological memory, even though bits
are shared, the sparsity and hyperdimensionality of the space ensure
that random bit removal has only local effects within the entire
knowledge base. The discrete nature of patterns prevents the kind of
ripple effects seen in conventional scalar weight-shifting.

\begin{quote}
\textbf{Local forgetting}\\
The stochastic dissolution of signal paths within the memory topology.
Unlike catastrophic forgetting, it causes a graceful thinning of
information that only renders a pattern inaccessible when its activation
falls below the pattern matching threshold.
\end{quote}

We can model stochastic information decay with two distinct approaches:
by decreasing entropy or by increasing entropy.

Information entropy decreases if we randomly clear memory locations as
we store new information. Here is a computationally convenient
mechanism: whenever we write \(B\) bits into memory, we clear
\(C \ge B\) bits chosen uniformly at random across the entire memory
space. The order of these operations does not matter --- collisions are
rare because the memory space is hyperdimensional. Writing or clearing a
bit means setting its value to \emph{on} or \emph{off}, respectively ---
regardless of its current state.

With this setup, the memory density \(\rho\) converges to a steady
state:

\[\rho_\infty \approx \frac{B}{B + C}\]

If the writing and deletion rates are the same, the memory density
remains at a safe value \(\rho^* = 0.5\). All active bits reflect
information that has been stored in memory. This setup is a suitable
model for a long-term memory with graceful, non-catastrophic information
decay.

The same mechanism also serves as a model of short-term memory where the
memory density always remains low because network connections quickly
dissolve. Memory traces degrade via decreasing entropy (the
\emph{monotone} effect). In the brain, the dentate gyrus (DG) shows this
behavior: a low density directly correlates with fast information decay.

Note that simulating short-term memory with this approach has a cost.
Maintaining a low density --- say, 5 percent as empirically observed in
the DG --- would require an aggressive deletion rate that is about 20
times the writing rate.

The reference implementation of the core retrieval algorithm includes an
optional decay mechanism that caps each hidden node's connections at
half of all possible connections \(N_B\). This is triggered whenever a
hidden node (the memory sets \(M_{ij}\)) is updated during a memory
write process.

As an interesting alternative to the information decay mechanism
outlined above, we can model forgetting by increasing entropy. This
involves writing random \emph{on} bits that do not correspond to known
information.

Consider a memory filled to capacity \(\rho^* = 0.5\) with random bits.
With respect to the topological memory retrieval algorithm, this memory
is effectively empty. Any query will fail to return a result. But if we
store an association on top of this random noise, it can be accurately
retrieved. (Note that memory retrieval works reliably for memory
densities \(\rho \lesssim 0.6\)).

To maintain a steady-state density as well as high information entropy,
we must add and clear random bits during each memory write operation.
For instance, when writing a pattern involving \(B\) bits, we also write
\(B\) random bits and independently clear \(2B\) bits randomly chosen
from all memory locations. This results in the memory density
fluctuating around 0.5 with fast information decay due to the \emph{heat
death} effect: most active bits do not reflect previously stored
information.

This entropy-increasing mechanism works well as a computational model of
short-term memory, as it requires a much lower deletion rate than the
entropy-decreasing mechanism.

When a stochastic memory decay mechanism is enabled, we can protect
commonly used memory traces from fading by writing them back to memory
after each successful retrieval. With a memory operating at its nominal
capacity, this write operation will trigger the stochastic deletion of
other, competing memories. This approach aligns with a phenomenon known
as retrieval-inspired forgetting (RIF) in cognitive psychology:
successfully recalling a memory suppresses and weakens competing, noisy
associations.

\subsection{Cross-associative memory}\label{cross-associative-memory}

Besides \hyperref[auto-associative-memory]{autoassociations} and
\hyperref[hetero-associative-memory]{heteroassociations}, the core
algorithm supports a third type: crossassociations. This structure,
first described in the \emph{Triadic Memory} paper, completes the basic
associative memory mechanisms in this computational framework.

\begin{quote}
\textbf{Crossassociation} A tridirectional memory relation
(\(A \times B \to C\)) that links all pairwise combinations of elements
in two sets to a third set.
\end{quote}

A crossassociation \(A \times B \to C\) links all pairwise combinations
of elements in \(A\) and \(B\) to all elements of \(C\). Here
\(A \times B\) denotes the usual Cartesian product between two sets. In
standard hetero-associative memory, storing a bundle \(A \cup B \to C\)
automatically writes the subcomponents \(A \to C\), \(B \to C\), and the
crossassociation \(A \times B \to C\).

This provides a direct mechanism for unbinding. A standard binding
\(A \cup B
\to A \cup B\) mathematically decomposes into \(A \to A\), \(A \to B\),
\(B \to A\), \(B \to B\), and the crossassociation
\(A \times B \to A \cup B\). Selectively clearing only the
cross-associative component decouples the tokens without destroying the
individual items or their other bindings.

Cross-associative topology is fully symmetric across the \(A\), \(B\),
and \(C\) triad. Storing \(A \times B \to C\) is mathematically
identical to storing \(B
\times C \to A\) or storing \(C \times A \to B\). Therefore,
crossassociations can be queried along all three directions with equal
accuracy.

\subsection{Cascaded kWTA retrieval}\label{cascaded-kwta-retrieval}

Cascaded k-winners-take-all (\hyperref[k-winners-take-all]{kWTA})
retrieval bypasses the combinatorial calculations of the core algorithm.
It applies thresholding in two stages: first to the contributions of
each individual input element, and second to the aggregated results.

This variant sacrifices a slight amount of noise tolerance and
false-positive resilience above the nominal memory capacity. For most
practical applications, this tradeoff is negligible.

Here are the algorithmic steps:

\textbf{Input:} Query set \(A\), hyperparameters \(T\) and \(P_B\)

\textbf{Output:} Best match \(Y\), or an empty set \(\varnothing\)

\begin{enumerate}
\def\labelenumi{\arabic{enumi}.}
\item
  \textbf{Initialize:} \(X \gets A\)
\item
  \textbf{Threshold check:} If \(|X| < T\) return \(\varnothing\)
\item
  \textbf{Expansion coding:} Map 2-subsets \(\binom{X}{2}\) to memory
  sets \(M_{ij}\)
\item
  \textbf{Aggregate:} \(R_i \gets \sum_{j} M_{ij}\) (per-element
  contribution)
\item
  \textbf{Select}: \(Y_i \gets (R_i)_{\top P_B}\)
  (\hyperref[k-winners-take-all]{kWTA} per input element)
\item
  \textbf{Aggregate}: \(Y \gets (\sum_i Y_i)_{\top P_B}\) (overall
  \hyperref[k-winners-take-all]{kWTA})
\item
  \textbf{Threshold check:} If \(|Y| < P_B\) return \(\varnothing\)
\item
  \textbf{Per-element weights:} \(w_i \gets |Y_i \cap Y|\)
\item
  \textbf{Refine:} \(X \gets X \setminus X_\text{min}\) (drop weakest
  contributors)
\item
  \textbf{Convergence test:} If no elements were dropped in the previous
  step, return \(Y\)
\item
  \textbf{Iterate:} Go to step 2
\end{enumerate}

On conventional CPUs, this variant runs slower than the reference
algorithm because the system must calculate steps 4 and 5 sequentially
for every input element. However, with-memory computing architectures
perform bitwise aggregation and kWTA in parallel, eliminating this
bottleneck. On specialized hardware, cascaded kWTA is likely the optimal
retrieval method.

\subsection{Raw memory retrieval}\label{raw-memory-retrieval}

The core memory retrieval algorithm relies on threshold checks and
noise-cancelling iterations to separate known from unknown query
patterns. This functionality is essential for processing symbolic tokens
(for example item stores and hyperassociations) where known items can be
reliably extracted from noisy signals.

When working with sparse distributed representations (SDRs) that encode
physical data, there is often no clear boundary between signals and
noise. In such scenarios, a single-pass retrieval algorithm without
threshold checks can be the most efficient choice. This process simply
calculates the k-winners-take-all response to the hidden layer
activations, always returning a non-empty set if there are at least
\(k = P_B\) output elements.

Here is an outline of the steps:

\textbf{Input:} Query set \(A\), hyperparameter \(P_B\)

\textbf{Output:} K-winners-take-all result \(Y\), or an empty set
\(\varnothing\)

\begin{enumerate}
\def\labelenumi{\arabic{enumi}.}
\item
  \textbf{Expansion coding:} Map 2-subsets \(\binom{A}{2}\) to memory
  sets \(M_{ij}\)
\item
  \textbf{Aggregate:} \(R \gets \sum_{ij} M_{ij}\) (overall response
  multiset)
\item
  \textbf{Select}: \(Y \gets R_{\top P_B}\)
  (\hyperref[k-winners-take-all]{kWTA} with cutoff 1)
\end{enumerate}

This method is equivalent to the \hyperref[retrieving-associations]{core
retrieval algorithm} with the pattern matching threshold set to
\(T = 2\). Like the core algorithm, it performs subset pattern matching,
but without determining the contributing elements of the query pattern.
Whereas the core algorithm isolates a unique best match and rejects
unknown inputs, this retrieval method always returns a raw statistical
result, even for completely random inputs.

\section{A Perspective on
Neuroscience}\label{a-perspective-on-neuroscience}

\subsection{Cerebellum}\label{cerebellum}

David Marr (1969) was the first to postulate that the cerebellum
functions as an associative memory. Independently, James Albus (1971)
developed a similar theory. Their combined work, the Marr-Albus model,
remains valid as a simplified computational model of the cerebellum.

The cerebellum's system architecture matches directly the class of
\hyperref[topological-associative-memory]{network topologies} from which
hetero-associative memory functionality emerges.

Mossy fibers, the cerebellum's primary input layer, are randomly
connected to a massive number of granule cells, forming the memory's
hidden layer. Whereas our reference architecture specifies hidden layer
nodes to be doubly connected, each individual granule cell receives
input from about four mossy fibers. This lowers the hidden layer size
required for memory operation, at the cost of a capacity reduction. (The
standard C implementation of the memory algorithm applies a similar
tweak to accommodate input dimensions that would otherwise exceed the
available RAM.)

Granule cells are massively connected via their parallel fiber axons to
Purkinje cells, the memory's output layer. Anatomical studies such as
those by Harvey (1988) and Napper (1988) showed that if a parallel fiber
passes through a Purkinje cell's dendritic tree, it has roughly a 54
percent probability of forming a synapse. This means about half of all
physical crossings result in a potential synaptic connection. This
finding aligns with our memory model, which saturates beyond a 50
percent memory density.

Climbing fibers project into the cerebellum, connecting approximately
one-to-one with each Purkinje cell. This is the memory's secondary input
(the teaching signal \(B\) in associations \(A \to B\)), which is the
signature of hetero-associative topologies and supervised learning.

Interestingly, climbing fibers fire continuously at an average rate of
about 1 Hz, mimicking irregular, Poisson-like noise. Hypothetically,
this random generator might be the source of new SHRs which get deployed
within the \hyperref[heteroencoder]{hetero-encoding} mechanism. Because
climbing fiber activation triggers a prolonged depolarization, such
complex spikes can be temporally integrated to form new sparse
representations, ready to be used as a new teaching signal.

The neural circuitry of the cerebellum is highly homogeneous. This
suggests that it applies the same algorithm to all information it
receives, regardless of what that information means. Based on the
similarity of the cerebellum's architecture with topological associative
memory, I propose that this universal algorithm is hetero-associative
memory retrieval, functioning via subset pattern matching.

\subsection{Neocortex}\label{neocortex}

Reverse-engineering the neocortex's architecture from the viewpoint of
our topological memory shows that it is auto-associative: input and
output functions are localized within the same neural layer. And the
architecture lacks a secondary input, analogous to the cerebellum's
climbing fibers, required for learning heteroassociations.

We see this dual-use design of the input/output layer in a phenomenon
known as \emph{temporal multiplexing}: a quick, synchronized burst (the
low-latency phase) acts as the input query, while the subsequent
repetitions (high-latency phase) represent the retrieved memory. These
repetitions likely serve to filter out noise, much like the iterative
steps in the algorithm described earlier. The precise timing of those
bursts does not encode information, but may be relevant for
synchronization.

Unlike the cerebellum, where a massive amount of specialized neurons
form the memory's hidden layer, the neocortex employs a recurrent
architecture with an intermediate dendritic network to realize its
associative memory topology. While compact and resource-efficient, this
design constrains the physical scale of a fully connected hidden layer
to a dimensionality permitted by dendritic anatomy. For this reason, the
neocortex is segmented into uniform units known as \emph{columns}.
Within the framework of topological associative memory, a cortical
column is the largest possible unit allowing an approximately fully
connected hidden layer.

Vernon Mountcastle (1957) pioneered the idea that the neocortex is
divided into uniform, repeating computing units which all use the same
universal algorithm to process information. Mountcastle left the
mechanics of this algorithm as an open question. The topological memory
framework suggests an answer: the algorithm performed by uniform units
of computation is an auto-associative memory which retrieves information
via subset pattern matching.

What specific computation a cortical column performs depends on the kind
of information being processed, and how it is encoded into sets.
Generally, auto-associative memory can be considered a data store with
the ability to complete missing information. Applied to to
highly-overlapping representations of sensory data, the memory fills in
and predicts missing perceptual data. It processes spatio-temporal
information encoded via grid-cell-like mechanisms. High-level cognitive
processes involve auto-associative memory operating on complex data
structures composed of symbols and pointers, combined to represent
knowledge as hypergraphs.

Mountcastle did not believe his columnar theory extended to other parts
of the brain, such as the cerebellum. In fact, he viewed the neocortex's
particular structure as the exclusive key to higher brain function. But
from the perspective of topological memory, the neocortex and the
cerebellum share the same abstract network topology from which
associative memory emerges. This suggests that both structures, despite
their evident differences, perform the same basic computation.

\subsection{The locus of learning and
memory}\label{the-locus-of-learning-and-memory}

The associative memory's storage model captures the topology of the
corresponding biological network. Its storage bits are Boolean
parameters which model the presence of a connection from a point in the
expanded hidden layer to a point in the output layer. A 0 value means
there is no route, and a value of 1 indicates that there is at least one
open route. The number of routes between two points is not counted.

In this topological model, the connections from the input to the hidden
layer are fixed, and learning takes place anywhere between the latent
and the output layer. This abstraction from specific biological
mechanisms accommodates both traditional synaptic plasticity and the
increasingly supported hypothesis of dendritic learning.

Standard neuroscience views synapses as the primary locus of learning
and memory. This is based on Donald Hebb's suggestion that learning
occurs through the modification of synaptic strength. Hebb (1949)
described synaptic plasticity from a biological perspective. Contrary to
common belief, he did not imply that connection strength is a carrier of
continuous information. Within the context of our topological
associative memory, synaptic weight is strictly binary. This is
consistent with Warren McCulloch's and Walter Pitts' original artificial
neuron model (1943). If synaptic strength is indeed a valid concept, its
function would be to encode the connection's permanence, not just a
fractional piece of a memory.

Synaptic long-term potentiation is traditionally viewed as the primary
mechanism of memory. But recent evidence points to dendritic plasticity
as another essential site of learning. In a model where synaptic
connections are unaffected by the learning process --- whether they stay
static or constantly fluctuate --- the locus of memory shifts to the
dendrites. This design is still fully consistent with Donald Hebb's
postulate that neurons that wire together fire together.

The following sketch illustrates this scheme: the hidden layer (shown as
parallel lines on top) connects via fixed synapses to three dendrites of
a postsynaptic neuron, with each dendrite behaving like an electric
switch.

\begin{figure}[H]
\centering
\includegraphics[width=0.542\linewidth,height=\textheight,keepaspectratio,alt={Dendritic on/off switches as the mechanism of topological plasticity.}]{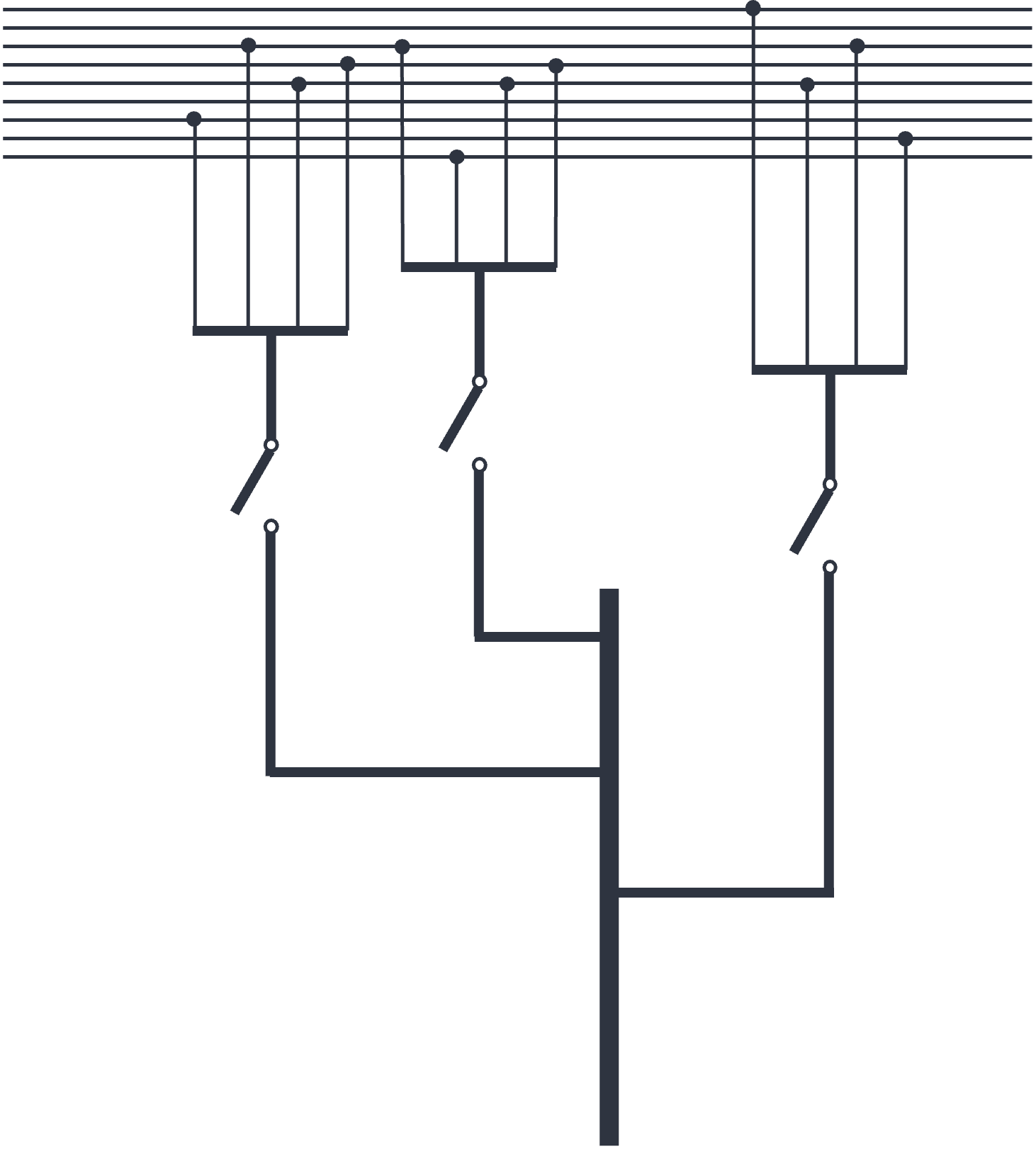}
\caption{Dendritic on/off switches as the mechanism of topological
plasticity.}
\end{figure}

A presynaptic neuron may have multiple connections to the same
postsynaptic neuron via different dendrites. Because the voltage across
every branch of a parallel circuit is equal, the same signal is
transmitted regardless of the number of possible connections. For this
reason, the topological associative memory uses a
\hyperref[non-binary-memory-storage]{binary storage model} as opposed to
integer weights.

The dendritic learning hypothesis aligns well with our architecture:
fixed synapses could be a natural source of pre-wired random encodings
(SHRs) that can be readily and efficiently deployed. Since dendrites are
anatomically closer to the neuron's soma (cell body) than individual
synapses, the proposed learning mechanism could function faster. Also,
the total number of variable parameters required for learning would be
significantly lower, as a single neuron has fewer dendrites than
synapses.

A simple binary switching mechanism within each dendrite would suffice
for topological associative memory to emerge --- there is no need for
dendritic plasticity to provide counters or continuous weights.

The model that dendrites function as dual-state switches is a clear,
testable hypothesis for experimental neuroscience.

\subsection{The binding problem and short-term associative
memory}\label{the-binding-problem-and-short-term-associative-memory}

The binding problem in cognitive science defines the challenge of
integrating separate sensory streams --- such as color, shape, motion,
and location --- into a coherent experience. In visual perception, this
requires precisely associating attributes with their respective objects,
such as binding ``red'' to an apple and ``green'' to a grape. Dropping
this precise linkage results in an illusory conjunction, such as
perceiving a red grape.

Computationally, this integration can be resolved through the binding
operation within auto-associative memory: storing a bundle of sparse
holographic representations \(A \cup B \cup C\) in memory binds them
together. Querying the memory with \(A \cup B\) retrieves the complete
pattern \(A \cup B
\cup C\) and returns \(C\) after subtracting the matching elements. This
mechanism constitutes the computational basis for ``chunking''.

So when the mind perceives ``red'' and ``apple'', it replaces those two
patterns with one combined ``redapple'' pattern, provided this is a
known combination. Chunking replaces two patterns with one dissimilar
pattern, which amounts to a higher-level representation. Repeated
application of this mechanism creates a hierarchy. And each step reduces
the number of active elements, which preserves overall sparsity. This
process happens ``in place'', operating on incoming data streams.

Encounters with novel combinations, such as ``blue banana'', trigger a
different process. Attempting to chunk those two items fails: the memory
query with the bundle ``blue banana'' gives an empty result. In this
case, the system needs to generate a new random holographic
representation for ``bluebanana'', and then bind it to the bundle ``blue
banana''.

In sparse distributed representations (SDRs), each element carries
specific semantic meaning. Our approach isolates semantic meaning to the
newly generated SHR, where each subset has the same meaning as the
whole.

An associative memory that can handle this mechanism at the time scale
of human perception must continuously generate and store random
representations on the fly. This implies a fast decay rate, guarding the
memory from filling up in no time. This short-term associative memory
works exactly like long-term topological memory, the only difference
being in the permanence of information. And we can also conclude that
this short-term associative memory is a source of higher-level sparse
holographic representations.

Fast associative learning correlates with short-term plasticity --- a
phenomenon observed primarily in the prefrontal cortex and parts of the
hippocampus.

\subsection{An explanation for childhood
amnesia}\label{an-explanation-for-childhood-amnesia}

Childhood amnesia --- the inability to recall memories prior to age
three or four --- lacks a consensus biological explanation. This
phenomenon turns out to be consistent with our topological model of
long-term memory.

A newborn possesses roughly one-quarter of the brain cells of an adult.
Most neurons develop during the first few years of life. Translated to
hyperdimensional terms, the universal set dimensionality quadruples
during early childhood. Assuming the brain maintains a constant sparsity
ratio throughout this growth, early memories are encoded by a much
smaller neural population. As the overall dimensionality expands, these
early encodings fall permanently below the adult pattern matching
threshold, making them inaccessible via associative memory retrieval.

\subsection{Step-wise response to threshold
modulation}\label{step-wise-response-to-threshold-modulation}

In the topological memory framework, the pattern matching threshold
\(T\) functions as a discrete guardrail, separating known symbols from
random noise. Because representations are based on discrete sets, the
memory cannot retrieve fractional elements.

This predicts a divergent, staircase-shaped ultrasensitivity in the
neural response to chemical modulation. Drug-induced lowering of \(T\)
results in a series of punctuated increases of activity in local
circuits that perform memory retrieval. Each ``step'' in the function
occurs at the moment the biological equivalent of the threshold drops by
one unit (for instance from 7 to 6), suddenly allowing smaller subsets
to match with information stored in memory. Due to the combinatorics of
subset matching, these steps become increasingly steeper.

This effect can manifest in various ways across different brain areas,
potentially causing sensory hallucinations as well as an impairment of
symbolic thinking. But this would be hard to observe at a system-wide
scale (via psychological experiments) where the aggregate behaviors of
many individual circuits average out. At a micro-circuit level, however,
a step-wise activity increase should be measurable via
electrophysiological experiments. Observing this effect would validate
our hypothesis that discrete set matching is a core operation of the
brain.

\subsection{A neural equivalent of stochastic
subsampling}\label{a-neural-equivalent-of-stochastic-subsampling}

Within the topological memory framework,
\hyperref[stochastic-subsampling]{stochastic subsampling} is a
fundamental mechanism which drives knowledge representation and
reasoning.

Associative memory retrieval involving composite symbolic data
structures --- SHR bundles, bindings and hypergraphs --- requires a
proportional decimation of active elements. This process breaks the
representational symmetry between bundled items, which allows a best
subset match to emerge.

This decimation factor \(D\) must lie within a narrow, optimal zone. If
decimation is too weak (\(D \lesssim 1\)), ties between competing
possible matches remain. If it is too aggressive (\(D \ll 0.8\)),
constituent SHRs are likely to be pushed below the memory's pattern
matching threshold. Values outside the optimal zone result in cognitive
failures, for instance the inability to recall known concepts.

Computational experiments show that subsampling by a decimation factor
\(D
\approx 0.8\) is most suitable to minimize these failure modes.

This model predicts the existence of a neural equivalent of stochastic
proportional decimation: a neuromodulatory mechanism that effectively
fine-tunes activity levels to optimize memory retrieval. This would
affect higher-level cognitive functions (driven by sparse holographic
representations) rather than perception (sparse distributed
representations). The effect would manifest as a peak in cognitive
performance, measured within a narrow interval around the modulatory
strength corresponding to \(D \approx 0.8\). This response curve is
shown qualitatively in the following figure.

\begin{figure}[H]
\centering
\includegraphics[width=0.675\linewidth,height=\textheight,keepaspectratio,alt={Cognitive performance as a response to a neuromodulation factor that corresponds to stochastic subsampling in memory retrieval.}]{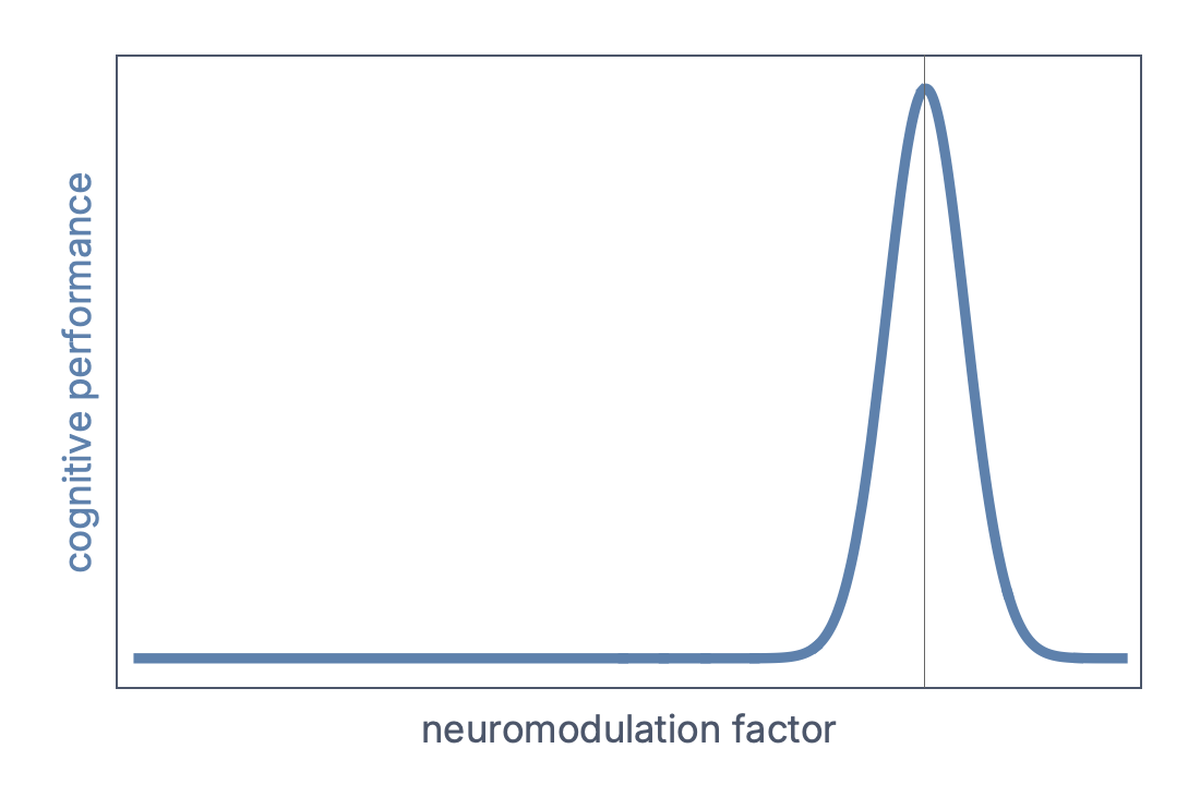}
\caption{Cognitive performance as a response to a neuromodulation factor
that corresponds to stochastic subsampling in memory retrieval.}
\end{figure}

\subsection{The signature of sparse holographic
representations}\label{the-signature-of-sparse-holographic-representations}

In our computational framework, hyperedges are bundles of sparse
holographic representations (SHRs) with uniform populations. This
predicts the existence of neural activity patterns where the number of
firing neurons manifests in tiers. These levels would roughly correspond
to integer multiples of the underlying SHR size, subject to small
interference from spurious overlaps.

Experimentally observing this signature pattern allows us to determine
the size of a biological SHR from the uniform distance between these
activity levels. This measurement provides an empirically grounded
hyperparameter for the development of biologically plausible AGI models.

\section{Appendix: Capacity
Calculations}\label{appendix-capacity-calculations}

This appendix details the derivations of the memory's storage capacity
for autoassociations, heteroassociations, and partitioned
representations. These calculations require only entry-level
combinatorics and should therefore be accessible to a wide audience.

We define the nominal capacity \(C\) of an associative memory as the
number of random associations that can be stored before half of the
memory's storage locations are set to 1. This corresponds to a memory
density of \(\rho^* = 0.5\).

Let \(V\) be the total number of memory storage locations (bits),
corresponding to the number of possible topological connections between
the memory's expanded hidden layer and every output element. This volume
depends on the memory's hyperparameters and the structure of
associations to be stored.

Each memory storage operation sets \(L\) bits to 1, regardless of their
previous state. Assuming associations are random and uniformly
distributed, the probability that any specific bit is not flipped to 1
during a single write operation is \(1 - L/V\).

The probability that a bit remains 0 after \(C\) independent write
operations is:

\[P_0 = \left(1 - \frac{L}{V}\right)^C\]

To calculate the nominal capacity, we set this probability to \(1/2\):

\[\left(1 - \frac{L}{V}\right)^C = \frac{1}{2}\]

Taking the logarithm of both sides:

\[C \ln\left(1 - \frac{L}{V}\right) =  -\ln(2)\]

Because the memory is highly sparse (\(L \ll V\)), we can apply the
approximation \(\ln(1 - x) \approx -x\) for small values of \(x\):

\[C \left(-\frac{L}{V}\right) \approx -\ln(2)\]

This gives the nominal capacity for a general associative memory:

\[C  \approx \ln(2) \frac{V}{L} \]

We now calculate the capacity of a hetero-associative memory, storing
associations of the form \(A \to B\). The input and output spaces are
defined by their dimensions \(N_A\) and \(N_B\), respectively, and by
their sparse populations \(P_A\) and \(P_B\). The total volume of binary
storage locations in this space is:

\[V_\text{hetero} = \frac{N_A (N_A - 1)}{2} N_B\]

Storing a single association affects \(L\) bits:

\[L_\text{hetero} = \frac{P_A (P_A - 1)}{2} P_B\]

Substituting \(V\) and \(L\) in the general formula derived above gives
the nominal storage capacity of hetero-associative memory:

\[C_\text{hetero}  \approx \ln(2) \frac{V}{L} = \ln(2) \frac{N_A (N_A - 1) N_B} {P_A (P_A - 1) P_B}\]

Auto-associative memory (\(A \to A\)) shares identical input and output
spaces (\(N_A = N_B = N\) and \(P_A = P_B = P\)). While one might expect
its capacity to be the same as the hetero-associative capacity using
these shared parameters, the auto-associative capacity is actually
higher by a factor of \(P/(P-2)\).

This increase is owed to the inherent redundancy of self-correlation. In
topological memory, a memory set \(M_{ij}\) links the input pair
\((i, j)\) to the output elements. But in an autoassociation, the output
pattern is identical to the input pattern. Therefore, the output always
includes \(i\) and \(j\) --- these elements are \emph{clamped}, which
means their topological connections do not carry information. The memory
only needs to record the connections from \((i, j)\) to the other
\(P - 2\) elements in the active set.

Adjusting our parameters for the informational space: The effective
number of output dimensions is \(N - 2\), resulting in an addressable
volume of

\[V_\text{auto} = \frac{N (N - 1)}{2} (N-2)\]

The effective number of bits flipped per pair during a write operation
is \(P - 2\). Storing a single autoassociation affects \(L\) bits:

\[L_\text{auto}  = \frac{P (P - 1)}{2} (P - 2)\]

Substituting these into the baseline capacity equation:

\[C_{\text{auto}} \approx \ln(2) \frac{N (N - 1) (N - 2)}{P (P - 1) (P - 2)}\]

Comparing this to hetero-associative memory (\(C_{\text{hetero}}\)):

\[\frac{C_{\text{auto}}}{C_{\text{hetero}}} =  \frac{P(N-2)}{N(P-2)}\]

For very large dimensions \(N\), we can approximate this scaling factor
by

\[\frac{C_{\text{auto}}}{C_{\text{hetero}}} \approx \frac{P}{P - 2}\]

We now determine the capacity of a bidirectional associative memory,
realized by partitioning auto-associative memory into two blocks,
\((A, B)\), each with dimension \(N\) and population \(P\). The combined
input has a total dimension of \(2N\) and a population of \(2P\).

To calculate capacity, we visualize the expanded latent space as a
triangular prism. The base of this prism represents all pair-wise
combinations of the \(2P\) input elements, and the height represents the
\(2N\) output elements. This base can be divided into three geometric
regions:

\begin{enumerate}
\def\labelenumi{\arabic{enumi}.}
\tightlist
\item
  Self-correlations in \(A\): A right triangle of size \(N \times N/2\).
\item
  Self-correlations in \(B\): A right triangle of size \(N \times N/2\).
\item
  Cross-correlations between \(A\) and \(B\): A square of size
  \(N \times N\).
\end{enumerate}

As random associations are stored, these regions do not fill at the same
rate. The memory's overall capacity is bottlenecked by the region that
reaches a density of \(\rho^* = 0.5\) first.

We first isolate the cross-correlating division. The total number of
addressable locations in this cross-correlating space involves the
\(N \times N\) pairs connecting to the \(2N\) output elements. But
because this memory represents autoassociations where two output
elements are clamped, the effective output dimension is only \(2N -2\).
The addressable volume of this space is given by:

\[V_{\text{cross}} = N^2 (2N-2)\]

The effective number of bits flipped per pair during a write operation
is \(2P - 2\). Storing an autoassociation affects \(L_{\text{cross}}\)
bits within the cross-correlating region:

\[L_{\text{cross}} = P^2 (2P - 2)\]

Substituting these values into the baseline capacity equation yields:

\[C_{\text{cross}} \approx \ln(2) \frac{N^2 (N - 1)}{P^2 (P - 1)}\]

Next, we calculate the capacity of the self-correlating divisions.
Adjusted for the two clamped output elements, each of the two divisions
with a triangular baseline has the volume

\[V_{\text{self}} = \frac{N (N-1)}{2}  (2N - 2)\]

The number of bits affected by storing an autoassociation within this
region is given by:

\[L_{\text{self}} = \frac{P (P-1)}{2}  (2P - 2)\]

The resulting capacity for the self-correlating division is therefore

\[C_{\text{self}} \approx \ln(2) \frac{N (N - 1)^2 }{P (P - 1)^2}\]

As expected, the capacity \(C_{\text{self}}\) is slightly larger than
\(C_{\text{cross}}\): the cross-correlating region fills up faster, and
therefore constrains the overall capacity of the bidirectional memory:

\[C_{\text{bam}} \approx \ln(2) \frac{N^2 (N - 1)}{P^2 (P - 1)}\]

We can compare this bipartite capacity to the capacity of a standard,
unpartitioned auto-associative memory. If we assume a total combined
dimension \(2 N\) and population \(2P\), the ratio is:

\[\frac{C_{\text{bam}}}{C_{\text{auto}}} = \frac{\ln(2) \dfrac{N^2 (N - 1)}{P^2 (P - 1)}}{\ln(2) \dfrac{2N (2N - 1)(2N - 2)}{2P (2P - 1)(2P - 2)}}  = \frac{N(2P - 1)}{P(2 N - 1)} \approx \frac{2P - 1}{2 P}\]

Thus, splitting an auto-associative memory with total population
\(P_\text{total}\) into two disjoint blocks reduces its nominal memory
capacity by approximately a factor
\((P_\text{total} - 1)/ P_\text{total}\).

Consider now the general case of a multi-partition auto-associative
memory. Following the derivation discussed for the bipartite memory, we
divide the footprint of the memory space (a prism) into self-correlating
and cross-correlating parts. This tiling is shown here for \(b = 4\)
partitions:

\begin{figure}[H]
\centering
\includegraphics[width=0.458\linewidth,height=\textheight,keepaspectratio,alt={Footprint of auto-associative memory partitioned in 4 equal-size blocks.}]{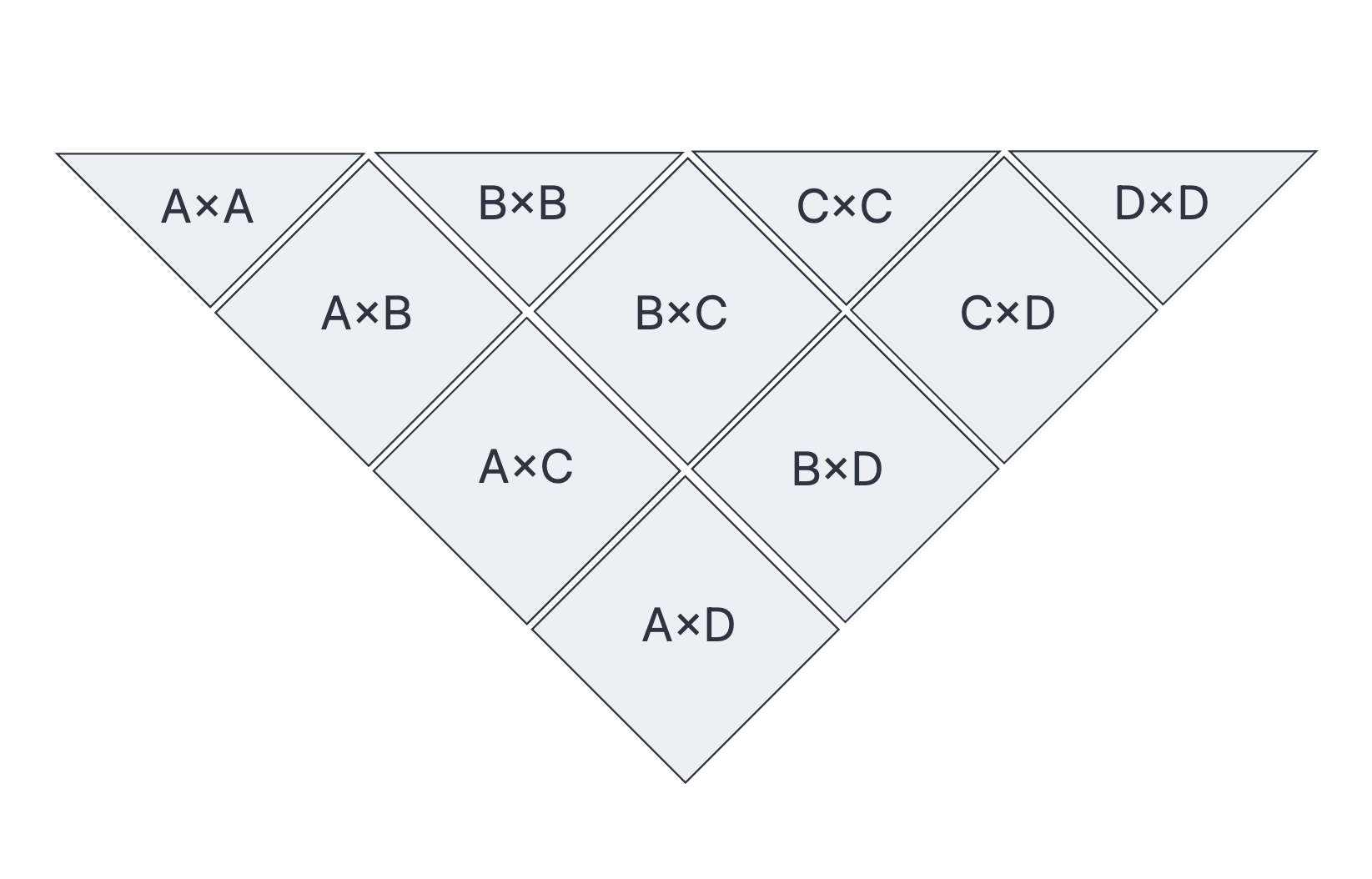}
\caption{Footprint of auto-associative memory partitioned in 4
equal-size blocks.}
\end{figure}

As in the bipartite case, the memory capacity is determined by the
cross-correlating divisions (the squares in the figure).

\[C_{\text{partitioned}} \approx \ln(2) \frac{N^2 (b N - 2)}{P^2 (b P - 2)}\]

Partitioning an auto-associative memory with total dimension \(b N\) and
total population \(b P\) into \(b\) blocks changes its capacity
according to this ratio:

\[\frac{C_{\text{partitioned}}}{C_{\text{auto}}} = \frac{\ln(2) \dfrac{N^2 (b N - 2)}{P^2 (b P - 2)}}{\ln(2) \dfrac{b N (b N - 1)(b N - 2)}{b P (b P - 1)(b P - 2)}}  = \frac{N(b P-1)}{P (b N-1)} \approx \frac{b P-1}{b P}\]

Because splitting the input layer into two blocks lowers the capacity,
one might expect this trend to continue as the memory is divided up into
more than two partitions. But the contrary is true: splitting an
auto-associative memory with total population \(P_\text{total}\) into
any number of disjoint blocks reduces its nominal memory capacity by
approximately a factor \((P_\text{total} - 1)/ P_\text{total}\),
independently of the number of blocks.

We conclude that block coding within auto-associative memory has only a
minor impact on storage capacity, regardless of the partitioning.

\section{Appendix: Relation to Deep
Learning}\label{appendix-relation-to-deep-learning}

The following table summarizes the main differences between the
connectionist deep learning paradigm and the computationalist framework
introduced in this work.

{\def\LTcaptype{none} 
\begin{longtable}[]{@{}
  >{\raggedright\arraybackslash}p{(\linewidth - 4\tabcolsep) * \real{0.2889}}
  >{\raggedright\arraybackslash}p{(\linewidth - 4\tabcolsep) * \real{0.3556}}
  >{\raggedright\arraybackslash}p{(\linewidth - 4\tabcolsep) * \real{0.3556}}@{}}
\toprule\noalign{}
\begin{minipage}[b]{\linewidth}\raggedright
\end{minipage} & \begin{minipage}[b]{\linewidth}\raggedright
\textbf{Deep Neural Networks}
\end{minipage} & \begin{minipage}[b]{\linewidth}\raggedright
\textbf{Topological Memory}
\end{minipage} \\
\midrule\noalign{}
\endhead
\bottomrule\noalign{}
\endlastfoot
AI paradigm & emerging from data & symbolic reasoning \\
network architecture & primarily feedforward & inherently recurrent \\
representation & emergent from weights & hypergraphs, nested data
structures \\
compositionality & soft (entangled) & strict (exact) \\
embedding & dense continuous dimensions & sparse hyperdimensional \\
memory model & scalar synaptic weights & discrete topology (open or
closed routes) \\
short-term memory & hidden states & topology with faster decay \\
core algorithm & dot product, matrix multiplication & subset matching \\
mathematical foundation & calculus, linear algebra & set theory,
combinatorics \\
learning rule & global (backprop) & local (topological plasticity) \\
learning dynamics & offline / batch & online / atomic \\
symbol grounding & entangled continuous spaces & heteroencoding SDRs to
SHRs \\
information decay & catastrophic forgetting / pruning & stochastic
decay \\
retrieval reliability & statistical approximation & exact pattern
matching \\
hardware architecture & von Neumann / memory wall & neuromorphic /
in-memory \\
\end{longtable}
}

\section*{References}\label{references}
\addcontentsline{toc}{section}{References}

\protect\phantomsection\label{refs}
\singlespacing

\end{document}